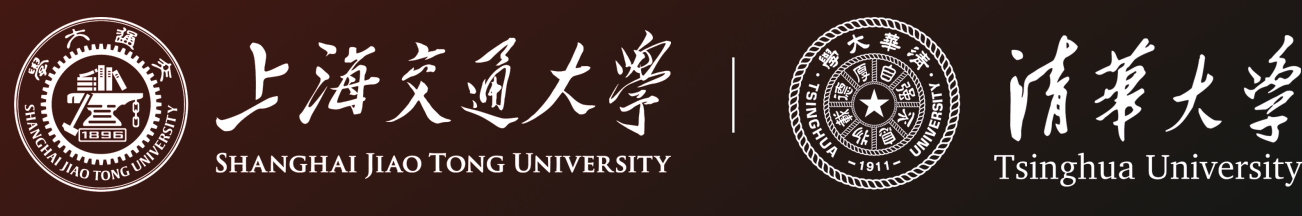

# AppCopilot: Toward General, Accurate, Long-Horizon, and Efficient Mobile Agent

*https://github.com/OpenBMB/AppCopilot*

Jingru Fan★†, Yufan Dang♠†, Jingyao Wu★, Huatao Li★, Runde Yang★, Xiyuan Yang★, Yuheng Wang★, Chen Qian★✉

★School of Artificial Intelligence, Shanghai Jiao Tong University

♠Department of Computer Science and Technology, Tsinghua University

fanjingru510@sjtu.edu.cn, dangyf21@mails.tsinghua.edu.cn, qianc@sjtu.edu.cn

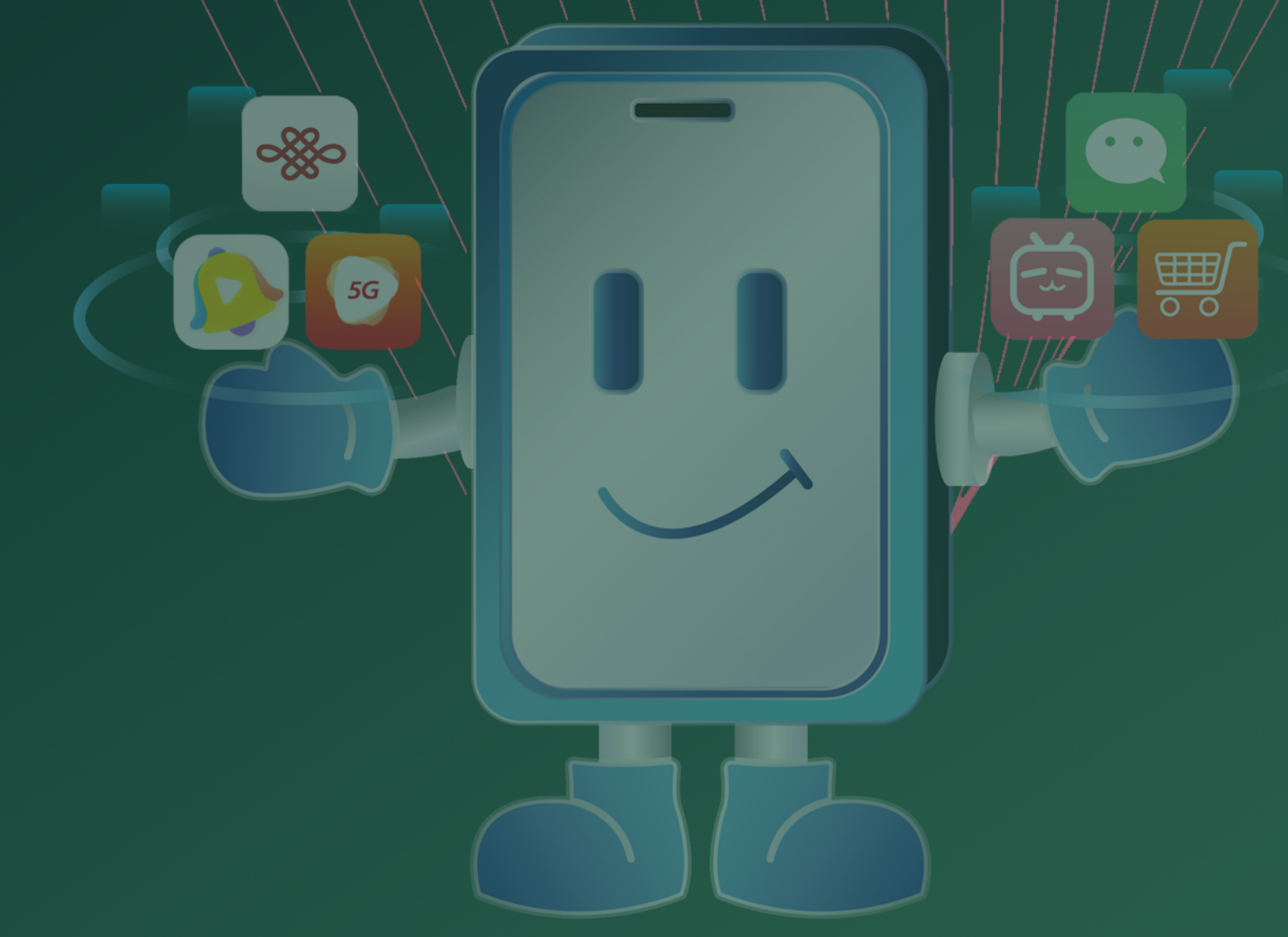

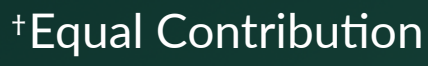

†Equal Contribution

✉Corresponding Author

# Release Notes

In version 1.0, we established the initial content framework and provided the core features along with basic documentation. Several chapters and examples were still being refined, and the preliminary build might contain errors that require further verification.

In version 2.0, we compressed and streamlined the content to highlight the key information, and correspondingly updated the author list. We improved the presentation with an optimized layout, clearer hierarchy, and more effective charts for better readability. We also corrected typos, terminology inconsistencies, broken links and citations, and other potential errors. Finally, we open-sourced the system by providing a public repository and appropriate licensing.

# Abstract

With the raid evolution of large language models and multimodal models, the mobile-agent landscape has proliferated without converging on the fundamental challenges. This paper identifies four core problems that should be solved for mobile agents to deliver practical, scalable impact: (1) generalization across tasks, APPs, and devices; (2) accuracy, specifically precise on-screen interaction and click targeting; (3) long-horizon capability for sustained, multi-step goals; and (4) efficiency, specifically high-performance runtime on resource-constrained devices.

We present AppCopilot, a multimodal, multi-agent, general-purpose mobile agent that operates across applications. AppCopilot operationalizes this position through an end-to-end autonomous pipeline spanning data collection, training, finetuning, efficient inference, and PC/-mobile application. At the model layer, it integrates multimodal foundation models with robust Chinese–English support. At the reasoning and control layer, it combines chain-of-thought reasoning, hierarchical task planning and decomposition, and multi-agent collaboration. At the execution layer, it enables experiential adaptation, voice interaction, function/tool calling, cross-APP and cross-device orchestration, and comprehensive mobile APP support. The system design incorporates profiling-driven optimization for latency and memory across heterogeneous hardware. Empirically, AppCopilot achieves significant improvements on four dimensions: stronger generalization, higher precision of on screen actions, more reliable long horizon task completion, and faster, more resource efficient runtime.

By articulating a cohesive position and a reference architecture that closes the loop from "data collection, training to finetuning and efficient inference", this paper offers a concrete roadmap for general purpose mobile agent and provides actionable guidance for both academic research and industrial adoption.

For updates and forthcoming releases, see the project page: https://github.com/OpenBMB/AppCopilot.

# Acknowledgements

We gratefully acknowledge the core contributions of the author team. Jingru Fan and Yufan Dang led the technical research and oversaw the full development workflow. Jingyao Wu, Huatco Li, Runde Yang, and Xiyuan Yang implemented key functional modules. Yuheng Wang contributed to the manuscript preparation and revisions. Chen Qian provided technical guidance and overall project management.

Additionally, we also thank Zhong Zhang, Yaxi Lu, Yankai Lin for technical consulting, Zhiyuan Liu, and Dahai Li for hardware support, and appreciate the suggestions from Yankai Lin and Zhiyuan Liu regarding paper compression.

# 1

# Research Background

## 1.1 Large Language Models

In the last couple of years, Pretrained Language Models (PLMs) have become the dominant technological paradigm in Natural Language Processing (NLP), profoundly driving the key process of artificial intelligence's evolution from perceptual intelligence to cognitive intelligence. This paradigm primarily consists of two stages: first, self-supervised pre-training on massive amounts of unlabeled text data to learn linguistic statistical structures and basic knowledge graphs; then, fine-tuning with limited supervision for specific tasks, enabling the model to adapt to particular downstream applications [69, 16, 53, 32, 47, 11, 38, 40]. With the continuous increase in model parameter scale, pretrained language models have gradually evolved into LLMs. In late 2022, OpenAI's release of ChatGPT demonstrated unprecedented capabilities in language understanding and generation, human–machine dialogue, and code generation, marking the formal establishment of the autoregressive model-dominated LLM paradigm [67, 64, 18, 2, 66, 5, 41]. Overall, model scale is widely regarded as a key factor influencing performance [30, 74, 73].

Along with technological evolution, research on optimizing performance and aligning behavior for practical applications of LLMs has also developed rapidly. First, how to achieve more efficient deployment and fine-tuning while maintaining performance for massive model parameters has become a significant topic. To this end, the academic community has proposed Parameter-Efficient Fine-Tuning (PEFT) techniques [26], which achieve performance close to full fine-tuning through the tuning of a limited subset of the model's parameters. Methods like LoRA [26], Prefix Tuning [37], and Adapter [25] are widely adopted. Additionally, Instruction Tuning guides models to understand and execute natural language instructions, enabling stronger task generalization in zero-shot or few-shot conditions, laying the foundation for the general interactivity of LLMs. To make a model's generated content more aligned with human expectations, Reinforcement Learning from Human Feedback (RLHF) [24] has been introduced. This mechanism guides the model to follow human value preferences and reasoning logic during generation, significantly enhancing its controllability and social acceptability.

However, despite the groundbreaking progress of LLMs on various tasks, they still have significant limitations, mainly in the following areas: 1) LLMs are essentially static response systems. Their interaction process is typically limited to a "single-turn input, single-turn output" mode. They lack dialogue memory, state maintenance, and dynamic behavior planning capabilities, making them unable to handle complex interactive tasks that require continuous decision-making and context preservation. 2) Traditional LLMs lack perception and control interfaces with external environments. They cannot acquire information, call tools, or perform operations in the physical world, which makes them inherently inadequate for tasks requiring real-time observation and execution feedback (e.g., web browsing, robot control). 3) LLMs are typically designed as general-purpose answerers, not as "agents" with autonomous goals, planning, and behavioral capabilities. They cannot decompose tasks based on long-term goals, manage resources, or coordinate strategies. They also lack the structured ability for multi-role, multi-step collaboration to complete a task. It is based on these technical limitations that academia and industry are gradually exploring how to embed LLMs as the core intelligence within an "agent" framework. This would give them the ability to think, act, and collaborate, thereby overcoming the bottlenecks of traditional language models in perception, task planning, and environmental adaptation. This *agentization* not only expands the boundaries of language models but also provides a key opportunity for building a new generation of cognitive intelligent systems.

## 1.2 Autonomous Agents

Amid the rapid development of LLMs, the concept of the autonomous agent is widely considered a natural extension of their capabilities, and a crucial step in the models' transition from "Q&A tools" to "cognitive agents". Although LLMs excel at tasks like language understanding, summarization, Q&A, and code generation, their core remains a passive language system. They lack goal awareness, behavioral control, and the ability to execute complex decisions, making it difficult for them to independently complete full task workflows in open environments. To overcome this limitation, academia and industry have proposed the "autonomous agent" paradigm, which seeks to endow language models with external tools and behavioral planning capabilities, enabling them to complete multi-stage tasks in a more intelligent and autonomous way.

The basic idea of an autonomous agent is to use the LLM as the central "cognitive hub", combining it with peripheral modules such as task planning, tool interfaces, memory systems, and role-playing. Together, they form a closed-loop system with the ability to perceive, think, act, and self-reflect. This type of system can autonomously formulate plans, call external tools, perceive environmental feedback in real time, and adjust its process based on a task goal, thereby completing complex tasks with minimal human intervention [81, 61]. From an architectural perspective, autonomous agents are highly modular. The language model itself is responsible for semantic understanding and high-level reasoning, while peripheral modules supplement its capabilities in environmental interaction, information processing, and state maintenance. These modules work together to build a closed-loop structure around the language model, turning it from a passive conversational system into a composite agent with a degree of autonomous behavior.

To support autonomous agents in completing complex tasks, their architecture is typically built around four key technical modules: Planning, Tool Use, Memory Mechanisms, and Role-Playing. These modules, in synergy with the LLM, provide the agent with a complete capability chain, from understanding the problem and formulating a plan to calling resources and generating feedback. Figure 1.1 shows a typical autonomous agent architecture.

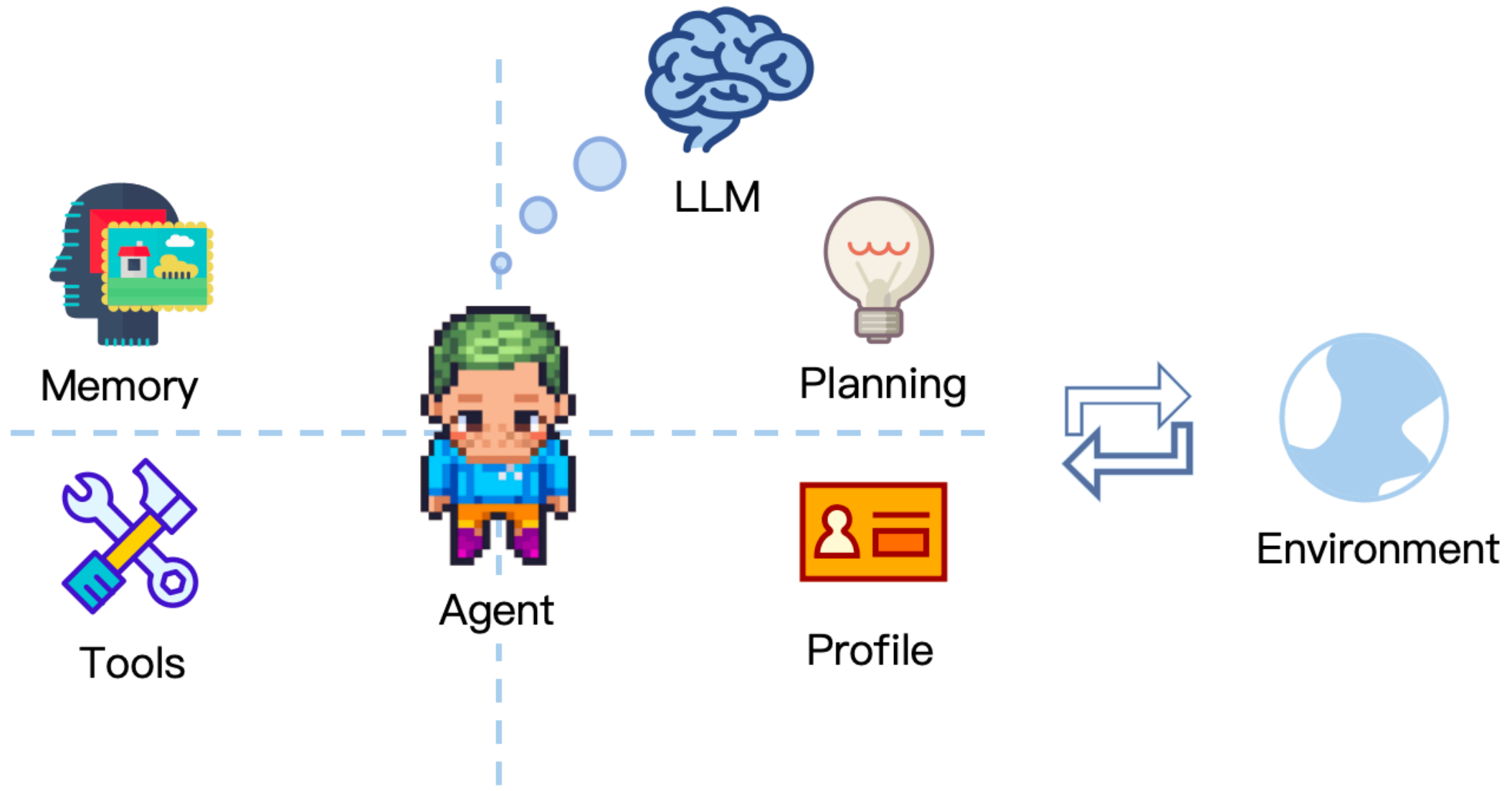

**Figure 1.1:** *Autonomous Agent Framework*

## 1.3 Multi-Agent Collaboration

With the continuous expansion of LLM capabilities, their performance in single-agent tasks has become quite mature. However, when faced with complex, dynamic, and heterogeneously distributed tasks, a single agent struggles to cover all processes. It is limited by a restricted cognitive perspective, tight task coupling, and a lack of behavioral flexibility. Therefore, to further enhance a system's task adaptability, scenario generalizability, and collaborative efficiency, research is shifting from building single, fully-functional agent systems to creating collaborative systems composed of multiple agents with different capabilities and specializations—that is, multi-agent systems. Through task division, negotiation, communication, and game theory among agents, complex tasks can be processed in parallel and completed efficiently, which is a key path for moving LLMs toward higher levels of intelligence. Multi-agent systems are not a new concept. Traditional research often used centralized or distributed control architectures, combined with methods such as reinforcement learning, game modeling, policy transfer, communication protocols, and hierarchical learning, to optimize individual agents' perception, reasoning, and decision-making capabilities, thus building systems that can collaborate effectively on specific tasks. By using natural language interaction as a unified coordination mechanism, LLM-driven multi-agent systems have achieved a deep coupling of cognition and collaboration. On one hand, agents can use language for efficient knowledge sharing, task synchronization, and decision transmission, reducing collaboration costs. On the other hand, the generalization capa-bilities of LLMs enable the system to show stronger adaptability when facing unknown scenarios and goals. This new paradigm not only enhances the naturalness and scalability of collaboration but also reduces reliance on manual rules and supervised signals, providing a feasible path for building general, multimodal, multi-task collective intelligence systems. A typical LLM-driven multi-agent system framework is shown in Figure 1.2.

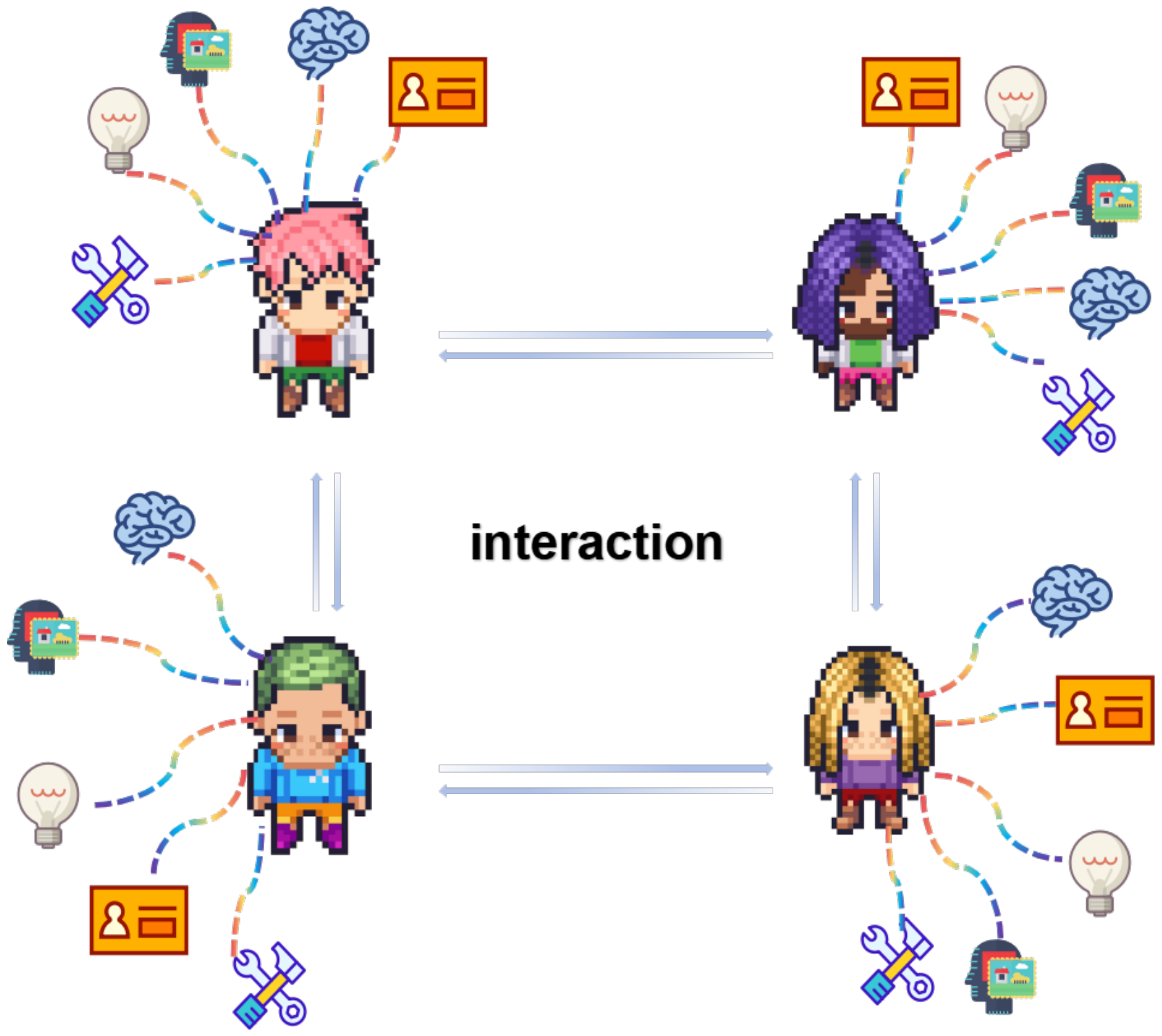

**Figure 1.2:** *Multi-agent System Framework*

## 1.4 Mobile Agents

As the demand for complex model interaction environments grows, the tool-use capabilities of large models are gradually extending to the visual modality. Compared to the text modality, visual tool learning is closer to real-world interaction, enabling complex task manipulation directly through the perception and understanding of visual information from graphical interfaces or real-world environments. In contrast to traditional scripted automation or rule-based interface control, a mobile agent can understand complex task intentions in both visual and linguistic modalities and automatically generate cross-application, multi-step interaction sequences. This brings new opportunities for fields like software testing, human–machine collaboration, and accessibility technology.

International research institutions have been at the forefront of this exploration. OpenAI, building on ChatGPT, launched a prototype for web and desktop operations called "Operator", demonstrating the language model's ability to automatically fill out forms and perform web searches through control recognition. Google DeepMind's AndroidEnv [68] uses reinforcement learning to train agents to automate operations on Android APPs, and it uses deep networks to in-tegrate visual information with environmental states, enhancing the feasibility of deployment on real devices. ByteDance has developed a cross-platform GUI operation model software, attempting to enable large models to achieve consistent interface perception and decision-making on various terminals such as mobile phones, PCs, and the Web [52]. The academic community has also produced numerous results. Tsinghua University proposed AutoWebGLM [31], which integrates the GLM model with web interaction data. It builds a diverse web scenario dataset through "human–computer collaboration + self-supervised learning" to enhance the model's adaptability in real networks.

Despite rapid research progress, mobile agents still face numerous challenges: recognition deviations and action failures are common in dynamic interfaces, model deployment performance on edge devices does not yet meet real-time requirements, secure read/write mechanisms for user interface information are incomplete, and task transfer and generalization capabilities are still insufficient. Most current achievements are concentrated on specific applications or benchmark scenarios and have not yet truly achieved general intelligence for diverse GUI environments.

*Representative Mobile Agents*

Recent vision–language agents [9, 90, 14, 46, 1, 12] focus on end-to-end UI understanding and action execution. UI-TARS is a purely vision-driven mobile agent that maps screenshots to low-level actions with unified action modeling, multi-step reasoning, and self-iterative data growth, outperforming closed-source baselines like GPT-4o and Claude on GUI benchmarks [52]. Qwen-VL provides strong multimodal abilities (visual localization, OCR, multi-image dialogue) via a vision encoder, staged training, and multilingual data, surpassing similarly sized models on standard tasks; its chat variant is optimized for real dialogues [3]. AgentCPM-GUI targets Chinese mobile APPs with an on-device design, mixed Zh-En data, progressive training (perception pretrain, SFT, GRPO), and compact JSON actions, achieving robust cross-APP generalization across 30+ APPs [86]. InternVL3 unifies multimodal and language training, introduces V2PE for long contexts, and applies SFT/MPO, reaching state-of-the-art results across multimodal benchmarks while retaining strong pure-language ability. Foundational GUI/action modeling efforts tackle data scarcity and cross-platform consistency. OS-Atlas releases a cross-platform GUI localization synthesis toolkit and the largest open corpus, resolving action naming conflicts to build a general high-accuracy action model that beats prior closed-source systems on desktop, mobile, and web benchmarks [77]. OS-Genesis proposes inverse task synthesis: agents first explore GUIs, then retrospectively generate high-quality tasks, boosting diversity and performance on AndroidWorld and enabling end-to-end data creation without manual labels [62]. AGUVIS removes dependence on text and closed-source models via a unified vision-only framework, a large cross-platform dataset with inner-monologue reasoning, and a two-stage training pipeline decoupling localization from planning, matching top closed-source performance on related datasets [79, 49]. More broadly, open reasoning agents integrate planning with multimodal perception. GPT-OSS adds agentic tool-use post-training to decompose instructions and act on screenshots, documents, and webpages [44]. GPT-5 advances a routed system combining a fast "smart model" with a deeper "thinking" model, improving multimodal understanding (voice, video, audio), lowering hallucinations, and providing more reliable complex-task performance [45, 82].

# 2

# Existing Problems

## 2.1 Generalization: Insufficient Generalization Capabilities across Applications

With the rapid development of mobile agents, although significant progress has been made in various benchmark tests, there are still many shortcomings in open and variable real-world application scenarios, highlighting the core bottleneck of "insufficient generalization". This bottleneck is not caused by a single factor but stems from a systematic disconnect between current training data and the complexity of the real world across multiple dimensions. This chapter will conduct an in-depth analysis of the key challenges constraining model generalization from three progressive dimensions: regional ecological differences, internal structural diversity, and the behavioral authenticity of data.

### 2.1.1 Data Resource Scarcity in Chinese Scenarios

In the process of building mobile agents with generalization, task execution, and environmental adaptation capabilities, high-quality data resources have always been a fundamental support. Training data not only carries the model's ability to understand interface structures, user intentions, and operational logic but also directly determines its stability and robustness when performing tasks in real scenarios.

However, despite the significant progress in mobile agent research in recent years and the emergence of a series of datasets and evaluation benchmarks—such as those designed for grounding, which aim to associate natural language instructions with on-screen controls, with representative works including Mind2Web [15] and OS-ATLAS [77]; and other datasets focusing on agent task execution, representing tasks as sequences of "action–observation pairs", such as AITZ [83], AndroidControl [36], and GUI-Odyssey [42] The existing training data as a whole still heavily relies on English semantic environments and English interface designs (as shown in Table 2.1). In the Chinese context, there is still an extreme scarcity of GUI operation task data for real-world APPs, which restricts the application and promotion of models in Chinese scenarios.

The language prompts, labels, and content in Chinese APPs are predominantly in Chinese, and their interface structures, control styles, and interaction sequences differ significantly from English applications, making direct transfer from English training data challenging. Chinese interfaces often follow local user habits, featuring complex nested menus, vertical text, multi-level navigation paths, and highly localized elements such as ambiguous icons, Pinyin abbreviations, and local symbols, which are rare in English APPs. This complexity makes it difficult for models to accurately identify, locate, and interpret controls, affecting operational accuracy. Moreover, existing Chinese GUI datasets are limited, mostly focusing on low-level tasks like static screenshots, OCR recognition, or control classification, and lacking dynamic "action-observation" sequences, continuous operational flows, clear user task objectives, and semantic annotations of user behavior. As a result, models struggle to learn complete task planning and execution logic from these datasets.

**Table 2.1:** *Overview Comparison of Major Mobile Agent Datasets*

| Dataset Name | Main Task | Data Scale | Language | Platform | Operating System |
|---|---|---|---|---|---|
| Mind2Web [15] | Grounding / Agent | 2,350 tasks; 137 websites | English | Web | N/A |
| ScreenSpot-Pro [34] | Grounding | 1,581 instructions | English | Desktop, Mobile, Web | Windows, macOS, Linux |
| OS-ATLAS [77] | Grounding | Over 13 million elements; 2.3 million screenshots | English | Desktop, Mobile, Web | Windows, Linux, macOS, Android |
| GUICourse [8] | Grounding / Agent | 73,000 web tasks, 9,000 mobile tasks | English | Web, Mobile | Android |
| UGround [23] | Grounding | 10 million elements; 1.3 million screenshots | English | Web, Desktop, Mobile | Windows, Linux, macOS, Android, iOS |
| AITW [54] | Agent | 715,000 trajectories; 30,000 instructions | English | Mobile | Android |
| AITZ [83] | Agent | 18,643 screen–action pairs; 2,500 instructions | English | Mobile | Android |
| AndroidControl [36] | Agent | 15,283 demonstrations; 14,500 tasks; 833 APPs | English | Mobile | Android |
| GUI-Odyssey [42] | Agent | 8,834 trajectories; 212 APPs | English | Mobile | Android |
| AMEX [6] | Agent | Over 104,000 screenshots; 110 APPs | English | Mobile | Android |
| AndroidWorld [55] | Agent | 116 programmatic tasks; 20 APPs | English | Mobile | Android |
| E-ANT [71] | Agent | 40,000+ trajectories | Chinese | Mobile | Android |
| Mobile3M [76] | Agent | 20,138,332 actions | Chinese | Mobile | Android |
| CAGUI [86] | Grounding / Agent | 3,000 grounding data, 600 agent tasks | Chinese | Mobile | Android |

### 2.1.2 Lack of Data Diversity in Specialized Scenarios

While achieving significant progress in benchmark tests, existing mobile agents still face severe challenges in their generalization capabilities in vertical domains. The lack of generalization mainly stems from systematic biases in training data across two dimensions: insufficient application coverage breadth and the homogenization of task types.

Firstly, at the application level, most existing training datasets show a long-tail distribution, focusing primarily on a few mainstream applications—such as YouTube and Amazon Shopping in English contexts, and WeChat and Taobao in Chinese contexts. Taking the large-scale Chinese Android dataset Mobile3M [76] as an example, its data distribution also confirms this phenomenon, with the vast majority of samples coming from a few popular applications. Annotated data for a large number of specific domains (such as finance, communication, etc.) or emerging applications is extremely scarce. This makes it difficult for the model to learn diverse interface layouts, interaction patterns, and user intentions. When the model is deployed in a new, unseen application environment, its performance often plummets, severely limiting its practical value.

Secondly, at the level of data content, a significant number of benchmark datasets exhibit a distinct predisposition towards specific task typologies. Notably, navigation tasks account for a disproportionately high percentage of the data [83] [42] [63]. The fundamental objective of these tasks is to direct an agent through a sequence of operations—such as clicks and swipes across various application interfaces-based on natural language commands, with the ultimate goal of arriving at a designated target page. This bias is especially evident in several major large-scale datasets. For example, in AITZ [83], one of the largest collections of real-world mobile interactions, most task trajectories follow a canonical navigational pattern, typically abstracted as "$\text{open APP} \rightarrow \text{multi-step click/swipe} \rightarrow \text{complete objective}$".

Such a distributional bias in the training data induces model overfitting to navigation-specific instruction patterns and environmental cues. While the resulting agent may exhibit proficiency in simple navigational tasks, it fails to acquire the more sophisticated capabilities required for understanding and decomposing complex instructions. Consequently, a significant degradation in performance is observed when the agent is confronted with composite tasks that require a fusion of distinct operational steps, such as navigating to a page before performing information extraction. Moreover, existing datasets are markedly deficient in their support for cross-APP tasks. Many real-world user workflows—for example, *find a restaurant in a map APP, copy its address, then switch to WeChat and send it to a friend*—necessitate seamless inter-application control and data transfer. The current datasets, with their overwhelming focus on intra-application operations, are insufficient for endowing agents with the generalization capabilities required for such real-world scenarios, thereby severely constraining their practical utility.

In summary, the lack of application breadth and task depth are intertwined, collectively creating a hard-to-break ceiling that limits the model's generalization ability. This dual dilemma indicates that simply increasing the amount of homogeneous data can no longer bring about fundamental improvements. How to design and construct a training dataset that is more balanced, diverse, and challenging in both the application and task dimensions is a significant current challenge.

### 2.1.3 Challenges in the Authenticity of Behavioral Data

In the development of generalizable mobile agents, the behavioral authenticity of training data emerges as a factor that is both decisive and exceptionally challenging. A robust foundation for generalization can only be established if the data fully encapsulates the inherent diversity and complexity of genuine user operation processes. This high-fidelity requirement manifests as two core challenges: interaction execution generalization at the micro-level, and task strategy generalization at the macro-level.

At the micro-level of interaction execution, even after an agent has successfully formulated a rational plan and identified the correct control for the next step, it must still address the challenge of spatial position generalization (see Figure 2.1). Note that contemporary mobile agents typically employ pixel-level localization, where an action is parameterized as a single coordinate point $(x, y)$ on the screen. This representation is fundamentally misaligned with the nature of GUI controls, whose interactive affordance spans a spatial region (i.e., a bounding box) rather than a discrete point. Training data often exacerbates this issue by recording only a single, specific location clicked by a human demonstrator (e.g., the geometric center), inducing a bias that causes the model to overfit to this point-based representation while failing to learn the concept of the broader interactive area where multiple points are viable.

This brittleness becomes particularly evident under conditions of UI dynamism, such as when controls are rearranged, resized, or partially occluded. For instance, a minor shift in a button's visual position on a different terminal device can cause a coordinate-based action to result in an "off-target" click, failing the interaction even if the control's semantic identity was correctly recognized. Therefore, to enhance interactional robustness, the agent must develop a structural understanding of a control's interactive region and acquire the ability to dynamically select a viable operation point within that valid area, rather than merely reproducing a previously observed coordinate. Second, at the task strategy level, the model must handle the diversity and non-determinism of paths leading to the same goal. Although these paths are functionally equivalent, their interaction logic, interface navigation structure, and reliance on contextual information are all different. Most current training datasets only cover one or two "golden paths", which makes the model more inclined to learn fixed operational flows rather than context-adaptive task planning strategies. Therefore, if the interface layout is adjusted or a recommended entry point is missing, the model may fail the task due to a path mismatch, reflecting its lack of a semantic-level generalized understanding of the task objective.

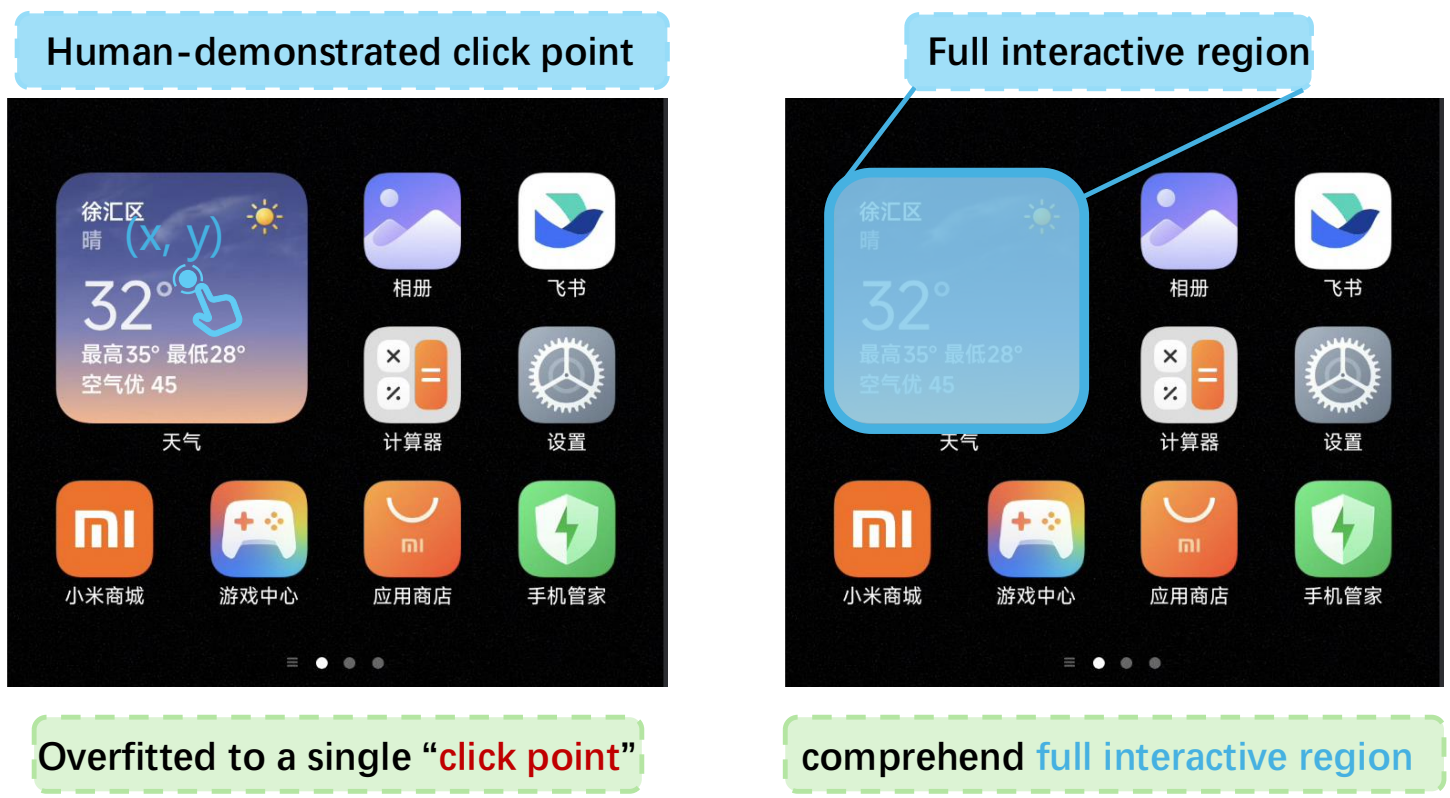

**Figure 2.1:** *Illustration of Target Control Spatial Generalization.*

## 2.2 Accuracy: Weak Single-Step Control Accuracy

During task execution, the ultimate success of a mobile agent depends not only on its high-level strategic planning capabilities but is also directly limited by the accuracy of each basic control step. Currently, even if an agent can correctly understand the task intent, it still exhibits significant fragility at the single-step operation level, specifically manifesting in three core problems caused by technical architecture, positioning granularity, and the model's own uncertainty.

### 2.2.1 Error Accumulation Caused by Non-End-to-End Models

In terms of technical architecture, some current mobile agent systems adopt a modular, non-end-to-end design. This paradigm is centrally reflected in the current agent frameworks. Its core mechanism is to decompose complex tasks through a preset Workflow and chain together a series of relatively independent functional modules. These modules—such as the "Interface Understanding Module" for visual perception, the "Task Reasoning Module" for decision-making, and the "Action Execution Module" for specific operations—are orchestrated and communicate via Prompt Engineering or external scripts [52]. Although this divide-and-conquer design has certain advantages in development efficiency and functional decoupling, its inherent flaw is that local errors in the reasoning process accumulate and amplify along the multi-stage chain, ultimately leading to unstable task performance and difficulty in holistic optimization towards a unified objective.

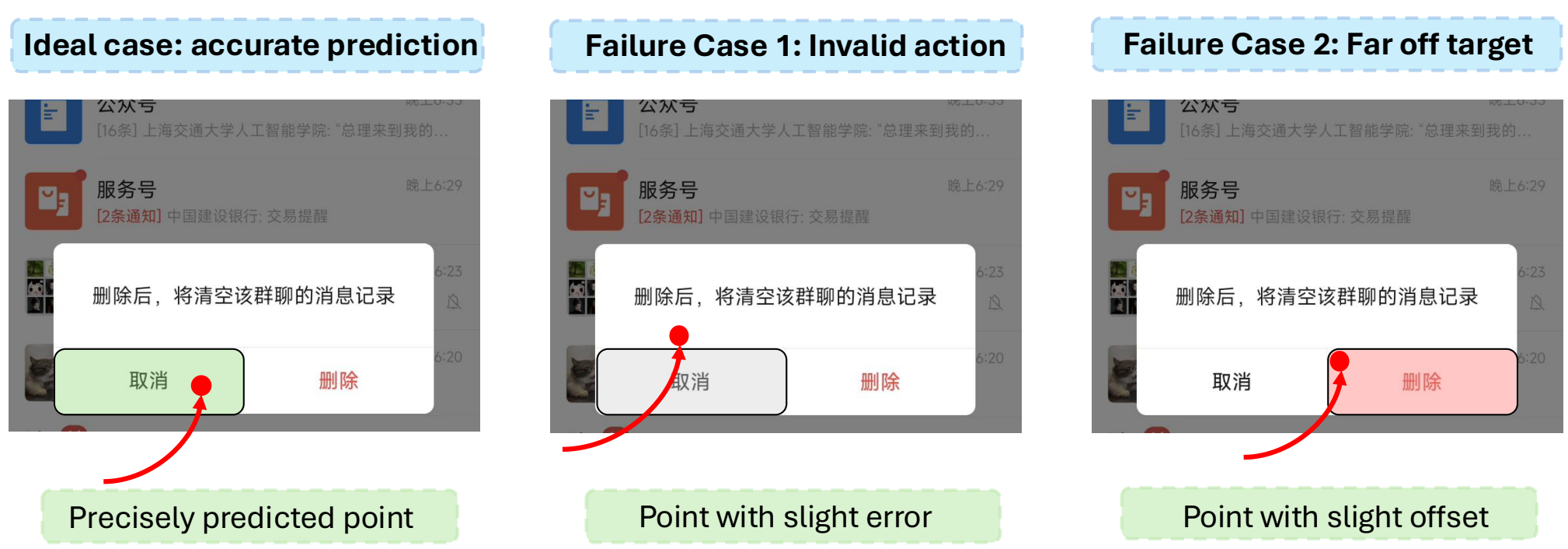

**Figure 2.2:** *Positioning Deviations in Control Interaction*

Under this "pipeline" processing mechanism, any slight error from an upstream module can be inherited and amplified by downstream modules, eventually having a fatal impact on task completion. For example, the interface understanding module might misidentify "Confirm Submission" as "Confirm Proposal" due to blurry fonts or insufficient recognition accuracy. When this incorrect input is fed into the task reasoning module, it may lead the language model to mistakenly judge that the current page lacks a submission entry, thus generating erroneous instructions like "scroll down to find the submission button". Once such an error occurs, subsequent modules can hardly correct it and can only execute the instruction, causing the operation to deviate from the correct trajectory. Overall, the system's single-step decision accuracy approaches the product of the accuracies of each module, forming a typical cascading error accumulation mechanism that significantly weakens the system's stability and interaction success rate.

More critically, this modular design faces the challenge of "indirect optimization" during the training phase. Each submodule is often optimized independently around its own local objective. However, these local optimization goals often deviate significantly from the global task success rate, which is what the system truly cares about. Taking the "Confirm Submission" recognition error as an example, from the visual module's perspective, this might just be an insignificant character-level mistake, but in the context of the global task, it could cause task failure. Because the current architecture lacks an end-to-end error attribution mechanism that propagates from the overall task success/failure result back to the perception and understanding modules, the system cannot effectively capture and correct these "high-impact" local errors. This disconnect between objective functions greatly limits the potential for synergistic optimization of the system's overall performance.

### 2.2.2 Excessively Pixel-Level Positioning Granularity Leads to Control Errors

At the action execution level, many current methods use pixel-level coordinates (x, y) as the final output to specify the click position. However, this excessively fine positioning granularity is misaligned with the physical reality of GUI interaction, making it extremely sensitive to prediction errors and a direct cause of control failures.

A fundamental discrepancy exists between the model's action parameterization and the physical reality of GUI interaction. While an interface control's interactive affordance is a spatial region (i.e., a bounding box), the agent is typically trained to predict a discrete coordinate point (x,y) within this area. This granularity mismatch renders the system highly sensitive to prediction errors. Consequently, even a minor deviation of a few pixels in the predicted coordinates can cause the action to fall outside the intended boundary, nullifying the operation. This vulnerability is exacerbated in dense interfaces where adjacent controls with opposing functions, such as "Delete" and "Cancel", are present in Figure 2.2. In such cases, a slight coordinate shift can trigger an unintended and potentially critical action. This problem is particularly pronounced in dynamic GUI environments characterized by responsive layouts, varying device resolutions, or asynchronous content loading, as these factors compromise the generalizability of a specific coordinate learned in a static context.

### 2.2.3 Generative Model Reliability is Affected by Sampling Fluctuations

In action generation, the core decision-making ability of modern mobile agents typically relies on a generative reasoning mechanism driven by LLMs. However, the inherent random sampling strategy in LLMs, while enhancing the diversity and flexibility of language generation, inevitably introduces non-deterministic fluctuations, becoming a significant factor that constrains stability and reproducibility.

In practical applications, to avoid monolithic generation results, LLMs naturally adopt decoding strategies with randomness, such as temperature sampling or top-k/top-p sampling. This mechanism means that even with identical inputs (e.g., the same screen state and task instruction), the intermediate chain of thought and the final action decision generated by the model may differ across different runs. This uncertainty in the generation process can easily lead to so-called "accidental failures" or "behavioral jitter".

This behavioral inconsistency induced by the sampling mechanism presents a significant challenge for mobile agents in high-reliability scenarios (e.g., financial transactions and medical applications). On the one hand, the system cannot ensure deterministic mapping from identical inputs to consistent decisions, which undermines task flow controllability and complicates debugging. On the other hand, such instability heightens user uncertainty, thereby eroding the foundation of trust in human-computer interaction.

## 2.3 Long-Horizon Capability: Limited Long-Range Task Orchestration

One of the core hallmarks of a general agent is its ability to understand, plan, and execute complex long-range tasks over extended time scales. However, current mobile agents are mostly confined to single-step or short-range operations, lacking effective planning and execution capabilities for long-term goals and complex processes. This deficiency in "long-range capability" is specifically reflected at multiple levels, from causal reasoning within sequences to collaborative work across applications and even across devices.

### 2.3.1 Lack of Supervised Reinforcement for Sequential Reasoning

In the process of interaction between a mobile agent and its environment, each time step follows the loop of "perceive–reason–act": first, it needs to perceive the current interface state; then, it reasons based on the high-level task goal and historical operation records to decide the most reasonable next action; finally, it translates this decision into a specific command and acts upon the environment. This process is not a one-time static calculation. Each action of the agent changes the state of the environment, and this new state becomes the input for the next "perceive–reason–act" loop. This series of interlocking loops constitutes a complete "decision chain" or "action trajectory". Therefore, the ultimate success of a task depends not only on the accuracy of single-step operations but more so on the logical coherence and cumulative effectiveness of the entire reasoning chain.

However, the training paradigm of many current mobile agents heavily relies on "single-step behavior supervision" based on imitation learning. Its core method is to train the model to imitate the next single-step action given the current screen state. The fundamental flaw of this method is that it "slices" a continuous task sequence along the time dimension, with the implicit assumption that each "observation–action pair" is independently and identically distributed. This oversimplified approach inevitably leads the model to focus only on the "local optimum", i.e., how to accurately reproduce the immediate step, while neglecting to model and reinforce the long-range logical dependencies between steps [85] [20].

Although methods based on single-step imitation learning have advantages such as simple implementation and clear supervision signals in model training, such "local optimum" strategies often perform poorly in actual multi-step tasks. The fundamental reason is the model's lack of ability to model the internal structure and causal relationships of the task [75] [60] [59]. In GUI scenarios, an agent's operations are not composed of a series of isolated actions but exhibit highly structured process characteristics, with significant temporal dependencies and logical prerequisites between steps. The single-step imitation learning approach focuses on learning "what to do" in a specific interface state, rather than understanding "why to do it". This causes the model to exhibit "locally reasonable but globally incorrect" behavior when executing multi-step tasks, which is essentially due to a lack of modeling capability for the causal structure between operations.

Furthermore, in long-sequence reasoning tasks, this training mechanism that ignores the global causal chain will lead to the "temporal credit assignment" problem: a seemingly harmless minor misoperation early on, such as incorrectly selecting a shipping address in an e-commerce scenario, may only manifest its impact as a task failure a dozen steps later at the payment stage, yet the model cannot reasonably attribute this failure to the preceding error [78] [35]. Therefore, without sequence-level supervision, the agent can hardly extract useful signals from task feedback and achieve self-correction, thus finding it difficult to complete long-range tasks.

### 2.3.2 Limited Complex Task Planning and Decomposition Capabilities

When executing long-range tasks in the real world, in addition to requiring the agent to have reasoning capabilities across long sequences of steps, a more critical part is its ability for structured planning and dynamic decomposition of complex tasks [28] [27] [84]. This means that the agent not only needs to interact specifically with the GUI interface and complete low-level operations but also needs to possess high-level planning capabilities at an abstract level. It should be able to accurately infer hidden intentions and constraints from high-level, semantically ambiguous instructions from the user, and autonomously break them down into a series of clear, executable sub-goals, thereby constructing a hierarchical and operationally feasible overall task plan.

However, at the current stage, mobile agents based on LLMs, despite having strong commonsense reasoning and language generation capabilities, still face multiple challenges in terms of truly effective task planning. First, effective planning requires a deep analysis of the user's ambiguous intent. The agent needs to be able to take high-level, semantically vague instructions from the user and accurately decompose them into a series of well-defined, logically consistent, and operationally feasible sub-tasks, ultimately forming a hierarchical overall task plan. Second, a plan that appears logically perfect at the text level may be ineffective if it is detached from a deep understanding of the real GUI environment. Current agents, when planning, often produce "hallucinations" and generate impractical steps due to their inaccurate understanding of the GUI environment. Finally, the agent lacks adaptability and correction capabilities when facing the dynamism of the real world. Once a plan is hindered by interface changes or operational failures, agents generally lack effective failure diagnosis and dynamic replanning mechanisms, making it difficult to recover from setbacks and adjust their strategies.

### 2.3.3 Insufficient Coordination and Linkage for Cross-Application Tasks

In reality, user tasks are often not confined to a single application but unfold naturally across applications, forming a "task chain" that spans multiple system components. For example, a user might receive an address in Enterprise WeChat, copy it, switch to AutoNavi Map for navigation, and then take a screenshot of the estimated arrival time to share back to Enterprise WeChat. Such cross-application operations are extremely common in real scenarios, reflecting the highly systemic and collaborative nature of user task flows.

However, the design and training paradigm of most current mobile agents is still confined within a single APP, lacking the ability to model cross-application tasks globally. Achieving efficient cross-application collaboration faces multiple technical bottlenecks: Loss and forgetting of context, Drift and loss of sub-task goals and Insufficient adaptability to heterogeneous interaction interfaces. Therefore, breaking through the limitation of "single-application closed execution" and enabling the agent with cross-application perception, memory, and planning capabilities—evolving towards a ubiquitous assistant with system-level context management and cross-domain orchestration capabilities is the key path to achieving automation of real-world long-range tasks.

### 2.3.4 Cross-Device Collaborative Mechanisms Not Yet Established

The ultimate challenge for long-range tasks lies in extending the agent's capabilities from serving a "single user" to supporting the collaborative work of "multiple users" [39] [48]. This is not just an extension of capability but a paradigm shift from a "personal intelligent assistant" to a "distributed collaborative network". In this vision, each user's device runs an autonomous agent representing their individual preferences and intentions. These agents can discover, communicate with, and collaborate with each other to jointly accomplish a complex goal that is beyond the capability of a single agent.

The establishment of this collaborative paradigm faces three core challenges. First, in a multi-user environment, the task states of different users are naturally distributed across different devices and private contexts. The agent needs to perform reasoning and coordination with incomplete information, possessing the ability to handle partially observable environments. Second, communication and negotiation mechanisms need to be built between agents. Agents must express intentions, divide tasks, and provide feedback through structured, semantically aligned interaction protocols to ensure policy consistency in asynchronous and heterogeneous environments. Finally, when their respective goals conflict (e.g., resource competition, temporal inconsistency), the system needs an effective consensus mechanism to achieve an overall optimal collaborative solution while satisfying individual constraints.

Therefore, the implementation of cross-device collaboration essentially requires the mobile agent system to upgrade from a single agent's perceive–reason–act loop to a system-level agent architecture with multi-agent collaboration, distributed state modeling, and mechanism design capabilities, marking the entry of mobile agent research into the "swarm intelligence" stage.

## 2.4 Efficiency: Low Inference and Decision-Making Efficiency

Although mobile agents show great potential in task automation, their bottlenecks in inference and decision-making efficiency severely constrain their practicality. This challenge stems from multiple levels: first, the reliance on large-scale models leads to high computational costs and response latency, posing high requirements for on-device deployment; second, systems generally lack long-term memory for user personalization and mechanisms to learn from past successful experiences, resulting in redundant interactions and an inability to quickly handle repetitive tasks; finally, because they cannot directly call the internal APIs of applications, agents must rely on slow and fragile graphical interface simulation, which greatly compromises both efficiency and stability.

### 2.4.1 Model Parameter Size Affects Inference Efficiency

The powerful capabilities of current mainstream mobile agents depend on their underlying pre-trained language or vision-language models with large-scale parameters, typically ranging from billions to hundreds of billions. Although larger language models possess excellent generality and expressive power, their inference costs are extremely high. Specifically, for input sequences of the same length, the model's forward computation cost grows roughly linearly with the number of parameters, as estimated in [56]. Consequently, larger models significantly increase response latency per interaction.

This enormous computational burden makes it nearly impossible to directly deploy fully functional agents on resource-constrained edge devices such as mobile phones, tablets. Consequently, most current systems are forced to adopt a cloud-based deployment solution, where interface information is sent to a server, inference is completed by powerful cloud computing resources, and the instructions are then sent back. Although this approach solves the computing power problem, it introduces three equally tricky new issues: first, the unavoidable network latency, which reduces the smoothness of interaction; second, the data privacy risks associated with uploading sensitive information like user screens to third-party servers; and third, the considerable operational costs required to maintain the cloud service.

Meanwhile, the other extreme lightweight models are not ideal solutions either. Although such models are easy to deploy and have fast response times, their limited parameter size and simplified structure cannot effectively support complex multimodal understanding capabilities. This leads to their severe inadequacy in parsing ambiguous, multi-step natural language instructions and generalizing to unseen interfaces and tasks, ultimately falling into the dilemma of being "deployable but not capable".

Therefore, exploring a viable path to migrate from large cloud-based models to lightweight on-device models has become imperative. The core of this path is not just about unilaterally applying compression techniques like distillation, quantization, and pruning, but more about finding the optimal balance (sweet spot) between model performance and resource consumption. The ultimate goal is to build an on-device model that is both "light" and "powerful"—it must meet the low-power, low-latency requirements of on-device deployment while retaining sufficient core reasoning and generalization capabilities. Achieving this goal will be the core challenge and key to enhancing the practicality and deployment flexibility of mobile agents.

### 2.4.2 Lack of Mechanisms to Leverage High-Value Historical Behaviors

An efficient mobile agent should also have the ability to summarize experience from its own historical behaviors and reuse efficient strategies [88] [50], thereby avoiding redundant "from-scratch reasoning" for repetitive tasks. Current systems generally lack this capability. Even when faced with a task that has been successfully executed multiple times, the agent tends to restart the complete perceive–reason–act loop, repeating interface analysis and task planning, which leads to redundant reasoning.

Taking "recharge 100 yuan for a mobile number" as an example, ideally, the agent should be able to recognize the consistency of this task with past records and directly invoke the operation sequence that has been verified as optimal in the past, rather than reconstructing the reasoning chain. This “no experience leveraged” behavior pattern not only wastes computational resources but also significantly increases the system’s response latency.

Therefore, establishing a generalizable mechanism for mining and reusing historical behaviors, enabling the mobile agent to achieve “experience-based rapid decision-making” when facing task repetitions or structurally similar tasks, is a key direction for improving its reasoning efficiency and user experience.

### 2.4.3 Third-Party Systems Cannot Directly Access Internal APIs

The mainstream interaction method for current mobile agents is to simulate the human “look at the screen–tap the screen” behavior path to operate with the front-end GUI. This method has strong universality, but it is essentially an indirect and fragile interaction mechanism. In contrast, directly calling the application’s function interfaces would significantly improve task execution efficiency and robustness. However, limited by the closed nature of operating systems and application ecosystems, mobile agents, as third-party systems, usually have difficulty obtaining access permissions to these internal interfaces, leading to the following problems:

- Lengthy interaction paths: For example, to get the “*order status”, an agent typically needs to complete several steps in sequence: “open APP → click ’My’ → click ’My Orders’ → locate the target order → read the status*”. However, if it could directly call an API (e.g., get_order_status(order_id)), the same task could be completed in milliseconds.
- Poor stability: UI-based interaction is highly dependent on the stability of the interface structure, and mobile interfaces are frequently updated. Any minor change in a button’s position or text could cause the operation to fail. In contrast, function call interfaces are usually published as a public contract by the system and have higher stability and backward compatibility.

In summary, how to break down the “interface barrier” between the agent and the application, and enable it to intelligently switch between “direct function call” and “UI operation simulation” modes, will be a key breakthrough direction for improving task execution efficiency and system robustness.

# 3

# Generalization: Enhancing Generalization Capability for Chinese Universal and Specialized Scenarios

The preceding chapters have provided an in-depth analysis showing that the generalization bottleneck of current mobile agents originates from a systematic mismatch between training data and the real world in three dimensions: regional and ecological differences, internal structural diversity, and behavioral authenticity. To address this core challenge, our team has designed and implemented a comprehensive data collection and annotation methodology targeting both Chinese–English universal scenarios and specialized application scenarios. We have designed and executed a complete data strategy consisting of three core stages: "Universal Capability Development", "Specialized Capability Enhancement", and "Real-Behavior Alignment". The final output of this strategy is a hierarchical, multi-source hybrid training dataset, whose overall architecture is presented in Table 3.1. To systematically elaborate on its construction process, the following three chapters will describe these key stages in sequence:

Bilingual Chinese–English Universal Data Foundation: This stage focuses on the universal data required for model training. We will describe how, by collecting open-source datasets and independently constructing Chinese universal data, we infuse the model with broad cross-lingual foundational capabilities to address the challenge of "regional and ecological differences".

Specialized Data Construction for Specialized Applications: This stage aims to collect the specialized data necessary for model training. We will present refined production methods for domain-specific agent data targeting particular application scenarios, thereby enriching the internal structure of the training data.

Collection of Real User Behavior Data: This stage is key to ensuring the data remains consistently aligned with real-world conditions. We will explain how to standardize the collection, filtering, and processing of real user behavior data, thereby fundamentally addressing the challenge of "behavioral authenticity".

**Table 3.1:** *Overall Composition of Training Data*

| Corresponding Chapter | Data Type | Language | Data Source |
|---|---|---|---|
| **1. Universal Data Construction** | Grounding Data | English | Open-source datasets |
| | Grounding Data | Chinese | Open-source datasets |
| | Agent Data | English | Open-source datasets |
| | Agent Data | Chinese | Proprietary dataset |
| **2. Specialized Data Enhancement** | Agent Data | Chinese | Proprietary dataset |

## 3.1 Construction of the Chinese–English Universal Data Foundation

To develop a mobile agent that can deeply understand and proficiently operate both Chinese and English interfaces, our model training relies on two core categories of data: Natural Language-GUI Grounding Data and Agent Task Data.

Natural Language-GUI Grounding Data is designed to establish precise correspondences between UI elements and natural language descriptions, enabling the model to "comprehend" the screen by accurately aligning visual elements with functional language. This alignment equips the model with essential visual comprehension capabilities. To achieve this goal, our dataset must not only include basic element recognition tasks but also encompass multi-dimensional visual understanding capabilities ranging from low-level to high-level, including fine-grained GUI icon classification, interface control detection, and optical character recognition (OCR) for key content. This design enables the model to develop a hierarchical perception ability akin to that of humans, progressing from "seeing clearly" (OCR) to "recognizing" (control detection) and ultimately "understanding" (icon semantics). Agent Task Data, on the other hand, is represented in the form of *(instruction, action sequence)* pairs. It aims to train the model to complete specific tasks based on natural language instructions by executing sequences of operations such as clicking, swiping, and text input—thereby enabling task-level intelligent interaction.

We posit that the model's bilingual capability is fundamentally rooted in the bilingual coverage of its training data. Accordingly, we have established a core principle: both data categories must be supported by high-quality Chinese and English datasets. To achieve this, our data construction approach is two-pronged: fully leveraging existing open-source resources while purposefully building high-quality proprietary Chinese datasets.

### 3.1.1 Integration and Analysis of Open-Source Datasets

We conducted a comprehensive and in-depth survey and integration of mainstream GUI-related datasets available in the community. First, we seek to "stand on the shoulders of giants" to rapidly build the foundational abilities of the model. Moreover, we aim to analyze existing datasets to clarify their strengths and limitations.

For both Natural Language-GUI Grounding tasks and Agent tasks, we have selected and integrated the following industry-leading open-source datasets as part of our training data preparation:

**Table 3.2:** *Datasets Related to Natural Language-GUI Grounding Capability*

| Dataset | Scale | Language | Core Task | Annotation Depth | Platform |
|---|---|---|---|---|---|
| AITZ [83] | 18k | English | Instruction-to-element mapping | Coordinates / Chain-of-Action-and-Thought | Android |
| GUICourse [8] | 700k | English | Text/region localization | Bbox / Text | Web |
| OS-Atlas [77] | 5.5M | English | General element localization | Bbox / Instruction | Cross-platform |
| SeeClick [10] | 280k | English | Visual element localization | Bbox / Instruction | Cross-platform |
| Wave-UI [65] | 16k | Chinese–English | UI understanding | Functional/intent description | Web |
| AMEX [6] | 100k | English | Multi-level element localization | Coordinates / Element function description | Android |
| UGround-V1 [23] | 6.5M | English | Visual localization | Bbox / Textual referring expressions | Cross-platform |

**Table 3.3:** *Comparative Analysis of Mainstream Agent Task Datasets*

| Dataset | Scale | Language | Task Type | Average Steps | Task Diversity | Annotation Depth |
|---|---|---|---|---|---|---|
| AITW [54] | 3M / 715k | English | Single/multi-step (single-APP) | 2-16 | Hundreds of APPs and websites | Low: OCR + icon annotation only |
| Android_Control [36] | 250k | English | Single/multi-step (single-APP) | 4.8 | Extremely high: 833 APPs, 15k+ tasks | Medium: high/low-level instructions; partial reasoning chains |
| GUI_Odyssey [42] | 90k | English | Multi-step (cross-APP) | 15.3 | High: 212 APPs, 1.4k+ combinations | High: semantic labels for decision reasoning and screen understanding |
| AITZ [83] | 13k | English | Single/multi-step (single-APP) | - | Based on AITW, covering 70+ APPs | Extremely high: complete CoAT reasoning annotation |
| AMEX [6] | 38k | English | Multi-step (single-APP) | 12.8 | High: 192 APPs, 3k+ complex instructions | Extremely high: triple-layer annotation (function, localization, instruction chain) |

Through the systematic integration of existing open-source datasets, we observe that model training in English contexts already enjoys a solid data foundation, particularly in terms of dataset scale and task diversity. However, a critical and increasingly evident issue has emerged: high-quality Chinese data remains severely lacking. Although certain datasets contain a small amount of Chinese data, in terms of overall volume, application coverage, and task complexity, such data is far from sufficient to support the development of robust generalization and reasoning capabilities in Chinese scenarios. This shortfall will directly constrain the model's generalization capability and real-world relevance in Chinese contexts.

### 3.1.2 Self-Built Universal Data Pipeline for Chinese Scenarios

To address the industry-wide shortage of high-quality Chinese Agent task data and to build the core competitiveness of our model in localized applications, we initiated a large-scale, high-standard self-built data project. This project employed a standardized data collection pipeline to obtain a vast quantity of user instructions originating from real-world scenarios.

#### Systematic Application Selection

The process began with a systematic selection of applications within the domestic APP ecosystem. We considered factors such as market penetration, user activity level, and the irreplaceability of application scenarios, ultimately constructing an application matrix consisting of more than 30 mainstream Chinese Android APPs. This matrix, covering the primary scenarios of Chinese users' digital lives (Table 3.4), includes a variety of nationally popular applications. This ensures diversity and representativeness in our dataset.

#### Task Design Based on Real Use Cases

To construct Agent task data with authentic semantics, diverse expressions, and large-scale coverage, we designed and implemented a structured task generation process based on the "instruction template and slot filling" paradigm. This process, grounded in expert knowledge modeling, enables systematic expansion of data scale while maintaining instruction authenticity and linguistic flexibility, forming one of the key pathways for building our Chinese agent data ecosystem.

**Table 3.4:** *Matrix of Mainstream APPs Covered in the Self-Built Chinese Dataset (36 APPs in Total)*

| | | | | | |
|---|---|---|---|---|---|
| Ele.me | Amap | Dianping | WeChat | Bilibili | Tencent Meeting |
| Xiaohongshu | Taobao | Browser | Alarm | Notes | Meituan |
| Ctrip | Baidu Maps | Tencent Video | Gallery | JD.com | QQ Music |
| NetEase Cloud Music | Cainiao | Kugou Music | Youku Video | Didi Chuxing | 5G Wide Vision |
| Tencent Maps | Douyin | Alipay | Kuwo Music | Feishu | Ximalaya |
| Tomato Novel | Weibo | Qunar | WeChat Reading | Meituan Waimai | iQIYI |

*Definition of Instruction Patterns and Template Library Construction* In the first stage, human experts lead the abstraction and definition of "instruction patterns" that cover the core functionalities of each APP. From a user perspective, we analyze the core usage scenarios of typical APPs and summarize several high-frequency, mission-critical task types. For each task type, we further construct multiple base instruction templates with rich linguistic diversity. These templates vary in syntactic structure, informational emphasis, and colloquial style, while reserving task parameters through "slot" placeholders. Through meticulous template design, we ensure that subsequent data maintains naturalness and diversity in linguistic expression while retaining structural consistency.

*Slot Filling for Instruction Templates* After template construction, the second stage involves users and annotators filling the slots based on real-life experience to generate complete instructions. This stage emphasizes contextual authenticity and clarity of intent, avoiding the semantic distortion caused by random word filling.

By leveraging this "expert template design + user semantic filling" mechanism, we achieve both structural and large-scale optimization of task instructions. On this basis, the platform can further apply enhancement strategies such as semantic perturbation and context reuse, ensuring comprehensive coverage across task types, expression styles, and contextual scenarios. This process not only guarantees the naturalness and diversity of linguistic expressions in task data but also greatly improves data generation efficiency and controllability.

Ultimately, through the above process, we supplemented and annotated approximately 50,000 high-quality Chinese task instructions covering multiple mainstream APP scenarios. Along with the integrated open-source datasets (see Table 3.2 and Table 3.3), these data constitute one of the core training corpora for our model.

## 3.2 Data Annotation and Expansion for Specialized Applications

For mobile agents, the quality and task relevance of data directly determine their actual performance in specific domains. In specialized industries such as telecommunications, business scenarios are often complex and diverse, encompassing a large number of domain-specific terms and standardized operational workflows. Although we have constructed a general-purpose dataset with broad coverage, such data typically falls short of meeting the stringent requirements of specialized domains for high professionalism, precision, and reliability.

Therefore, obtaining and processing specialized domain data closely tied to real-world business is not only the starting point for building high-performance mobile agents but also a critical factor determining their ultimate effectiveness. Such domain-specific data not only provides the semantic foundation for the model to understand business logic but also serves as the core guarantee for making precise decisions and executing efficiently in actual operations.

To this end, we conducted in-depth analyses of representative applications in real business contexts, systematically mapping out their core business processes and high-frequency user needs. As an illustrative case, we selected China Unicom's APP ecosystem, which is typical of telecommunications applications, and carried out structured task analysis. Based on this methodology, we organized expert teams to design task instructions and completed high-quality domain-specific data annotation work. As a result, we constructed datasets closely aligned with real business scenarios of specialized industries such as telecommunications, providing a solid foundation for building mobile agent capabilities that can be broadly applied across different application ecosystems.

### 3.2.1 Systematic Application Mapping in Representative Ecosystems

To systematically cover diverse business scenarios, we first conducted a comprehensive review and categorization of application ecosystems in specialized domains. As an example, in the case of China Unicom, we grouped its diverse applications and services into logically coherent and functionally cohesive categories (see Table 3.5), laying a solid foundation for subsequent instruction creation and data annotation work. This methodology is equally applicable to other enterprises and domains with complex service ecosystems.

**Table 3.5:** *Core Application Function Categories in the China Unicom Ecosystem (as an illustrative case)*

| Function Category | Representative Applications / Functional Examples |
|---|---|
| **Communication & Messaging** | **Unicom Video Ringtone** (incoming call video display and interactive service) |
| **Network & Broadband** | **China Unicom APP** (account management, plan subscriptions, balance inquiry), **Unicom Smart Home APP** (intelligent networking) |
| **Shopping & Cultural Tourism** | **China Unicom APP** (includes Unicom Mall for mobile phones, accessories, smart devices, and cultural tourism products) |
| **Entertainment & Content** | **5G Wide Vision** (HD video, VR content, online performances, and other immersive multimedia services) |
| **Cloud Storage & Utilities** | **Unicom Cloud Drive** (file storage, backup, and multi-terminal synchronization services) |

### 3.2.2 Instruction Determination and Augmentation for Applications

Instruction construction is the key bridge between user intent and application operations, aiming to produce a high-quality instruction set that covers diverse scenarios. So we adopted a biphasic strategy of "expert knowledge definition + LLM augmentation".

First, an expert team authored "seed instructions" with clear intent and complete elements for the core functions of representative applications, thereby forming the high-quality nucleus of the instruction set. Next, to simulate the varied linguistic habits of real users, we designed precise system prompts to strictly regulate the LLM's behavior, applying augmentation-based paraphrasing to the seed instructions. This process follows the principles below:

- Strict constraint on intent preservation: Ensure that LLM-generated instructions are 100% consistent with the core intent of the seed instruction. For example, an augmented instruction aimed at "buying a phone" must not drift into "checking phone information".
- Protection of core keywords: Key entities in the instruction, such as the application name "China Unicom APP", product name "Apple phone", or constraints like "4,000-7,000 yuan", are set as immutable "protected keywords" to prevent loss of critical task elements.
- Injection of harmless procedural phrasing: Without altering the core operation, the model may insert minor, natural, conversational flow words to improve naturalness.
- Output format control: Length and formatting are constrained to maintain conciseness and structural consistency, facilitating automated downstream processing.

By utilizing the generative abilities of LLMs while enforcing precise rule constraints in the prompts, we avoided semantic drift and information loss, ultimately producing a high-quality, wide-coverage instruction dataset. This methodology, while exemplified using China Unicom, is broadly applicable to other specialized applications requiring precise and reliable mobile agent operations.

## 3.3 Real User Behavior Data Collection Strategy

### 3.3.1 Challenges in Data Collection

Through the processes described in the preceding sections, we have constructed a rich instruction dataset covering mainstream Chinese applications as well as specialized domains. However, these data contain only the textual description of user intent—i.e., the "instruction". A complete, trainable sample requires not only the instruction but also the strictly corresponding "ground truth" user behavior trajectory, namely the *action sequence*. Efficiently and reliably acquiring these action sequences is the core challenge in transforming instruction data into usable training assets.

Two main industry approaches exist to address this challenge:

*Approach 1: LLM-based automatic annotation* This approach uses advanced multimodal large models to "understand" both the instruction and current UI screenshot, then autonomously generate the required action sequence to complete the task. While potentially enabling fully automated collection at minimal human cost, it suffers from inherent limitations:

- Lack of behavioral authenticity: The actions generated are based on the model's parameterized knowledge rather than actual human behavior. They may contain "hallucinations" or non-human-like operational habits, and cannot fully replicate real decision-making in complex scenarios.
- Uncertain accuracy and generalization: When facing unfamiliar or domain-specific complex UIs, the model's operation success rate is unreliable—especially for long-sequence, fine-grained tasks.

*Approach 2: Human-in-the-loop behavior trajectory collection* Here, trained annotators execute task instructions on real devices, with their full interaction trajectories recorded. Compared to automated generation, this approach offers clear advantages:

- Native data with high fidelity: All trajectories originate from real user operations, faithfully reflecting decision-making processes and behavioral patterns in real interactive contexts. Such "gold standard" data provide highly reliable supervision for model training.
- Diversity in task completion strategies: Different users may choose different paths to the same goal via navigation menus, search entries, or history records based on personal habits, UI prompts, or prior experience. This natural strategy diversity enriches the dataset, enabling models to generalize across paths rather than rely on a fixed procedure.
- Spatial generalization at the interaction level: Human operations inherently tolerate spatial variation. For example, in click actions, users often click anywhere within a control's active area rather than a single pixel coordinate. Thus, collected data naturally cover multi-point distributions within interactive regions, helping models build structural understanding of "clickable areas" and improving robustness to layout changes or occlusions.
- High accuracy and adaptability: Humans are adept at handling complex layouts and diverse task logic, flexibly responding to abnormal states, loading delays, and other issues yielding higher collection success rates and data quality.

In pursuit of data authenticity and accuracy, we conclude that high-quality ground truth data are the foundation for building reliable autonomous agents. Therefore, we ultimately adopt the "human behavior trajectory collection" approach, supplemented by engineering methods to address potential efficiency and standardization challenges.

### 3.3.2 Standardized Collection Tool Implementation

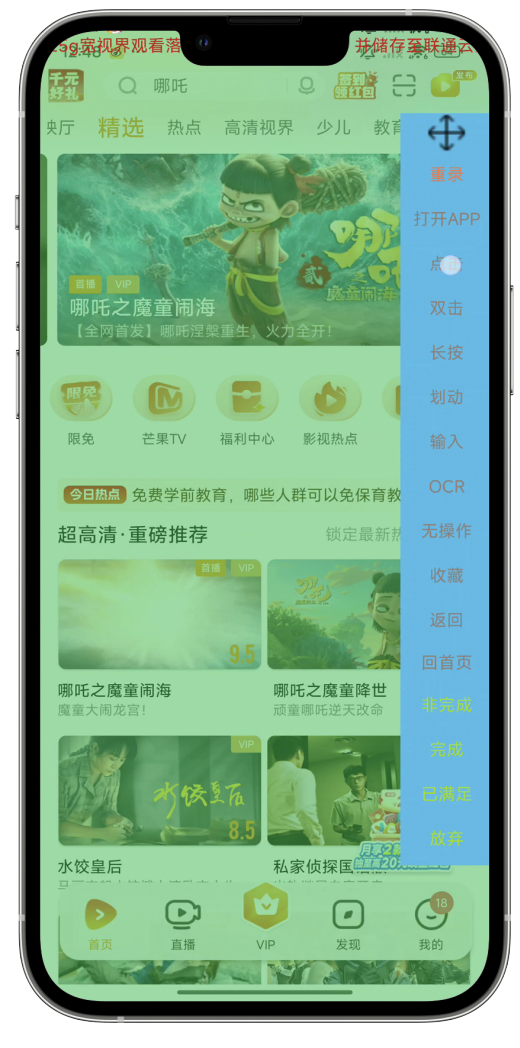

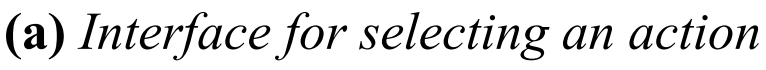
**(a)** *Interface for selecting an action*

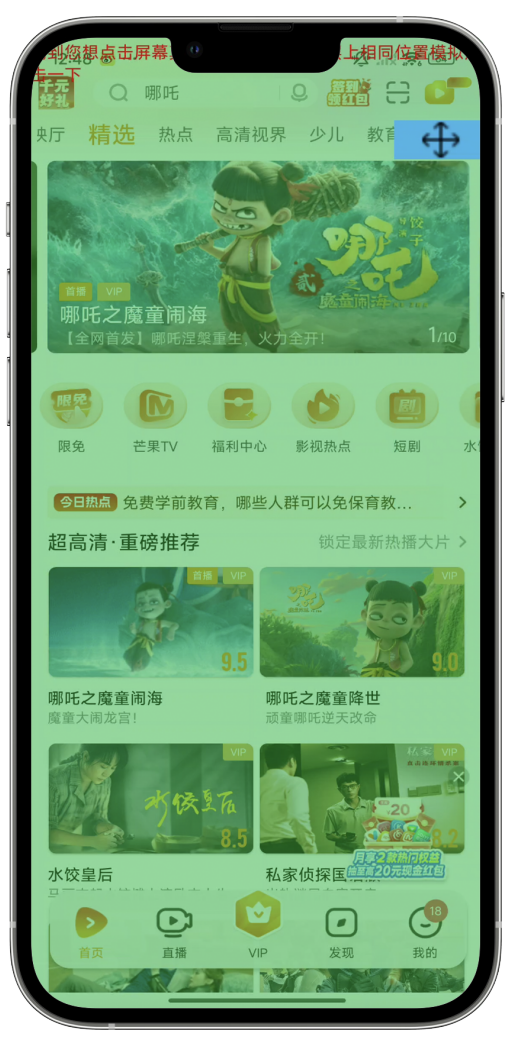

**(b)** *Interface with action selected but not executed*

**Figure 3.1:** *Data collection APK interface examples*

To scale and standardize the human annotation approach, we used a dedicated data collection APK. This tool is not a simple screen recorder but an integrated system combining task distribution, process guidance, and structured data capture. Its core design centers on an "intent-before-action" collection workflow. As shown in Figure 3.1, before performing each task step, annotators must explicitly declare the type of action they intend to execute (e.g., "click", "swipe", "type") via a floating menu on the right side of the screen. The collection tool then enters the corresponding capture mode, guiding and precisely recording the action's parameters (e.g., click coordinates, swipe trajectory, input text).

**Table 3.6:** *Atomic operations and parameter specifications of standardized action space* $\mathcal{A}$

| **Atomic Operation / Parameter** | **Description** |
|---|---|
| POINT | Locates an operation coordinate, accepting an integer tuple $(\mathrm{x},\ \mathrm{y})$ normalized to $[0,1000]$. Defaults to click; can be combined with $\mathrm{to}$ or $\mathrm{duration}$ to perform swipe or long press. |
| to | Defines swipe direction or target coordinate. Can be set to a preset direction (e.g., "$\mathrm{up}$", "$\mathrm{down}$") or combined with $\mathrm{POINT}$ to specify a swipe trajectory. |
| TYPE | Inputs text into the current focus element, accepting a string as parameter. |
| PRESS | Triggers a system-level key, such as "$\mathrm{HOME}$" (home screen), "$\mathrm{BACK}$" (back), or "$\mathrm{ENTER}$" (confirm/newline). |
| STATUS | Updates the current task execution status, such as "$\mathrm{finish}$" or "$\mathrm{impossible}$". |
| duration | Specifies the duration of the action (in milliseconds), used for waits or long presses. |

In summary, by employing our self-developed collection tool built on the "intent-before-action" workflow, we have successfully scaled the human annotation approach. This strategy not only ensures the authenticity of behavioral data but, through its mandatory intent declaration and guided parameter capture, fundamentally guarantees the standardization, consistency, and accuracy of the collected data—laying a solid and reliable foundation for subsequent model training.

# 4 Accuracy: Enhancing Single-Step Manipulation Accuracy in GUI Mode-Driven Systems

## 4.1 End-to-End Inference Architecture Based on Multimodal Large Language Models (MLLM)

In order to fundamentally cope with the prevalent issues of "error accumulation" and "optimization fragmentation" in traditional modular technology pipelines, we introduce an end-to-end inference architecture grounded in a MLLM as the core foundation of the mobile agent. This architecture abandons the "pipeline-style" design that relies on multiple independent submodules in series, and instead integrates visual perception, instruction comprehension, and action decision-making into a unified model, achieving tighter cross-modal collaboration and a more natural reasoning flow.

Our system is built upon the industry-leading open-source pretrained multimodal model AgentCPM-GUI [87], and is further fine-tuned and reinforced on downstream tasks specific to mobile agent scenarios, thereby enhancing its adaptability in interactive execution domains. This section introduces the fundamental structural design of the multimodal large model we adopt, and further explains how this architecture effectively supports the end-to-end task execution process, along with the systemic advantages it brings.

### 4.1.1 Fundamental Principles and Advantages of the MLLM

The MLLM integrates a vision encoder, a large language model core, and a vision-language merger to achieve unified modeling of heterogeneous visual-text information, providing a solid foundation for building intelligent and efficient end-to-end interactive systems.

*Vision Encoder* The vision encoder is responsible for converting raw visual inputs such as GUI screenshots into high-dimensional feature vectors, namely "visual tokens". Mainstream im-plementations are typically based on the Vision Transformer (ViT) [17], which partitions images into patches and applies a Transformer architecture to capture both local details and global layout information.

**Figure 4.1:** *Input-output schematic of the mobile agent with end-to-end architecture*

*Large Language Model Core* The large language model serves as the reasoning and decision-making center of the system. Our model is initialized from the pretrained language model AgentCPM-GUI, inheriting its powerful language understanding, logical reasoning, and knowledge generalization capabilities.

*Vision-Language Merger* Located at the interface between the vision and language modalities, this module is primarily used to compress the long sequence features output by the vision encoder, thereby reducing computational load, and to map visual features into a representation aligned with the embedding space of the language model. This mechanism improves processing efficiency while preserving the visual-semantic information crucial for downstream tasks.

Supported by the above architecture, an MLLM-based end-to-end system demonstrates three key advantages over traditional approaches. First, in terms of deep integration of heterogeneous information, an MLLM can directly process raw, pixel-level low-level visual features rather than relying on structured text generated by intermediate modules, thereby avoiding semantic loss caused by information conversion. This allows the model to capture detailed associations among appearance, color, and icons of UI elements and their functional semantics, resulting in deeper understanding of GUI interfaces. Second, the architecture possesses end-to-end global optimization capability: as a continuous, differentiable neural network, it supports backpropagation from the final task objective, enabling unified optimization of all module parameters, which significantly enhances stability and efficiency in real manipulation tasks. Finally, regarding the paradigm of capability inheritance and development, the model not only inherits the strong generalization and zero-shot reasoning abilities of the pretrained LLM, but also supports task-oriented fine-tuning to further shape professional operational strategies for mobile agents. This enables the effective unification of generality and domain-specific expertise, with the ability to continuously adapt and expand in novel interfaces and complex environments.

### 4.1.2 Mobile Agent with End-to-End Architecture

Based on the multimodal LLM, we are able to construct a truly end-to-end inference system. As shown in Figure 4.1, the core design philosophy of this architecture is to simulate the most natural human interaction logic: a person perceives the GUI interface through "vision", integrates it with the "language-based" task intent, and finally forms an action decision also expressed in language. The adopted MLLM directly maps multimodal inputs (GUI screenshot + user text instruction) to structured text outputs. This output is an explicit, language-based action command (e.g., click("Login button") or type("hello", on="Input field")), which directly corresponds to specific executable operations. Throughout this process, there are no manually designed intermediate modules or feature engineering; the entire information flow is seamlessly and naturally completed within the model.

This end-to-end "text generation" paradigm has its most significant advantage in greatly facilitating holistic model optimization. It supports backpropagation from the success or failure of the final task via a unified loss function, optimizing the full chain of parameters from visual perception to textual decision-making. Whether subtle deviations in perception or flaws in reasoning logic, both can be "captured" and corrected by the final task objective. This "result-driven model improvement" training paradigm is difficult for traditional methods to match. In addition, the architecture provides two further benefits: since the model always directly accesses the most raw and complete input information, it completely avoids the cascading amplification of errors across independent modules, significantly improving the robustness of the reasoning process; secondly, the overall system architecture is more concise, requiring no complex interface protocols or submodule logic, thereby greatly reducing development and maintenance costs.

In summary, the MLLM-driven end-to-end inference architecture not only improves accuracy and generalization at the model level but also achieves a return to human natural interaction logic in design philosophy. By building a deeply integrated chain from perception, understanding, to decision-making, it provides a solid technical foundation for creating truly intelligent and reliable mobile agents.

## 4.2 OCR- and OR-Based Region Localization Calibration Mechanism

At the action execution level of the mobile agent, many existing methods adopt pixel-level coordinates as the final output form to specify the click position. However, excessively fine-grained localization makes the model extremely sensitive to minor errors, often resulting in manipulation failures. To mitigate this issue, we introduce a control semantic-assisted localization mechanism based on OCR and OR. This mechanism utilizes the OCR+OR module to identify semantic labels and bounding box regions of interface elements, which are then used to geometrically calibrate the model's output coordinates.

### 4.2.1 Widget Region Recognition and Semantic Parsing

A GUI serves as the central hub connecting users with the functional capabilities of an application. Through graphical elements such as buttons, icons, and text, it fulfills the dual role of information presentation and interaction control. Fundamentally, a GUI functions as a visual language, conveying operational intent and system state via the spatial layout of widgets, visual style, and textual content. Consequently, the task of widget region recognition and semantic parsing in GUIs must go beyond the mere detection of visual objects and recognition of textual content; it must also achieve a structured understanding of widget semantics and interaction intentions, enabling machines to emulate human-like GUI comprehension capabilities [19][13].

Traditional computer vision approaches often focus on "detection" and "recognition", e.g., locating a bounding rectangle in the interface or identifying the text contained within it. However, the real challenge lies in integrating these raw perceptual results into high-level semantic structures with functional meaning—bridging the "semantic gap" between visual pixels and interaction functionalities.

Although OCR and generic OR technologies have matured and achieved wide success in practical applications, their isolated use still exhibits significant limitations in the context of complex GUI understanding.

*Semantic Blind Spots of OCR* The primary goal of OCR is to convert text within images into structured string information. The conventional OCR pipeline consists of image preprocessing, text detection, character recognition, and result integration. While effective in tasks such as document digitization, this approach shows two notable shortcomings in GUIs with complex layouts and semantically interdependent widgets [19]:

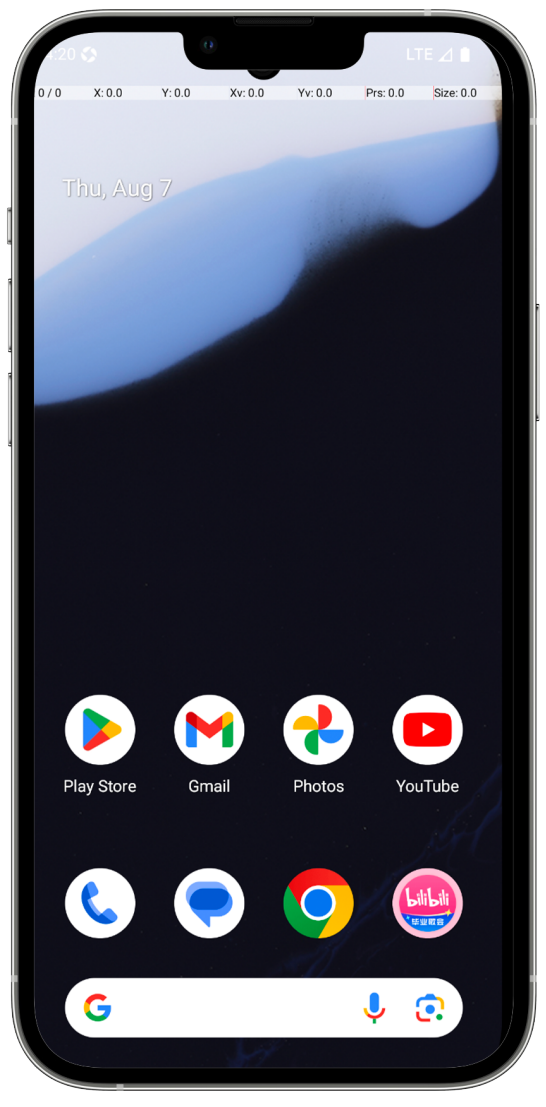

**(a)** *Original interface image*

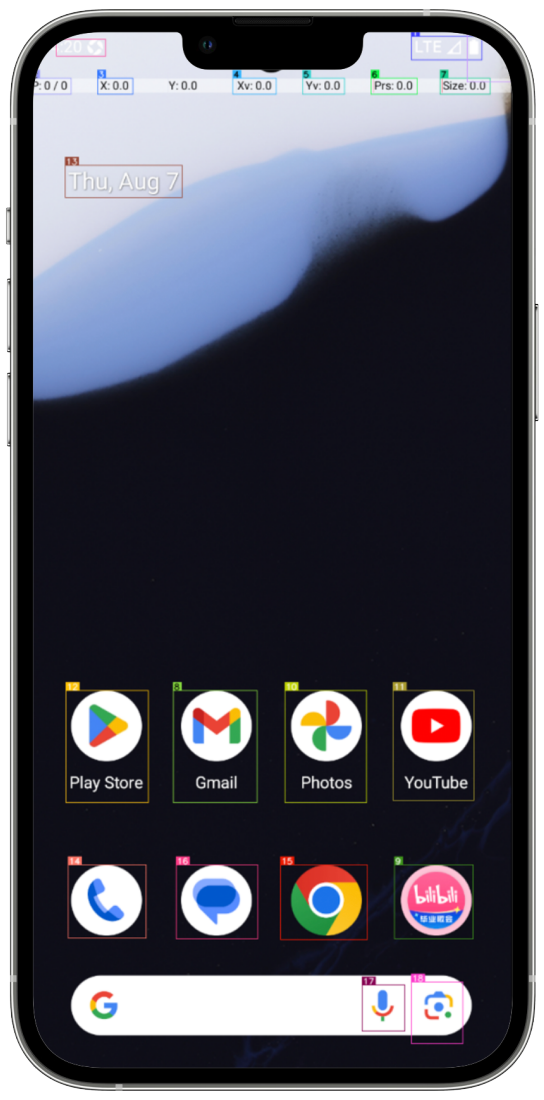

**(b)** *Visualization of OCR + OR recognition results*

**Figure 4.2:** *Example of widget region recognition: from raw interface to structured semantic region extraction*

**Table 4.1:** *Example of structured widget information output by the OCR + OR module*

| Widget Index | Widget Type | Content | Interactivity | Normalized Position (bbox) |
|---|---|---|---|---|
| 11 | Icon (Application) | YouTube | Yes | [0.74, 0.63, 0.90, 0.73] |
| 12 | Icon (Application) | Play Store | Yes | [0.10, 0.63, 0.26, 0.73] |
| 13 | Text | Thu, Aug 7 | No | [0.12, 0.14, 0.18, 0.17] |
| | | … | | |
| | | … | | |

First, lack of layout and structural awareness. GUI semantics are not only embedded in textual content but also in the spatial relationships and arrangements of widgets. For example, a "Username" label positioned next to an input box forms a logical unit, yet OCR merely extracts the text without capturing this semantic grouping.

Second, lack of interaction-function understanding. While OCR can recognize the word "Submit", it cannot determine whether the button containing it is interactive, nor can it infer the action triggered by clicking it. Therefore, OCR inherently lacks the ability to model semantic functions.

*Generalization Bottleneck of OR* OR focuses on extracting bounding box information for specific object categories from images. As deep learning models like YOLO and Faster R-CNN have progressed, widget detection accuracy has improved considerably. Nevertheless, in the context of GUIs, OR still struggles with weak recognition of semantic specificity. For instance, an OR model may classify multiple widgets as "Button" but cannot distinguish their functional differences, such as "Save", "Cancel", and "Delete", which may share similar visual styles but differ entirely in purpose. This inability to capture instance-level semantic distinctions limits the support such models can provide for higher-level interaction decisions [7][13].

*Collaborative OCR + OR Parsing Paradigm* Given the above, it is necessary to deeply integrate OCR and OR techniques into a GUI parsing framework that possesses both structural awareness and semantic understanding capabilities. The aim of this collaborative paradigm is not only to identify "what elements exist" but also to understand "what they mean", "whether they are interactive", and "what consequences interaction would produce".

We adopt Microsoft's open-source OmniParser model [43], which exemplifies this collaborative design philosophy. OmniParser combines object detection and text recognition with semantic description generation, thereby enabling an end-to-end pipeline from visual perception to semantic modeling. Its core idea is to "tokenize" a GUI screenshot into a structured, semantic, DOM-like representation suitable for downstream reasoning and decision-making by language models. The framework comprises three parallel modules:

- Interactable Region Detection: A fine-tuned YOLOv8 model detects bounding boxes for interactive widgets such as buttons, icons, and text boxes. This module answers the fundamental questions: "What interactive elements are present on the interface?" and "Where are they located?"
- Semantic Description Generation: A pretrained vision-language model generates image-to-text descriptions for widget regions, including not only visual attributes but also anticipated functional roles. For example, a gear icon is described as "Settings menu entry" rather than simply "gear icon".
- OCR: The OCR module extracts all textual information from the interface, including labels, menus, and tooltips providing linguistic cues for constructing a semantic structure.

OmniParser's output preserves the spatial layout and visual characteristics of interface elements while integrating textual semantics and interaction properties. This structured output facilitates semantic understanding and action planning. As shown in Figure 4.2, in a typical smartphone interface, OmniParser can identify semantically meaningful widgets such as "Gmail" and "YouTube" icons, along with their precise locations and interactivity labels. The structured output, illustrated in Table 4.1, specifies widget type, semantic content, interactivity, and normalized spatial position (bbox). This representation provides stable and precise target regions for subsequent action coordinate calibration, mitigating uncertainty inherent in pixel-level targeting.

### 4.2.2 Action Coordinate Calibration Based on Widget Regions

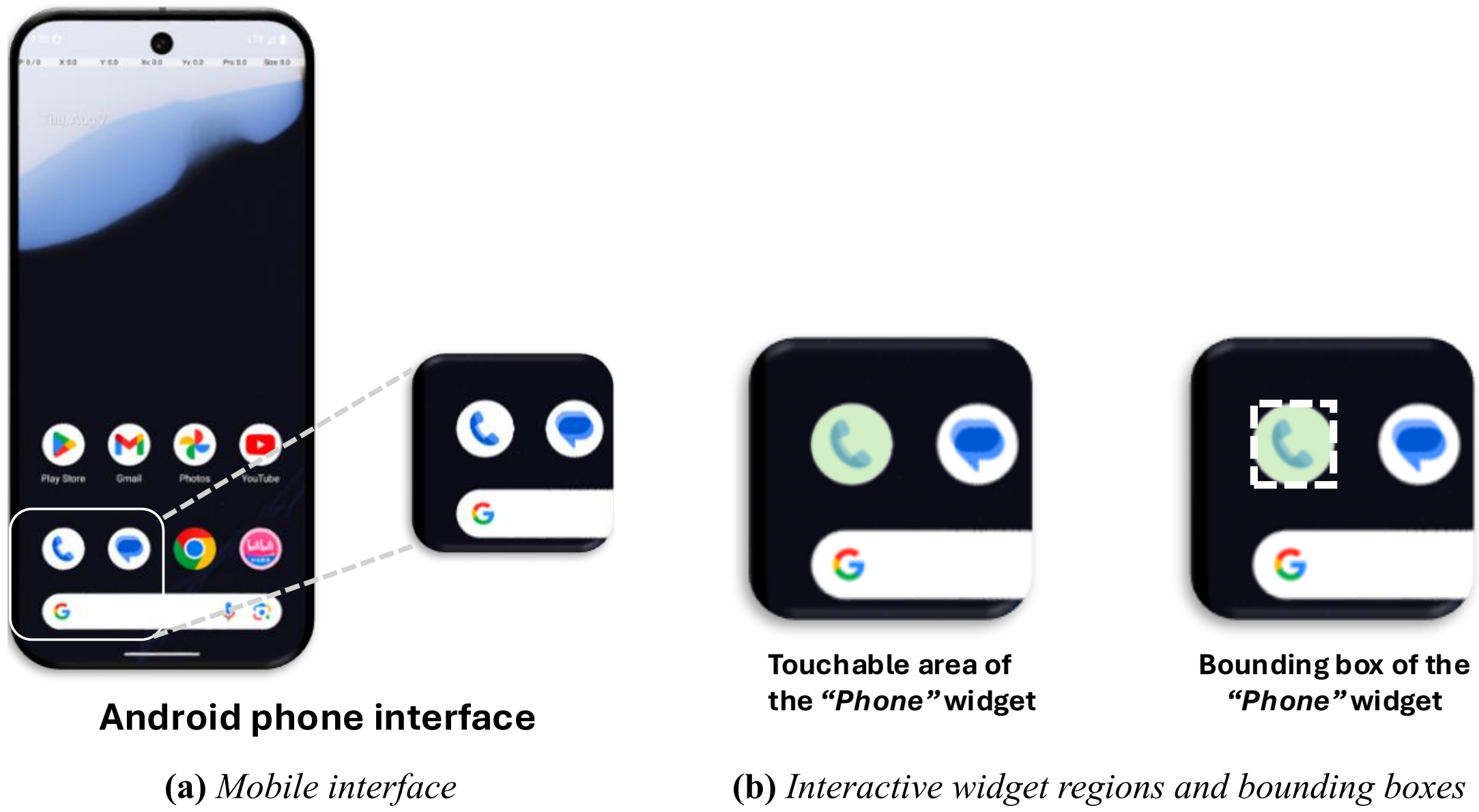

**(a)** *Mobile interface* **(b)** *Interactive widget regions and bounding boxes*

**Figure 4.3:** *Example of interactive regions and bounding boxes corresponding to widgets*

To improve the interaction accuracy and robustness of mobile agents operating in complex visual environments, we design an action coordinate calibration strategy that combines static spatial constraints with dynamic historical memory mechanisms, leveraging the spatial and semantic information of widget regions.

Each interactive widget is associated with a bounding box (bbox) that approximately delineates its operational area (Figure 4.3).

Static Bounding Box Constraint Correction: For a model-generated click point $p = (x, y)$, we first check whether it falls within any widget bounding box in the set $\mathcal{B} = \{b_1, b_2, \ldots, b_n\}$:

$$\exists b_i \in \mathcal{B}, \quad p \in b_i$$

If $p$ is inside a widget region, the click is considered valid and the original coordinate is retained. Otherwise, we treat it as a potential misalignment requiring correction. We compute the minimum perpendicular distance from $p$ to each bounding box $b_i$ and relocate $p$ to the center of the closest widget $b^*$:

$$b^* = \text{argmin}_{b_i \in \mathcal{B}}\, d(p, b_i), \quad p' = \text{center}(b^*)$$

where $d(p, b_i)$ is the minimum Euclidean distance from $p$ to rectangle $b_i$:

$$d(p, b_i) = \sqrt{\max(x_1 - x, 0, x - x_2)^2 + \max(y_1 - y, 0, y - y_2)^2}$$

with $b_i = [x_1, y_1, x_2, y_2]$ representing bbox coordinates (top-left and bottom-right corners).

The rationale is to leverage detected widget bounding boxes to monitor and correct mobile agent click commands in real time, ensuring operational validity and preventing failures due to coordinate drift. This is grounded in the behavioral assumption that a user's intention is typically to interact with a specific widget on the interface; hence, a valid click should fall within an interactive region. When a click lies outside all recognized widget regions, it is reasonable to infer a targeting error and adjust it to the nearest widget center.

Dynamic Historical Correction Avoidance: For each frame $I$, the system computes a perceptual hash $h(I)$ and records the set of previously corrected click coordinates that failed to produce interaction $\mathcal{F}(h(I))$. Before applying a correction, if:

$$p' \in \mathcal{F}(h(I))$$

then the system bypasses the correction and executes the original $p$, avoiding repeated erroneous adjustments. This mechanism is based on the assumption that a valid interaction should cause a state change in the interface; if a click at a given location has historically failed to elicit any change, it is deemed ineffective, and repeating it should be avoided to prevent infinite loops.

## 4.3 Multi-Agent Collaborative Decision-Making Strategy

Multi-agent voting among large language model (LLM) agents is a key ensemble decision-making strategy [33][29]. Its central premise is that aggregating the "opinions" of multiple autonomous LLM agents can produce a decision that is more accurate and robust than any individual model. This approach mitigates hallucinations and biases common to single models, while leveraging "collective intelligence" to address complex reasoning and decision-making tasks.

Various forms of implementation exist, ranging from basic majority voting [33], to weighted voting based on historical performance or confidence scores [51], to ranked-choice voting capable of expressing complete preference orderings [89].

In our system, we employ a multi-agent ensemble strategy with majority voting to enhance decision-making robustness. At each decision step, multiple independent agents receive the exact same environmental state and independently propose an action. The system then aggregates these proposals via voting to select a consensus-based action for execution.

Furthermore, in our collaborative framework, each decision is decomposed into two stages to ensure consensus from "what to do" to "how to do it":

*Stage 1: Action Type Voting* The system collects all proposed action outputs and conducts majority voting on their action type. Action types include: POINT (click coordinate), PRESS (press a key), TYPE (enter text), STATUS (status check), CLEAR (clear field), etc. The type with the highest vote count becomes the primary action type for that step.

*Stage 2: Action Parameter Aggregation*

To consolidate outputs from parallel decision-making agents into a unified and effective instruction, we design a parameter aggregation strategy tailored to the data characteristics of each action type (Table 4.2). Instead of a one-size-fits-all majority rule, the method applies different voting or fusion mechanisms: for continuous spatial data like $(x, y)$ coordinates, geometric centroid voting identifies a consensus location; for discrete actions like PRESS or candidate texts, frequency counting selects the most common choice; and for the decision rationale (thought), we preserve coherence by tracing back to the originating agent. This differentiated aggregation scheme condenses collective agent intelligence into a single, logically consistent action, thereby enhancing decision accuracy and interpretability.

**Table 4.2:** *Multi-agent action aggregation strategy*

| **Action Type** | **Parameter Aggregation Strategy** |
|---|---|
| (x,y) coordinates | Geometric centroid voting: compute the centroid of all coordinates, select the closest candidate |
| PRESS/STATUS/CLEAR | Discrete count voting: count occurrences, choose the most frequent value |
| text | Text frequency voting: count suggested texts, choose the most common |
| duration | Nearest-to-mean selection: compute the mean, choose the original duration closest to it |
| thought | Source tracing: adopt the reasoning chain from the agent whose action was ultimately selected |

Through this mechanism, we build a careful and robust decision-making system that synthesizes multiple agents' inputs while fine-tuning parameter handling, significantly improving overall task execution success rates.

# 5

# Long-Horizon Capability: Enhanced Planning through Task Decomposition

## 5.1 Supervised Reinforcement Algorithms for Sequential Reasoning

The interaction between mobile agents and their environment is inherently a multi-turn, sequential decision-making process. In such processes, the ultimate success or failure does not depend on any single correct action itself, but rather on the logical coherence and cumulative effectiveness of the entire action sequence. However, as discussed previously, traditional training paradigms exhibit limitations in addressing certain core challenges, particularly in credit assignment over time and modeling long-horizon causal dependencies.

To overcome these limitations, we adopt the training approach of Reinforcement Learning (RL). Specifically, instead of providing the model with a "ground truth" answer for each step, we allow it to autonomously decide on the action to take for each step, thereby generating complete action sequences. Then, the system provides a fine-grained reward signal based on the overall task outcome. Therefore, by iteratively optimizing its policy, the model is compelled to learn the causal chain between single-step behaviors and final outcomes. Consequently, the learning objective naturally shifts from achieving "local optima" to attaining "global optima".

In practice, we use the Group Relative Policy Optimization (GRPO) algorithm to conduct Reinforcement Finetuning (RFT). GRPO [58] is a simplified and improved variant of the classical Proximal Policy Optimization (PPO) algorithm [57]. It replaces the complex value function network in PPO with relative comparisons among a group of candidate generations. This modification reduces architectural complexity while improving gradient stability and efficiency with direct relative reward signals.

This algorithm proceeds in four core steps:

1. Sampling: Given an input query $q$, the algorithm starts by using the old version of the current policy $\pi_{\theta_{old}}$ for sampling. Generated candidate responses are denoted as the set $\{o_1, o_2, \ldots, o_N\}$.
2. Evaluation: For each generated response $o_i$, One or more external reward functions are applied to evaluate it, which produces a scalar task reward $r_i$. Thus, we obtain the reward set corresponding to the response set: $\mathbf{r} = \{r_1, r_2, \ldots, r_N\}$.
3. Advantage Calculation: The GRPO algorithm does not rely on a learned value network to estimate the expected return of a state. Instead, it normalizes the reward of each response within its group to compute its relative advantage. Specifically, the advantage value $\hat{A}_i$ of response $o_i$ is obtained by calculating the Z-score of its reward:

$$\hat{A}_i = \frac{r_i - \text{mean}(\mathbf{r})}{\text{std}(\mathbf{r}) + \epsilon} \tag{5.1}$$

   where mean($\mathbf{r}$) and std($\mathbf{r}$) represent the mean and standard deviation of the reward set $\mathbf{r}$, and $\epsilon$ is a small constant introduced for numerical stability. The advantage value $\hat{A}_i$ directly quantifies the relative quality of response $o_i$ compared to the average level generated by the current policy.
4. Policy Update: After computing the advantages, the algorithm updates the policy network using an objective similar to PPO. This objective includes a Clipped Surrogate Objective and a KL Divergence penalty. The goal of the policy update is to maximize the following function $J_{\text{GRPO}}(\theta)$:

$$J_{\text{GRPO}}(\theta) = \mathbb{E}_{q \sim P(Q), \{o_i\} \sim \pi_{\theta_{old}}} \left[ \sum_{i=1}^{N} \mathcal{L}_{\text{CLIP}}(\theta, o_i) - \beta D_{\text{KL}}(\pi_{\theta_{old}} \parallel \pi_\theta) \right] \tag{5.2}$$

   where $\mathcal{L}_{\text{CLIP}}$ is the clipped surrogate objective function, defined as:

$$\mathcal{L}_{\text{CLIP}}(\theta, o_i) = \min \left( \rho_i(\theta) \hat{A}_i, \quad \text{clip}(\rho_i(\theta), 1 - \varepsilon, 1 + \varepsilon) \hat{A}_i \right) \tag{5.3}$$

   Here, $\rho_i(\theta) = \frac{\pi_\theta(o_i|q)}{\pi_{\theta_{\text{old}}}(o_i|q)}$ denotes the probability ratio between the new and old policies for response $o_i$. $D_{\text{KL}}(\pi_{\theta_{old}} \parallel \pi_\theta)$ represents the KL divergence between the old and new policies, which measures the magnitude of the policy update. The hyperparameter $\beta$ controls the penalty strength to ensure the stability of the training process.

In summary, we adopt the GRPO algorithm for RFT of the model to address the long-horizon dependencies and credit assignment problems faced by single-step supervised learning in sequential tasks. This end-to-end reinforcement learning paradigm enables the agent to better establish causal relationships between individual actions and final outcomes, thereby enhancing its strategic planning capability and overall task completion rate in complex GUI tasks.

## 5.2 Long-Horizon Task Planning and Decomposition Mechanism

Many real-world user tasks often require execution across different operating environments, which can be categorized into two forms: cross-application collaboration, i.e., coordinating multiple applications on a single device, and cross-device collaboration, i.e., performing cooperative operations across heterogeneous devices such as smartphones and computers.

Although these two scenarios differ in their manifestation, they share a common requirement for task planning and decomposition: a high-level, complex user intent must be systematically planned into a logically coherent workflow composed of multiple atomic operations. Each operation within the workflow must be precisely mapped to a specific “execution endpoint” (i.e., a concrete application or physical device) for completion.

Accordingly, we propose a unified hierarchical planning and execution framework. This framework involves a central planning agent, $\pi_{\text{central}}$, to decompose and allocate tasks to different execution endpoints (applications or devices). Furthermore, it incorporates an efficient state synchronization mechanism to ensure that all subtasks are executed accurately and sequentially according to their dependency structure.

### 5.2.1 Task decomposition planning mechanism

To transform ambiguous natural language commands into structured tasks that machines can understand and execute, we design a task decomposition and planning mechanism led by a High-level Planning Agent $\pi_{\text{plan}}$. This agent takes a complex task command $T$ as input, then deeply parses user intent and ultimately outputs a structured Task Dependency Graph $G_T$, laying a solid foundation for subsequent task allocation and execution.

Specifically, when $\pi_{\text{plan}}$ receives a command, it does not adopt a traditional linear decomposition approach, as this fails to effectively capture parallel relationships among subtasks. Instead, we innovatively represent the task as a Directed Acyclic Graph (DAG) to more precisely capture its internal structure. The task dependency graph $G_T$ is defined as:

$$G_T = (V, E), \quad V = \{st_1, st_2, \ldots, st_n\}, \quad E \subseteq V \times V \tag{5.4}$$

Here, $V$ is the set of $n$ nodes corresponding to sub-tasks, where each sub-task $st_i$ is a fine-grained operation executable independently on a certain endpoint. The directed edge set $E$ describes the execution dependencies among subtasks. An edge from $st_i$ to $st_j$, $(st_i, st_j) \in E$, specifies that the completion of $st_i$ is a prerequisite for the execution of $st_j$, formally:

$$(st_i, st_j) \in E \Rightarrow st_i \prec st_j \tag{5.5}$$

This graph-based dependency modeling provides a rigorous mathematical foundation for handling complex task structures. Unlike linear sequences, DAGs naturally represent both *parallel* and *sequential* relationships, thereby enabling parallel scheduling and improved execution efficiency.

The mechanism proceeds through three major stages: task decomposition, task allocation,

and task state synchronization, as described below.

Task Decomposition: When the system receives a complex task command $T$, the high-level planning agent $\pi_{\text{plan}}$, powered by the LLM, first parses and decomposes it.

Task Allocation: In parallel with constructing the task dependency graph, the central planning agent must determine the execution endpoint for each subtask. The proposed framework is general: an endpoint may be either a physical device or a specific application on a device.

Task State Synchronization: To ensure correct execution of cross-device tasks, we design a synchronization protocol: each subtask $st_i$ is assigned a state $S(st_i)$ from defined state sets (Pending, Running, Completed, Failed) and any subtask assigned to device $d_j$ can only start if all its predecessors have been completed:

$$\forall st_p \in \text{Pred}(st_i), \quad S(st_p) = \text{Completed} \tag{5.6}$$

where $\text{Pred}(st_i)$ denotes the set of all predecessor nodes pointing to $st_i$ in $G_T$.

Through this architecture, both complex cross-device collaboration tasks and long-horizon cross-application tasks can be decomposed into goal-oriented subtask sequences executed by low-level execution strategies on individual devices. At the same time, the state synchronization mechanism ensures orderliness and reliability in cross-device execution, thereby guaranteeing the overall task completion.

## 5.3 Cross-Application Task Transition Mechanism

**Table 5.1:** *Types and Core Dimensions of Cross-Application Tasks*

| **Task Type** | **Requires Memory Maintenance** | **Proactive APP Switching** | **Linear Control Flow** |
|---|---|---|---|
| Passively triggered system linkage tasks | ✗ | ✗ | ✓ |
| Proactively triggered data-passing tasks | ✓ | ✓ | ✓ |
| Proactively triggered collaborative multi-tasks | ✓ | ✓ | ✗ |

We define a cross-application task as one where, in order to accomplish a user's final intent, the action sequence must actively or passively invoke, interact with, or reference the functionalities or data of two or more independent applications.

Cross-application tasks introduce unique challenges absent in single-application tasks, which can be deconstructed along several key dimensions:

- Memory Maintenance: Whether the agent needs to actively maintain internal memory across applications during execution, e.g., temporarily storing text fragments, screenshots, or other intermediate results.
- Triggering Mechanism: Whether cross-application switching is explicitly performed by the agent through multiple atomic operations (proactive triggering), or automatically executed by the system based on a single action (passive triggering).

- Control Flow: Whether the task execution process is linear and sequential, or non-linear, which requires concurrency or alternating handling.

Based on the above three dimensions and their combinations, we categorize the wide variety of cross-application tasks into some representative types. These types essentially form a progressive ladder of difficulty, imposing increasing demands on the agent's planning, perception, and execution capabilities. The classifications and corresponding solutions are described as follows:

*Passively Triggered System Linkage Tasks* This category of tasks, in our dimensional analysis, is characterized by no need for memory maintenance, passively triggered APP transitions, or any linear control flow. The agent only needs to perform a single atomic action such as clicking a link or button, after which the operating system's built-in deep linking mechanism automatically handles the application switch and context transfer. The agent itself does not need to proactively retain information or plan complex navigation steps. For example, when clicking the "Pay with Alipay" button in an e-commerce APP, the system automatically transitions to the Alipay APP for verification.

The main challenge here lies in accurately recognizing the key "trigger" UI element. Our solution focuses on enhancing the agent's perception and recognition ability. Specifically, we augmented the training data with a large number of cross-application linkage samples. This targeted fine-tuning enables the agent to reliably identify the correct switching button at the right time, effectively simplifying complex system-level interactions into a single successful recognition and click.

*Proactively triggered data-passing tasks* These tasks require memory maintenance, proactive application switching, and linear control flow. The agent must retrieve information (e.g., text, images, files) from one APP, store it temporarily in memory, and then perform a sequence of operations (e.g., switching APPs, locating an input field, pasting) to transfer the information to another APP.

Such tasks demand explicit step-by-step planning and memory capabilities, making them core application scenarios of the task decomposition and planning mechanism described earlier. In this context, the high-level planning agent $\pi_{\text{plan}}$ decomposes the user's intent into a linear task graph $G_T$ consisting of multiple atomic operations. This structured task graph serves as an execution blueprint, precisely guiding the agent through a sequence of actions while transferring necessary data across steps.

*Proactively triggered collaborative multi-tasks* This is the most complex type of all. It is characterized by extensive memory maintenance, proactive application switching, and non-linear control flow. The agent must simultaneously track the states of two or more APPs and dynamically perform operations in one APP based on information from another. The process often requires repeated switching and cross-referencing across applications. For example, in split-screen mode, a user may follow a tutorial in a video APP while simultaneously operating and recording in a notes APP.

Due to their high non-linearity, dynamic perception demands, and complex decision logic, these tasks often involve deep human-agent collaboration. They currently lie beyond the scope of our present work and are regarded as an important direction for future research.

## 5.4 Cross-Device Collaboration Mechanism

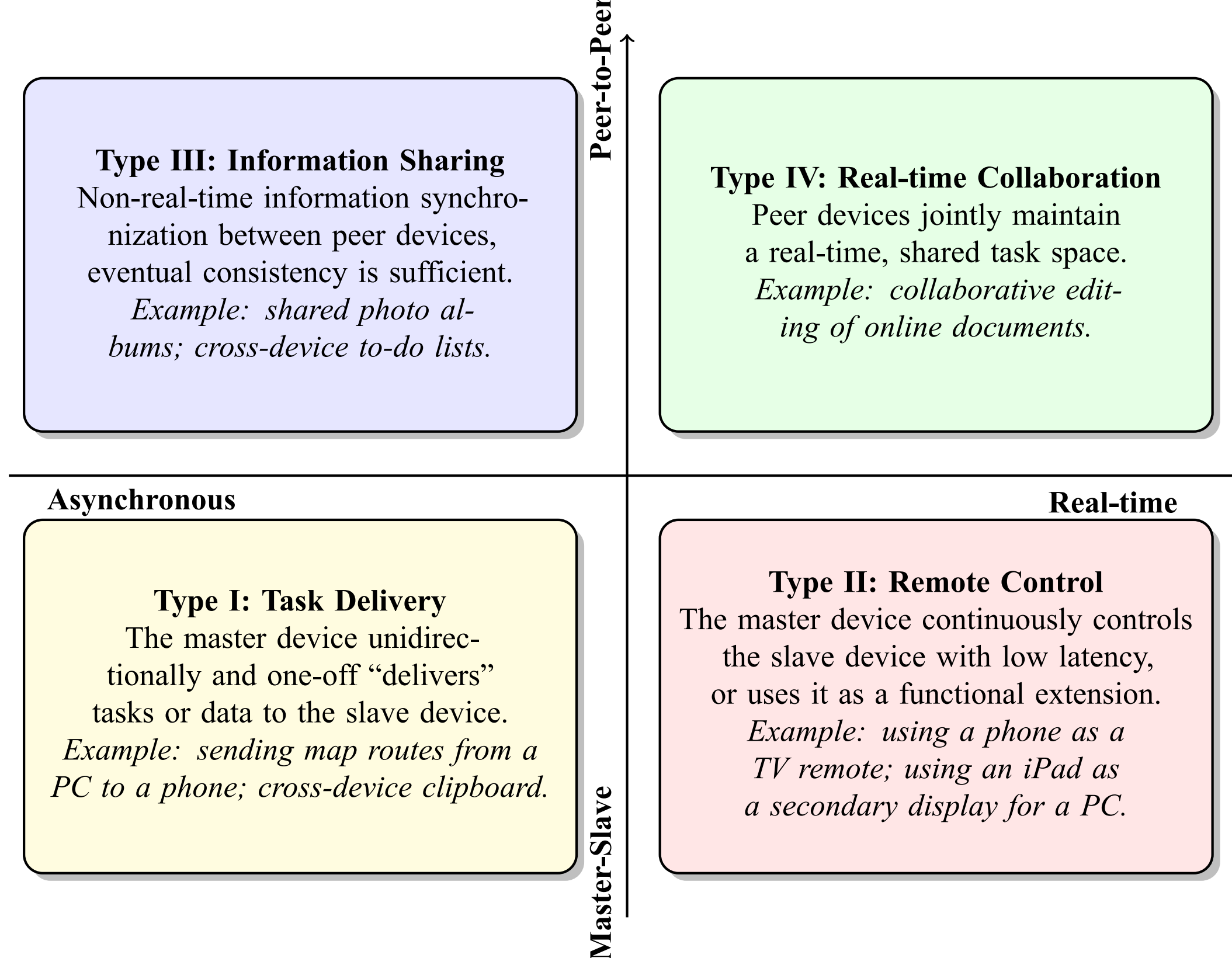

**Figure 5.1:** *Cross-device task classification quadrant chart based on synchronization mode and role relationship*

We define a cross-device task as one where, in order to accomplish a user's final intent, the execution path must be jointly carried out by two or more physically distinct computing devices (e.g., smartphone, computer, or devices belonging to different users), and task states or data must be synchronized across these endpoints. The essence of "cross-device" lies in inter-device collaboration and data exchange, not merely the independent execution of the same functionality on multiple devices.

Similar to cross-application tasks, cross-device tasks face inherent challenges due to contextual isolation and state differences among physical devices. We analyze these challenges along two key dimensions:

- Data & State Synchronicity: Whether data exchange between devices requires immediate response via real-time synchronization, or whether delayed, non-blocking asynchronous synchronization is acceptable.
- Role Relationship: Whether devices operate under a master-slave paradigm with asymmetric functionality, or in a peer-to-peer mode with functional parity.

By combining these two dimensions, all cross-device tasks can be mapped clearly into the four quadrants shown in Figure 5.1.

Regardless of whether tasks fall into the "task delivery" (Type I) in master-slave mode or "information sharing" (Type III) in peer-to-peer mode, they can essentially be decomposed into a set of well-defined, bounded subtasks. The previously described task decomposition framework has been proven to effectively manage subtask dependencies and execution flows across devices, ensuring successful task completion.

However, for tasks that require real-time synchronization (Types II and IV), the technical challenge shifts from the macro-level "task planning" to the micro-level "system engineering". For example, bottlenecks in "remote control" tasks lie in establishing low-latency, high-throughput event streams rather than generating semantic steps. Similarly, "real-time collaboration" tasks demand complex distributed consistency protocols to resolve conflicts in concurrent modifications. Addressing these problems requires a fundamentally different communication and state management infrastructure than our current framework. Consequently, our present work focuses on asynchronous collaboration scenarios, leaving real-time synchronization tasks as an important direction for future research.

# 6

# Efficiency: Three-layer Optimization of Reasoning Efficiency across Model Data Patterns

## 6.1 Efficient Model Selection for Edge Deployment

**Table 6.1:** *Comparison of large language models with different parameter sizes for GUI scenarios*

| Model | Developer | Open-source | Parameters (B) | Multimodal Input Support |
|---|---|---|---|---|
| GPT-4o | OpenAI | ✗ | N/A | ✓ |
| Claude 3.5 Sonnet | Anthropic | ✗ | N/A | ✓ |
| Gemini 2.5 Pro | Google | ✗ | N/A | ✓ |
| Qwen2-VL-72B | Alibaba | ✓ | 72B | ✓ |
| CogAgent | THUDM | ✓ | 18B | ✓ |
| Llama 4 | Meta | ✓ | 17B | ✓ |
| Ferret-UI 2 | Apple | ✓ | 8.4B | ✓ |
| AgentCPM-GUI | OpenBMB | ✓ | 8B | ✓ |
| Qwen2.5-VL-7B | Alibaba | ✓ | 7B | ✓ |
| Phi-3-vision | Microsoft | ✓ | 4.2B | ✓ |
| Llama 3-3B | Meta | ✓ | 3B | ✗ |
| Gemma 3-1B | Google | ✓ | 1B | ✗ |

Current powerful mobile agents generally rely on large-scale pretrained models, but this brings high inference cost and significant response latency. Such computational burdens make it difficult to deploy these models directly on resource-constrained edge devices such as smartphones. Meanwhile, existing cloud-based solutions have another set of problems, including network latency, data privacy concerns, and operational costs. Therefore, exploring a technical path for migrating from cloud-based large models to efficient and lightweight edge-side models has become critical to the advancement of this field.

Between the ultimate vision of fully on-device deployment and the substantial overhead of current cloud-based large models, we must identify a balance between model capability and inference efficiency. To this end, after careful evaluation, we selected a state-of-the-art open-source model with 8B (8 billion) parameters as the backbone of our mobile agent.

This was because a 8B-scale model occupies a strategic "sweet spot", as shown in Table 6.1. On the one hand, compared with giant models with tens or hundreds of billions of parameters, a 8B open-source model significantly reduces inference cost and latency, enabling the construction of an economical and efficient on device service. On the other hand, compared with smaller models (e.g., 1B-3B scale), the 8B model crosses the critical threshold of complex reasoning and multimodal capabilities. Smaller models often struggle to reliably execute multi-step complex instructions, whereas the 8B model demonstrates sufficiently strong logical planning and tool-use abilities, making it a solid backbone for a reliable mobile agent.

More importantly, the 8B-scale model is at a critical inflection point. It paves the way toward fully on-device deployment. It already has the potential for local deployment on high-end edge devices, such as flagship smartphones and laptops. This enables our research not only to optimize current cloud-based services but also to provide a practical technical route toward realizing a fully on-device mobile agent with no reliance on network connectivity, enhanced privacy protection, and zero latency.

In summary, adopting the 8B model as the backbone represents the best practice under current technological conditions, balancing core model capabilities with deployment and operational costs. This approach not only allows us to deliver high-performance, low-latency agent services, but also constitutes a forward-looking strategic layout for achieving fully autonomous on-device intelligence in the future.

## 6.2 Efficient Reuse of Historical Behavioral Data

To optimize the performance of agents in executing repetitive operational tasks, we propose a novel experience reuse framework. This framework systematically accumulates historical behavioral data and leverages semantic retrieval based on LLMs to enable precise replay of previously executed tasks. The framework aims to significantly reduce redundant reasoning overhead and execution latency in long-horizon or high-frequency tasks. Once a matching experience is successfully retrieved, the system bypasses complex planning and perception modules, entering a "zero-reasoning" execution mode, thereby enhancing efficiency and stability.

### 6.2.1 Structured Accumulation of Experience Data

The foundation of experience reuse lies in the systematic and structured recording of historical task execution processes. We design an automated logging and accumulation mechanism to provide high-quality data sources for subsequent experience retrieval.

Each independent task execution is regarded as a potential experience. The system automatically captures and records the complete execution trajectory. A structured experience entry $E$ can be formalized as a tuple:

$$E = (q, S, A, \mathcal{T})$$

where: $q$ denotes the original user query, expressed as a natural language string. $S$ represents the final execution status, with $S \in \{\text{success}, \text{failure}, \text{in-progress}\}$. $A$ is an ordered action sequence, $A = \langle a_1, a_2, \ldots, a_n \rangle$, where each atomic action $a_i$ is itself a structured entry containing the UI screenshot and the executed operation (e.g., click, input). $\mathcal{T}$ includes task metadata such as start/end timestamps and execution duration.

Only when a task is successfully completed ($S = \text{success}$) is the record $E$ deemed valuable for reuse and archived into the experience pool $\mathcal{P}$. The pool $\mathcal{P} = \{E_1, E_2, \ldots, E_m\}$ grows dynamically as more tasks are executed, ensuring continuous evolution and expansion of coverage. This accumulation mechanism provides a robust data foundation for achieving high recall in experience matching.

### 6.2.2 Semantic-driven Experience Retrieval

The effectiveness of retrieval directly determines the performance of the entire reuse framework. Traditional keyword-based and template-based matching struggle with the complexity and diversity of natural language. To overcome this, we introduce a semantic matching strategy powered by LLMs, enabling high-accuracy and robust task recognition.

When a new task query $q_{\text{current}}$ arrives, the system does not immediately proceed with the standard reasoning pipeline. Instead, it first attempts retrieval from the experience pool $\mathcal{P}$ using a semantic matching function $\mathcal{M}$. The goal of $\mathcal{M}$ is to identify a semantically equivalent query from the set of historical queries $Q_{\mathcal{P}} = q_i \mid (q_i, S_i, A_i, \mathcal{T}_i) \in \mathcal{P}$.

Formally:

$$E_{\text{matched}} = \mathcal{M}(q_{\text{current}}, \mathcal{P})$$

Specifically, $\mathcal{M}$ leverages an LLM $\mathcal{L}$ to evaluate the semantic similarity between $q_{\text{current}}$ and each $q_i \in Q_{\mathcal{P}}$. By constructing prompts that incorporate contextual information, the model considers multiple dimensions such as operation intent and target entities. If the model determines that a semantically equivalent query $q_k$ exists, it returns the corresponding experience $E_k$; otherwise, it returns $\emptyset$:

$$\mathcal{M}(q_{\text{current}}, \mathcal{P}) = \begin{cases} E_k & \text{if } \exists E_k \in \mathcal{P} \text{ s.t. } q_k \text{ matched } q_{\text{current}} \\ \emptyset & \text{otherwise} \end{cases}$$

The performance of this mechanism is evaluated by the Experience Hit Rate $H$, defined as the ratio of tasks successfully matched to reusable experiences $N_{\text{hit}}$ over the total number of tasks $N_{\text{total}}$:

$$H = \frac{N_{\text{hit}}}{N_{\text{total}}}$$

A higher hit rate is directly associated with overall system efficiency improvement.

### 6.2.3 Log-based Task Replay Execution

Once a matching experience $E_{\text{matched}}$ is retrieved, the system bypasses perception, planning, and decision-making modules, and activates the log replay mechanism.

The replay module directly reads the recorded action sequence $A_{\text{matched}} = \langle a_1, a_2, \ldots, a_n \rangle$ and executes each atomic action $a_i$ strictly in the original order. Since this process involves no real-time reasoning or visual analysis, execution speed far exceeds that of the standard pipeline.

We define the efficiency gain $\eta$ as follows. Let $T_{\text{std}}$ be the average execution time of the standard reasoning pipeline, and $T_{\text{replay}}$ be the average execution time of log replay:

$$\eta = \frac{T_{\text{std}} - T_{\text{replay}}}{T_{\text{std}}} = 1 - \frac{T_{\text{replay}}}{T_{\text{std}}}$$

According to our preliminary evaluation, in scenarios with stable UI layouts and fixed task structures, this mechanism can achieve efficiency gains of $40\% \sim 60\%$. Furthermore, log replay offers determinism and interpretability, providing strong support for task debugging, auditing, and the construction of more advanced memory systems.

## 6.3 Control Optimization via Function Calling

The function calling paradigm (Function Calling[1]) enables LLMs to automatically invoke external tool functions or APIs through structured data formats. Although GUI-based control provides broad generality, it inherently suffers from limitations in execution efficiency, stability, and completeness of information acquisition. To build a more efficient and robust agent, we introduce an API-based function calling paradigm and integrate it with GUI control, forming a hybrid control framework.

The core idea of this framework is to expand the agent's action space. We define the total action space $A_{\text{total}}$ as the union of the GUI atomic operation space $A_{\text{GUI}}$ and the function calling action space $A_{\text{Function Call}}$:

$$A_{\text{total}} = A_{\text{GUI}} \cup A_{\text{Function Call}} \tag{6.1}$$

Here, $A_{\text{GUI}}$ includes operations that directly interact with UI elements such as "click(element)" and "input(element, text)" (see Table3.6). Each action $a_i \in A_{\text{Function Call}}$ is a structured function call represented as a tuple:

$$a_i = (f_i, P_i), \quad P_i = \{p_k \mapsto v_k\}_{k=1}^{m} \tag{6.2}$$

where $f_i$ is a predefined function name, and $P_i$ is the set of key-value parameter pairs. In this framework, to ensure accuracy and reduce model dependency, function call actions are actively driven and selected by the user rather than determined by the model. Based on intent $T$, the user matches a suitable function and parameter list $a_{\text{match}} = (f_{\text{match}}, P_{\text{match}}) \in A_{\text{Function Call}}$ from the predefined domain-specific API toolkit. For each predefined parameter $p_k \in P_{\text{match}}$, the user may provide a custom value or use the system's default value $p_k^{(0)}$. Finally, the user executes the function call in a structured format on the terminal, completing the task at lower cost and shorter latency, with GUI operations serving as a general fallback. Moreover, the function-calling optimization is implemented as an independent, minimal, realizable module, enhancing control efficiency.

[1] https://platform.openai.com/docs/guides/function-calling

*Implementation and Application of Domain-specific API Toolkits* To validate the effectiveness of this hybrid framework, we developed and integrated two domain-specific API toolkits to extend the agent's capabilities.

First, we incorporated an *email service toolkit* $Tool_{\text{email}}$, which encapsulates low-level mail transfer protocols into high-level semantic interfaces. For example, a core function $\mathrm{send_email(to, subject, body, attachments)}$ abstracts what would otherwise be a complex GUI sequence, launching an APP, navigating the interface, entering text, selecting files, and transferring files across devices, into a single deterministic function call. This dramatically improves automation efficiency in scenarios such as batch invoice sending or system alert notifications. Moreover, through its parameterized attachment interface, it resolves the challenge of cross-application file transfer on Android devices.

Second, leveraging existing open-source libraries, we integrated an *automated social media interaction toolkit* $Tool_{\text{social}}$, exemplified by the Bilibili platform. This toolkit provides atomic function calls such as $\mathrm{like(video_id)}$, $\mathrm{coin(video_id, count)}$, and $\mathrm{search(keyword)}$. By invoking these tools, the agent can conduct deeper social interactions and information retrieval tasks. For instance, it can automatically search for content creators based on user preferences, or perform keyword-driven searches (e.g., "LLM pretraining"), which are difficult to achieve efficiently in a single step through pure GUI control.

*Methodological Analysis and Discussion* Integrating function calling into a GUI-agent framework fundamentally represents a trade-off between *generality* and *efficiency*.

Function calling methods demonstrate superior accuracy and speed in predefined tasks, with determinism and structured data advantages unmatched by pure GUI methods. However, this advantage comes at the cost of generality. The core bottleneck lies in dependency on external APIs: the agent's functional boundary is constrained to the set of encapsulated interfaces, leaving it unable to handle unseen tasks and vulnerable to changes in third-party APIs.

Therefore, we argue that the optimal agent architecture is not a single pathway but an organic integration of both. Our proposed hybrid control framework embodies such a pragmatic solution: GUI control provides a universal fallback ensuring broad applicability, while function calling delivers performance leaps in domains where APIs are available. This design allows the agent to operate any interface like a human while achieving superhuman efficiency and reliability in specific tasks like a program.

# 7
# Experimental Results and Analysis

The experimental results and analysis are divided into two main parts, with distinct evaluation focuses. Basic capability evaluation adopts a quantitative approach to assess the agent's fundamental capabilities. Scenario capability evaluation employs a qualitative method to evaluate the agent's practical execution capabilities.

## 7.1 Basic Capability Evaluation

This evaluation supports long-horizon capability and generalization, further enhancing accuracy. It assesses action prediction in multi-step scenarios, requiring multimodal reasoning to generate precise actions based on user instructions, history, and screenshots. The evaluation uses five benchmarks, including AndroidControl [36] [Low/High], GUI-Odyssey [42], AITZ [83] and CAGUI [86] benchmark. The core metrics for evaluation are Type Match (TM) and Exact Match (EM), where TM verifies the correctness of action types and EM requires a full match of parameters. Baseline Models include closed-source ones (GPT-5 [45], GPT-5-mini, GPT-4o [46], Gemini 2.0 [22], Claude [1]) and open-source ones (Qwen2.5-VL-7B [4], UI-TARS-7B [52], OS-Genesis-7B [62], OS-Atlas-7B [77], Aguvis-7B [79], OdysseyAgent [42], AgentCPM-GUI [86] ). Table 7.1 presents the results of the main methods across different datasets. Many closed-source models underperform in Chinese scenarios. The fine-tuned model performs reasonably well on Chinese data. AppCopilot, optimized on AgentCPM-GUI, achieves comparable performance in multi-dataset generalization evaluations. This validates AppCopilot's cross-lingual generalizability and ability to handle general and specialized tasks, with only a minor performance drop. The following will vividly demonstrate its execution in different scenarios through case studies.

**Table 7.1:** *Step-Level Action Prediction on Five Benchmarks (TM/EM). Bold/underline: best/second-best. *OS-Atlas uses different train/test split.*

| Model | AC-Low | | AC-High | | Odyssey | | AITZ | | CAGUI | |
|---|---|---|---|---|---|---|---|---|---|---|
| | TM | EM | TM | EM | TM | EM | TM | EM | TM | EM |
| *Closed-Source* | | | | | | | | | | |
| GPT-5-mini | 16.8 | 16.8 | 15.6 | 15.5 | - | - | 16.2 | 11.7 | 12.6 | 9.7 |
| GPT-5 | 69.8 | 69.1 | 56.3 | 55.2 | - | - | 56.0 | 31.3 | 70.6 | 33.0 |
| GPT-4o | - | 19.5 | - | 20.8 | - | 20.4 | 70.0 | 35.3 | 3.67 | 3.67 |
| Gemini 2.0 | - | 28.5 | - | 60.2 | - | 3.27 | - | - | - | - |
| Claude | - | 19.4 | - | 12.5 | 60.9 | - | - | - | - | - |
| *Open-Source* | | | | | | | | | | |
| Qwen2.5-VL-7B | 94.1 | 85.0 | 75.1 | 62.9 | 59.5 | 46.3 | 78.4 | 54.6 | 74.2 | 55.2 |
| UI-TARS-7B | **95.2** | **91.8** | **81.6** | 74.4 | 86.1 | 67.9 | 80.4 | 65.8 | 88.6 | 70.3 |
| OS-Genesis-7B | 90.7 | 74.2 | 65.9 | 44.4 | 11.7 | 3.63 | 20.0 | 8.45 | 38.1 | 14.5 |
| OS-Atlas-7B | 73.0 | 67.3 | 70.4 | 56.5 | 91.8* | 76.8* | 74.1 | 58.5 | 81.5 | 55.9 |
| Aguvis-7B | 93.9 | 89.4 | 65.6 | 54.2 | 26.7 | 13.5 | 35.7 | 19.0 | 67.4 | 38.2 |
| OdysseyAgent | 65.1 | 39.2 | 58.8 | 32.7 | 90.8 | 73.7 | 59.2 | 31.6 | 67.6 | 25.4 |
| AgentCPM-GUI | 94.4 | 90.2 | 77.7 | 69.2 | 90.9 | 75.0 | **85.7** | **76.4** | **96.9** | **91.3** |
| AppCopilot | 91.5 | 90.3 | 78.3 | **76.6** | **92.2** | **78.4** | 85.4 | 75.6 | 96.6 | 90.9 |

## 7.2 Scenario Capability Evaluation

### 7.2.1 Basic Scenario Task Evaluation Results and Analysis

#### Basic Communication Tasks

Task 1.1: Send SMS to grandson: "Are you coming back for dinner tonight?" / send message to mom, and tell her I will go back home

Task Objective: This task demonstrates capabilities in cross-lingual intent recognition, task decomposition, and contextual understanding.

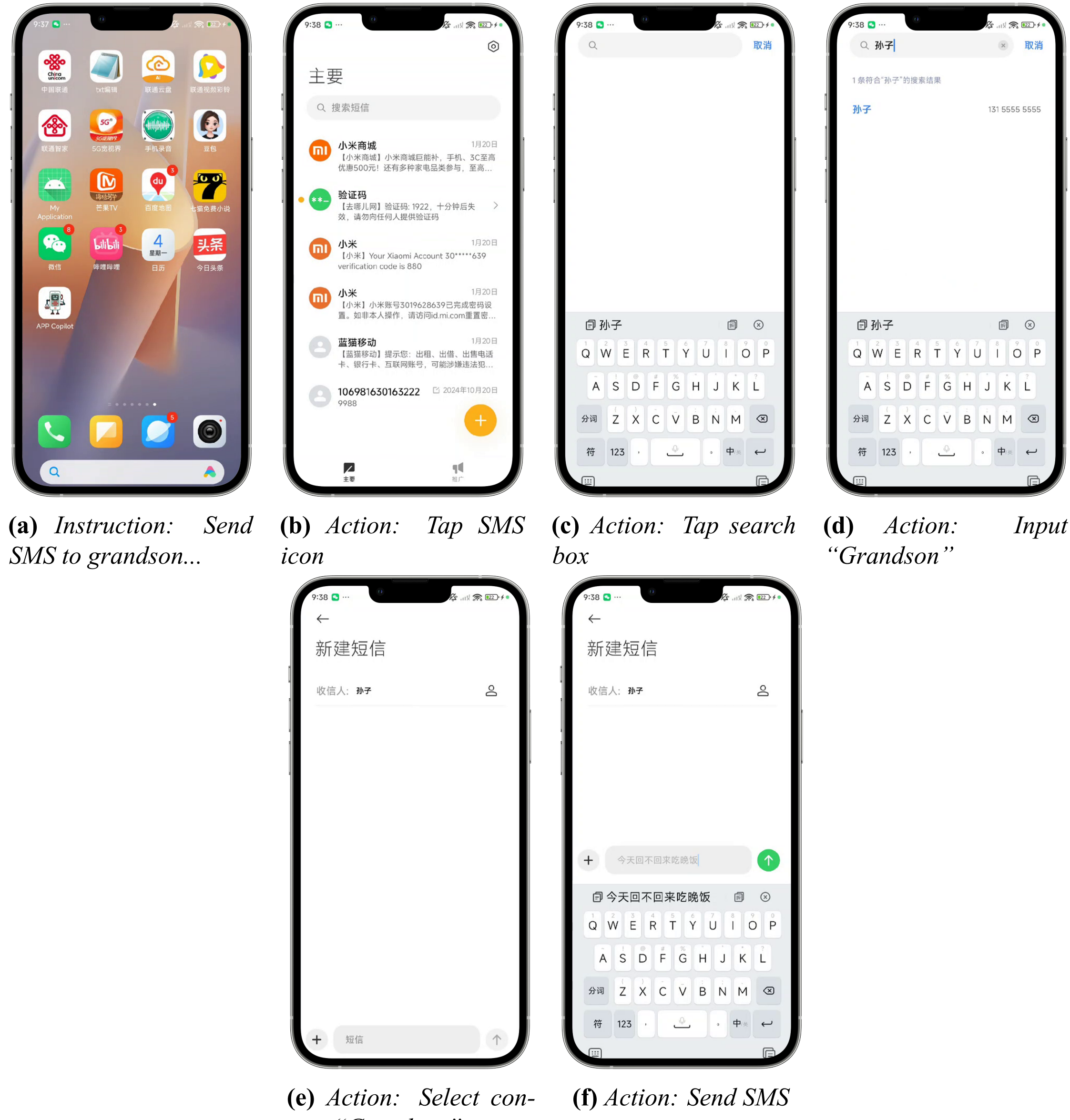

**(a)** *Instruction: Send SMS to grandson...*

**(b)** *Action: Tap SMS icon*

**(c)** *Action: Tap search box*

**(d)** *Action: Input "Grandson"*

**(e)** *Action: Select contact "Grandson"*

**(f)** *Action: Send SMS*

**Figure 7.1:** *Instruction: Send SMS to grandson:"Are you coming back for dinner tonight?"*

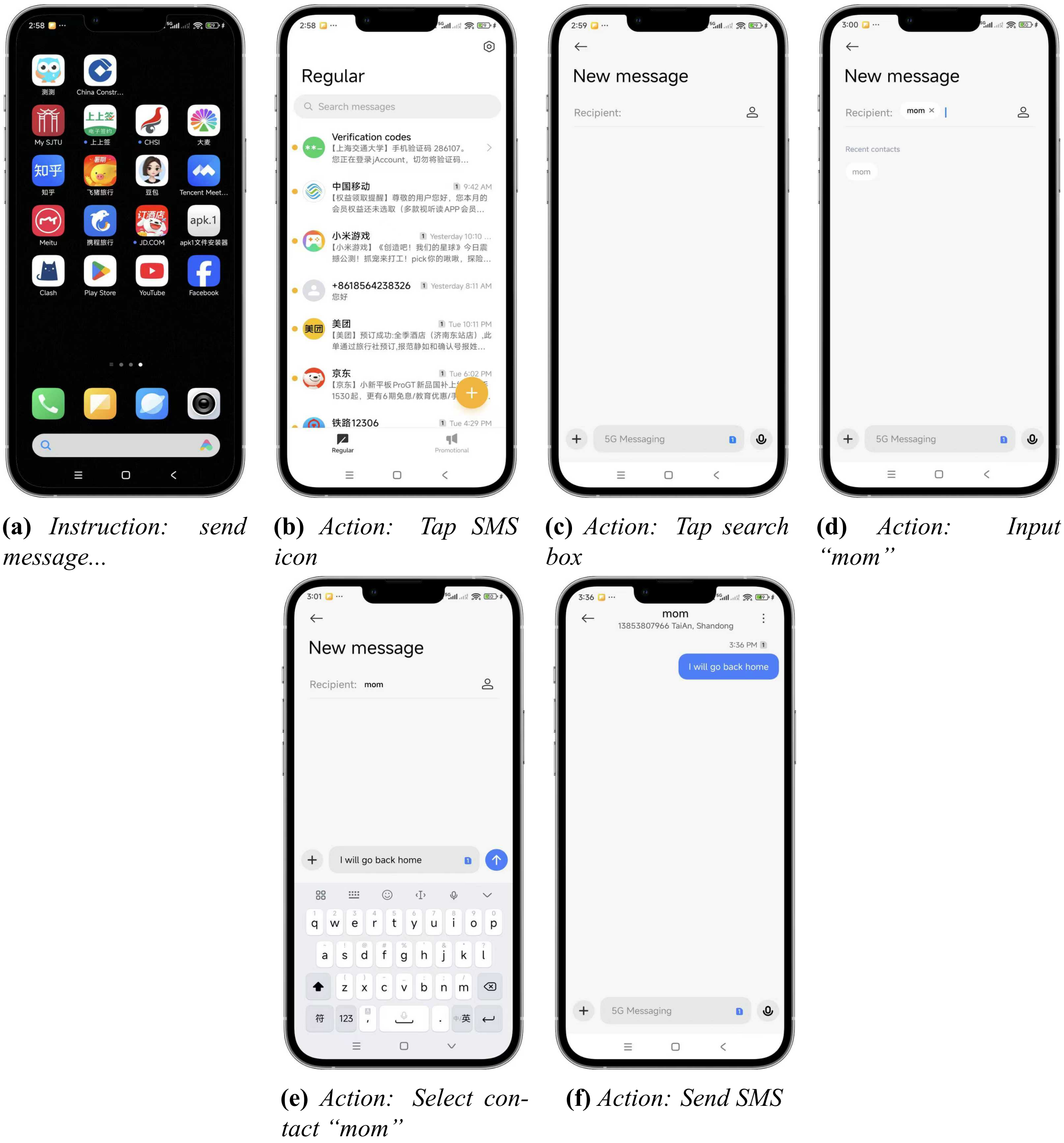

**(a)** *Instruction: send message...* **(b)** *Action: Tap SMS icon* **(c)** *Action: Tap search box* **(d)** *Action: Input "mom"*

**(e)** *Action: Select contact "mom"* **(f)** *Action: Send SMS*

**Figure 7.2:** *Instruction: send message to mom, and tell her I go back home*

*Execution Process* As shown in Figure 7.1, during initial execution, the agent completes multi-level parsing of the instruction. It accurately captures the core action intent "send SMS" and clearly distinguishes two key elements: the recipient "Grandson" and the message content "Are you coming back for dinner tonight?" This structural decomposition capability allows the agent to clarify element boundaries and logical relationships through intent recognition, preventing confusion between recipient and content.

During action, the agent directly searches for the "Grandson" contact. This process demonstrates precise intent recognition and contextual understanding. The core of task planning lies in mapping abstract kinship terms in natural language to specific contact entities, requiring both awareness of contact data structures and commonsense understanding.

Additionally, the agent accurately populates the message content without deviation, reflecting high-fidelity information capture and restoration capabilities. In the transformation from natural language to interactive content, the agent avoids information omissions, semantic distortions, or formatting errors. The SMS content perfectly matches the instruction description, indicating high-fidelity information transformation capabilities crucial for executing complex text instructions.

The task successfully completes sending specific content to a specific recipient. Through the complete demonstration of the "send specific SMS to grandson" task, it showcases the agent's capabilities in intent recognition, task decomposition, and contextual understanding. The system can accurately split and map natural language contacts and message content to specific interaction steps, maintaining execution accuracy and consistency even with incomplete information.

In the execution of the English instruction "send message to mom, and tell her I will go back home" (Figure 7.2), the agent similarly demonstrates precise parsing capabilities, accurately identifying the core action "send message", recipient "mom", and content "I will go back home", while fully inputting the English content. This verifies the system's language-agnostic parsing capability.

Through bilingual scenario demonstrations, this task comprehensively showcases the agent's capabilities in intent recognition, task decomposition, and contextual understanding. The system can accurately map natural language contacts and message content to interaction steps, maintaining execution accuracy and consistency in multilingual environments.

Task 1.2: Call Mom

Task Objective: This task highlights capabilities in ambiguous semantic parsing and user module invocation.

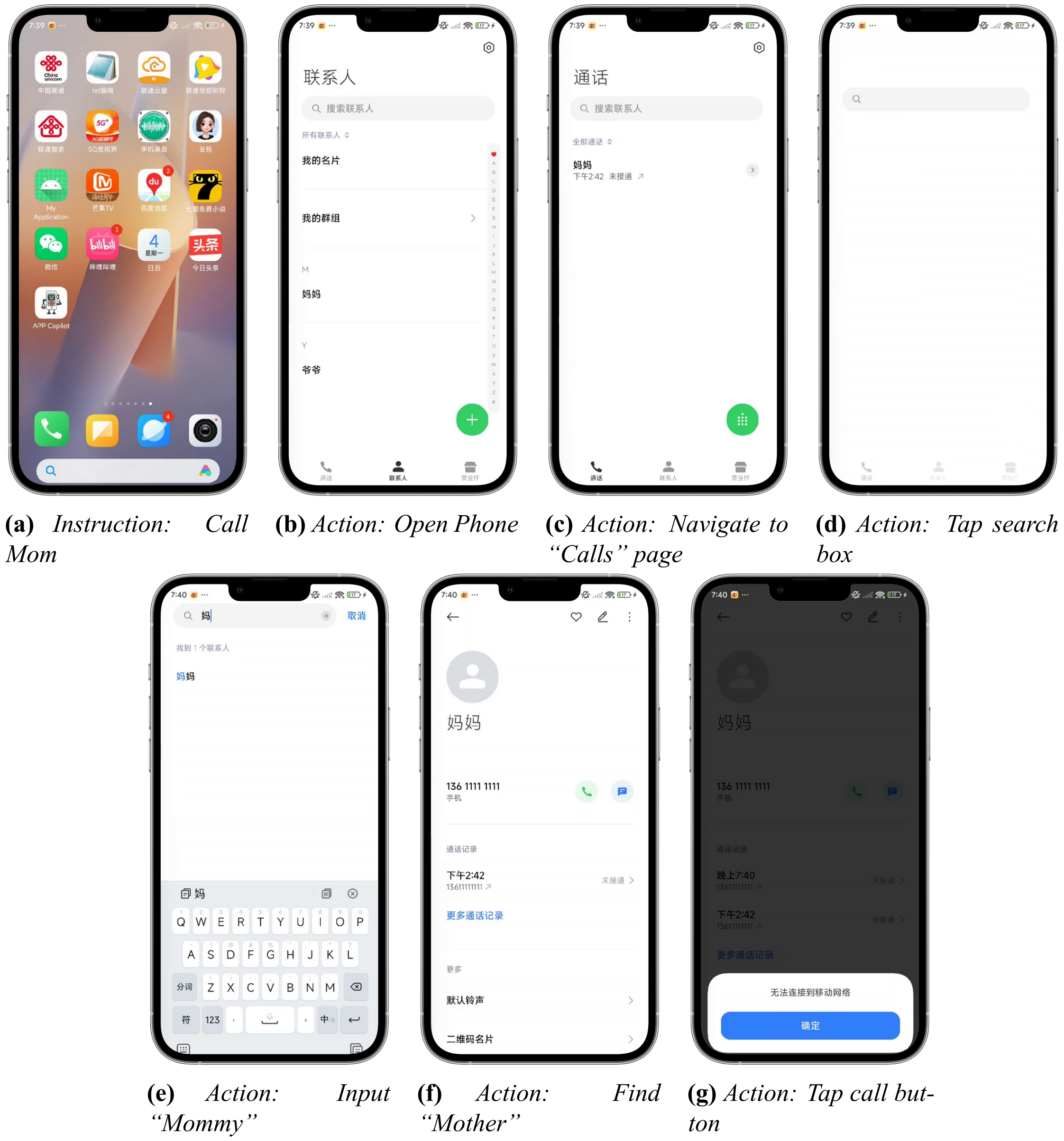

**(a)** *Instruction: Call Mom* **(b)** *Action: Open Phone* **(c)** *Action: Navigate to "Calls" page* **(d)** *Action: Tap search box*

**(e)** *Action: Input "Mommy"* **(f)** *Action: Find "Mother"* **(g)** *Action: Tap call button*

**Figure 7.3:** *Instruction: Call Mom (First Execution)*

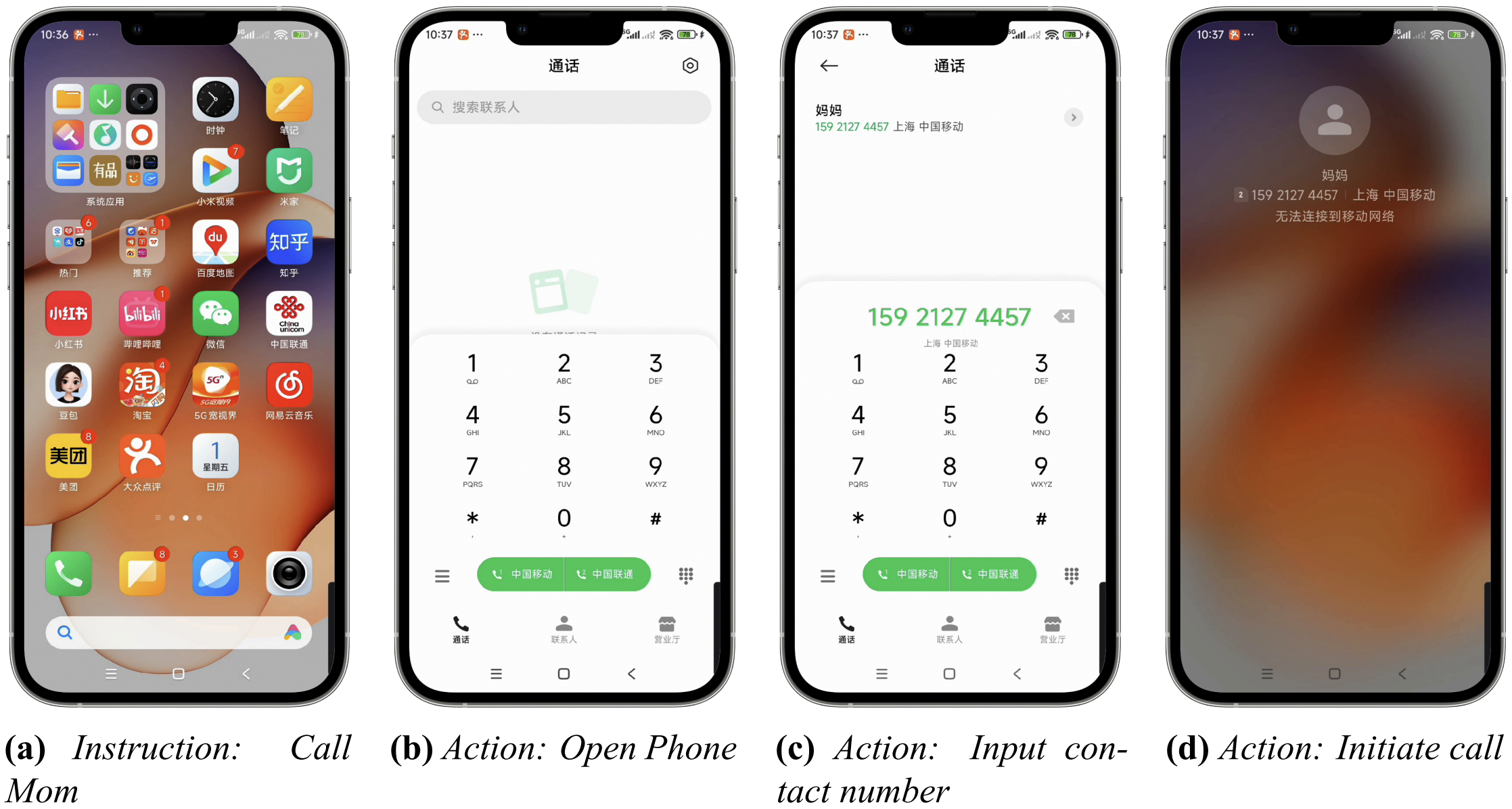

**(a)** *Instruction: Call Mom* **(b)** *Action: Open Phone* **(c)** *Action: Input contact number* **(d)** *Action: Initiate call*

**Figure 7.4:** *Instruction: Call Mom (Second Execution)*

*Execution Process* As shown in Figure 7.3, during the first execution without user module parameters, the agent relies entirely on real-time semantic parsing and interface interaction logic. During intent recognition, the agent accurately identifies the core action "make call" and generalizes the colloquial "Mom" to the more universal search keyword "Mommy". This process focuses on dynamic matching of ambiguous entities rather than precise positioning. The agent completes entity association through interface search functions, essentially an autonomous exploration behavior without prior knowledge. During execution, the agent strictly follows standard GUI interaction processes: open phone APP → enter search interface → input keyword "Mommy" → match contact results → finally locate the contact corresponding to "Mother". This approach relies on the synonym expansion capability of the natural language understanding module and the element recognition accuracy of the GUI control module, adapting to possible variations in contact naming while providing visual verification opportunities.

As shown in Figure 7.4, during the second execution with the user module enabled, the agent's processing logic changes. The pre-stored "Mother" entity information in the user module becomes the core basis for intent recognition and task decomposition and planning. During intent recognition, the agent directly links "Mom" in the instruction with the "Mother" entity recorded in the user module. During execution, the agent retrieves the phone number corresponding to "Mother" by calling the user module interface and passes the number to the dialing function to complete the call. This path offers significant efficiency advantages, eliminating intermediate steps like search and matching, greatly reducing end-to-end response time. At the same time, since it relies on verified entity relationships in the user module, accuracy is more strongly guaranteed, avoiding matching errors caused by semantic ambiguity.

The system successfully completes the task of calling a person with a specific title. Through the two execution processes demonstrated above, it showcases capabilities in ambiguous semantic parsing, dynamic entity matching, and user module invocation.

### Transaction Service Tasks

Task 1.3: Recharge 50 yuan to this device's number via China Unicom

Task Objective: This task highlights capabilities in human-AI collaboration.

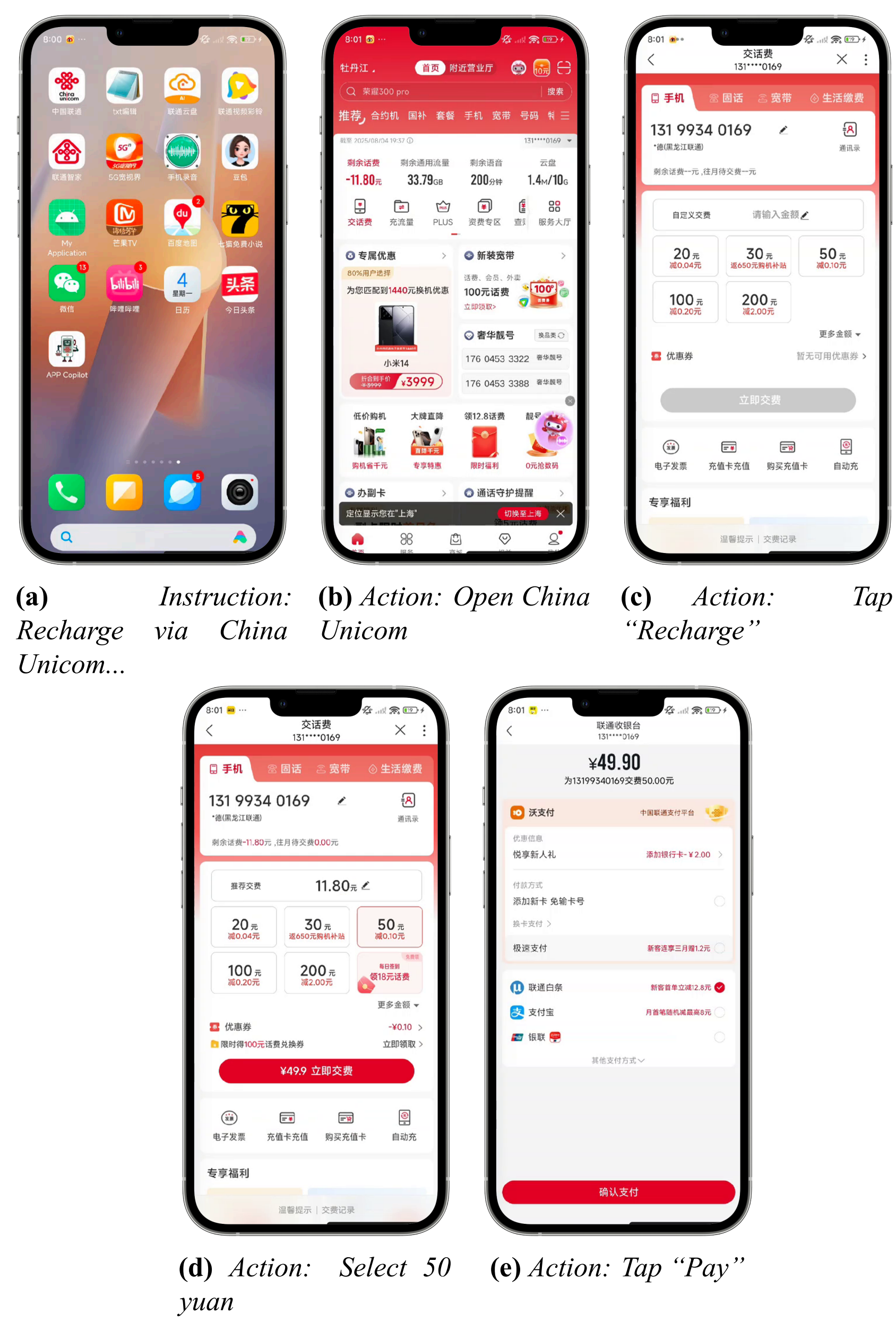

**(a)** *Instruction: Recharge via China Unicom...* **(b)** *Action: Open China Unicom* **(c)** *Action: Tap "Recharge"*

**(d)** *Action: Select 50 yuan* **(e)** *Action: Tap "Pay"*

**Figure 7.5:** *Instruction: Recharge 50 yuan to this device's number via China Unicom*

*Execution Process* As shown in Figure 7.5, during initial instruction execution, the agent accurately identifies and launches the China Unicom APP, directly locating the "Recharge" core function module. This process relies on its feature learning in vertical application scenarios, establishing direct mapping from instructions to functional entry points by recognizing UI components.

After entering the recharge interface, the agent automatically matches "this device's number" by reading device-bound user data, transforming the abstract "this device" expression into a specific phone number and filling it into the input box. For the "50 yuan" amount parameter, the agent correctly selects the corresponding numerical value. The entire process shows no number mismatches or amount selection errors, validating its precise parsing capability for entity information and numerical parameters.

When the process advances to the payment stage, the agent successfully navigates to the payment interface and actively pauses. This design is not accidental but incorporates reinforcement training for mechanisms requiring user feedback at critical nodes, aiming to enhance the agent's control over practical application security and understanding of human-AI interaction boundaries. Since recharges involve sensitive financial transactions, users typically need to confirm transaction information accuracy and choose payment methods independently. Therefore, we emphasize in training that the agent must actively stop the automation process at the payment interface rather than directly executing the payment action.

The task successfully completes the full process from application launch to payment interface, ultimately pausing at the pending payment interface awaiting user confirmation, demonstrating capabilities in vertical scenario function localization, precise entity parameter parsing, and human-AI collaboration.

Task 1.4: Please select and purchase appropriate benefits in China Unicom based on my video viewing habits

Task Objective: This task demonstrates capabilities in user preference analysis, task decomposition, and personalized decision-making.

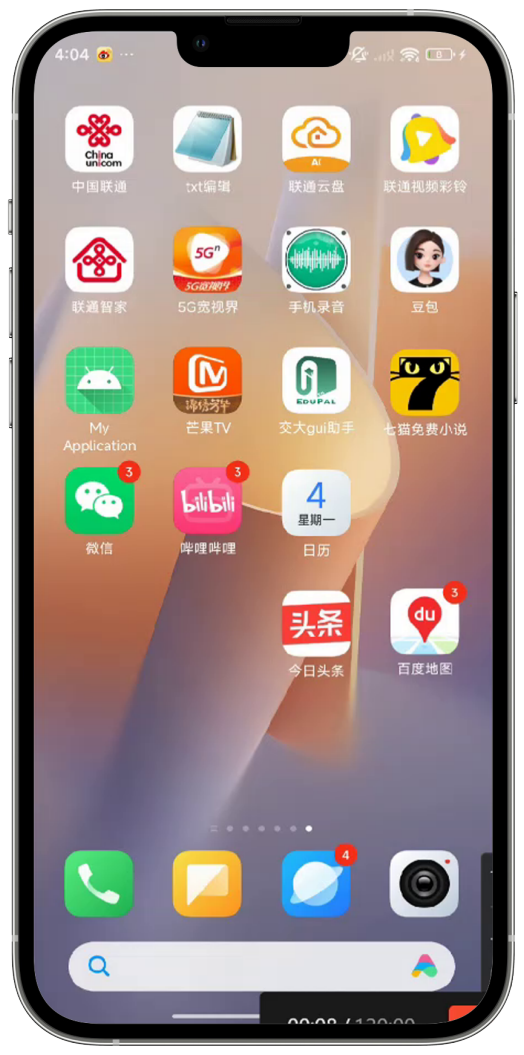

**(a)** *Instruction: Please select benefits...*

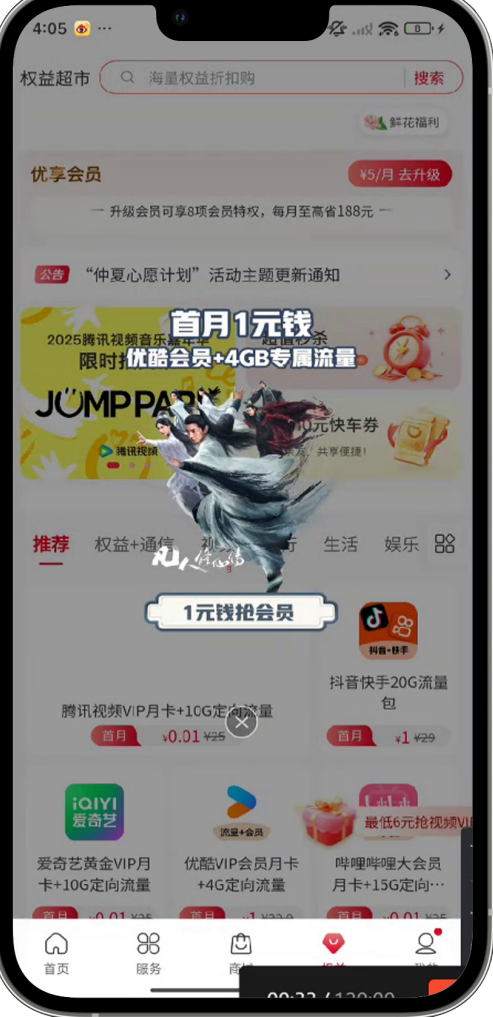

**(b)** *Action: Open China Unicom*

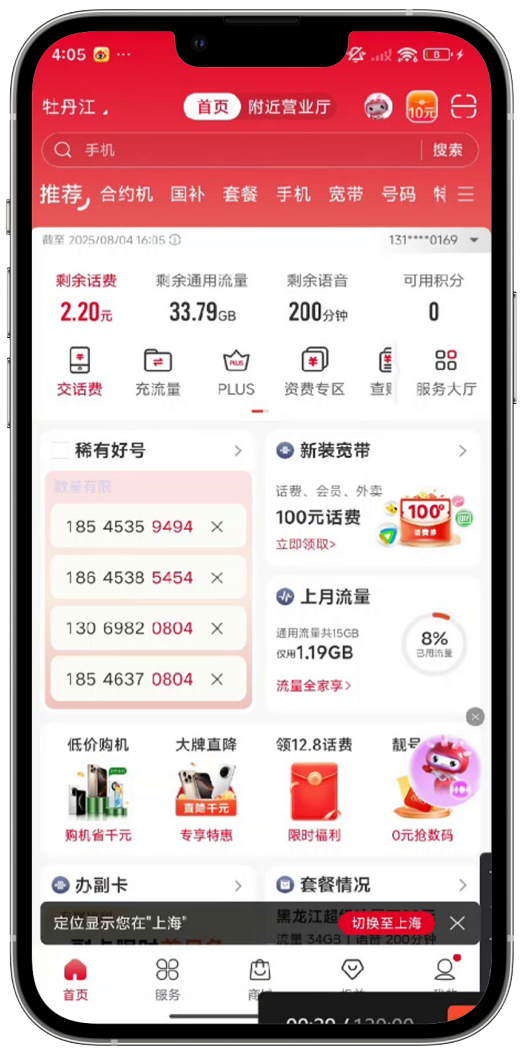

**(c)** *Action: Close ad*

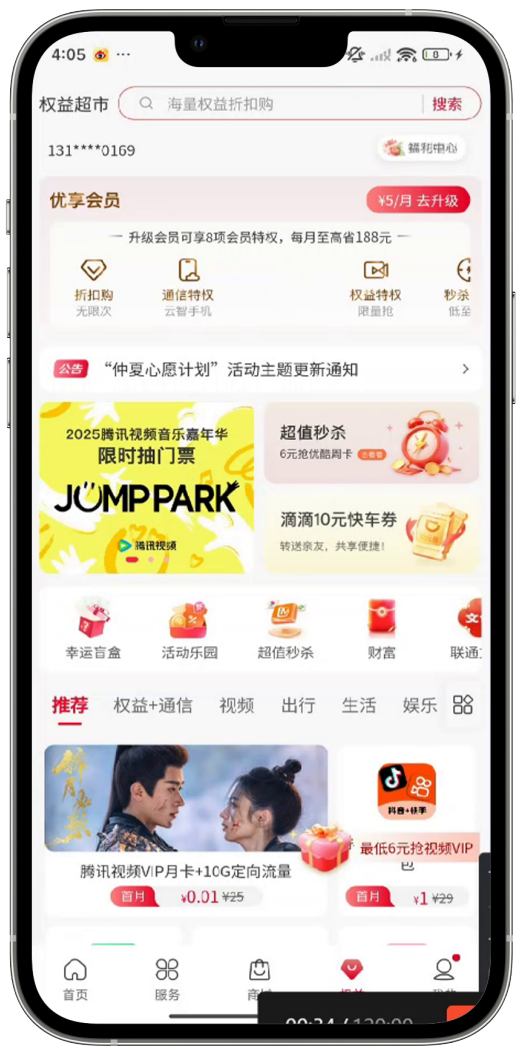

**(d)** *Action: Enter benefits interface*

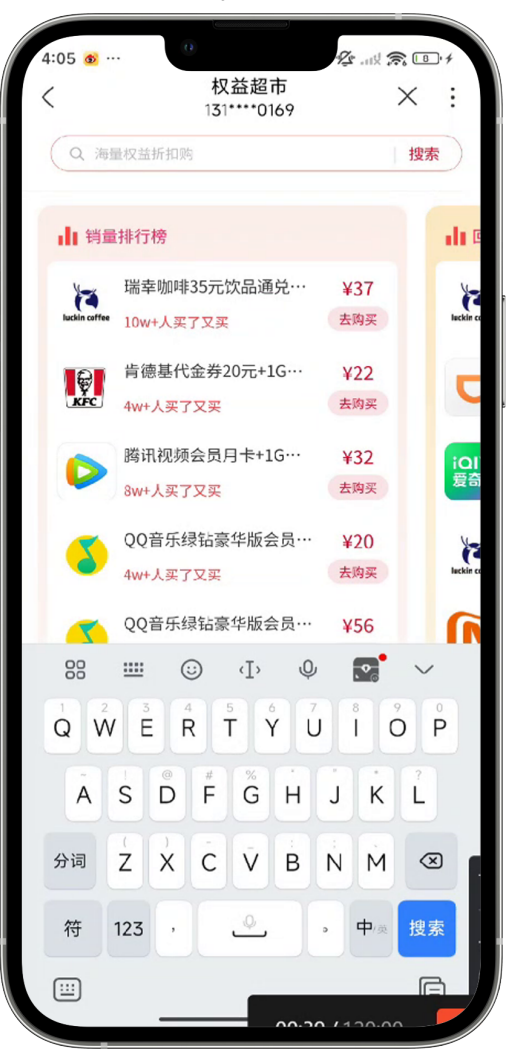

**(e)** *Action: Tap search box*

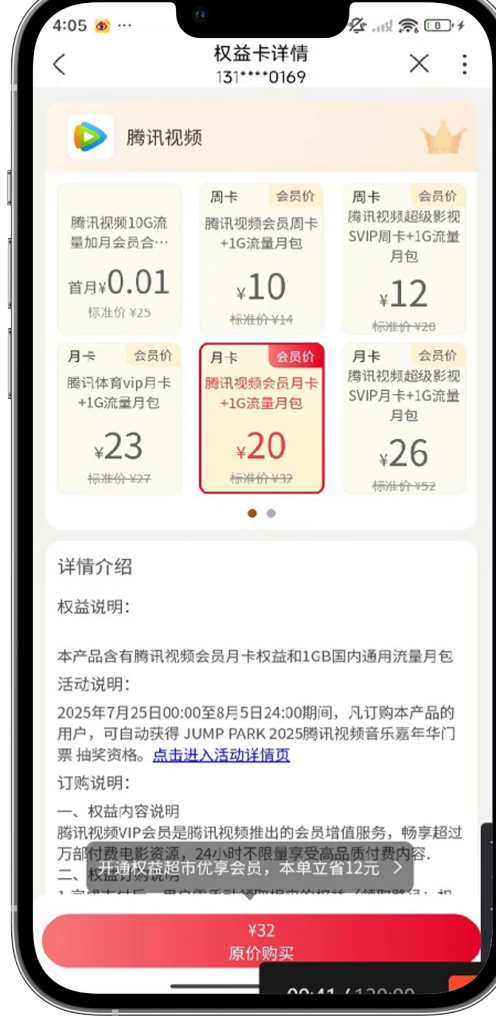

**(f)** *Action: Select corresponding benefit*

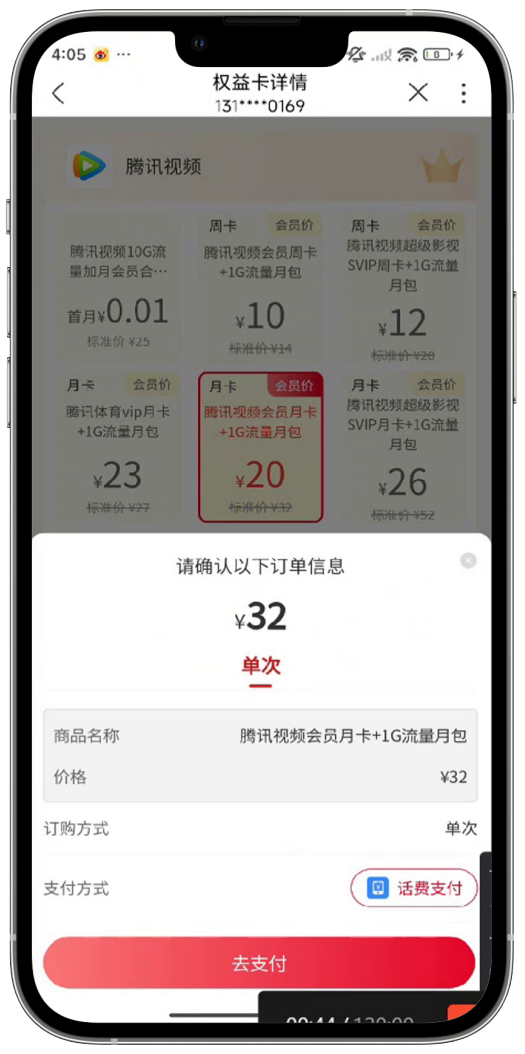

**(g)** *Action: Tap "Pay"*

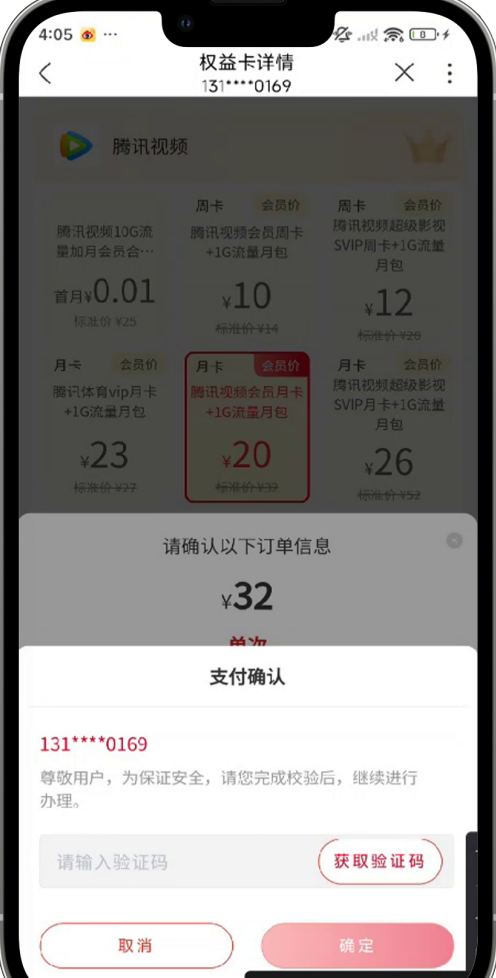

**(h)** *Action: Confirm payment*

**Figure 7.6:** *Instruction: Please select and purchase appropriate benefits in China Unicom based on my video viewing habits*

*Execution Process* As shown in Figure 7.6, during initial instruction execution, the agent demonstrates precise extraction of user preferences. The abstract requirement "video viewing habits" is resolved by invoking the user module to locate specific "Tencent Video" preference data. This process requires both stored behavioral records and the agent's ability to interpret data, extracting core preferences from scattered behavioral data. This transformation of habits into specific service types represents the concrete application of user profiling.

After obtaining the "Tencent Video" preference, the agent completes the entire screening and purchase process within the China Unicom APP. This choice reflects its accurate understanding of the instruction boundary to provide services within the Unicom ecosystem. The agent achieves full-chain automation of data invocation, preference analysis, benefit screening, and purchase guidance. In practical operation, the agent first autonomously skips interface ads, demonstrating generalization capabilities in unknown interaction scenarios and task execution efficiency. It then navigates to the benefits supermarket module, locates Tencent Video-related benefits through search, and guides to the purchase interface. This process involves multiple interface transitions and element interactions, yet the agent maintains process continuity without path deviation.

This task demonstrates the core value of agent personalized service. Unlike deterministic instructions like "recharge 50 yuan", "select appropriate benefits" requires the agent to make subjective judgments based on user data. This judgment both conforms to user preferences and fits the service scenario. This capability breaks through the limitations of traditional GUI mechanical operations, achieving active decision-making based on user data, essentially the integration of user understanding and business understanding capabilities.

The task is successfully completed by invoking user module preference data for personalized recommendation and purchase, demonstrating capabilities in user preference analysis, business scenario boundary control, and personalized decision-making.

#### Entertainment Content Tasks

Task 1.5: Browse short videos in Unicom Video Ringtone and like favorites

Task Objective: This task demonstrates capabilities in user preference analysis, multimodal information interaction, and autonomous execution.

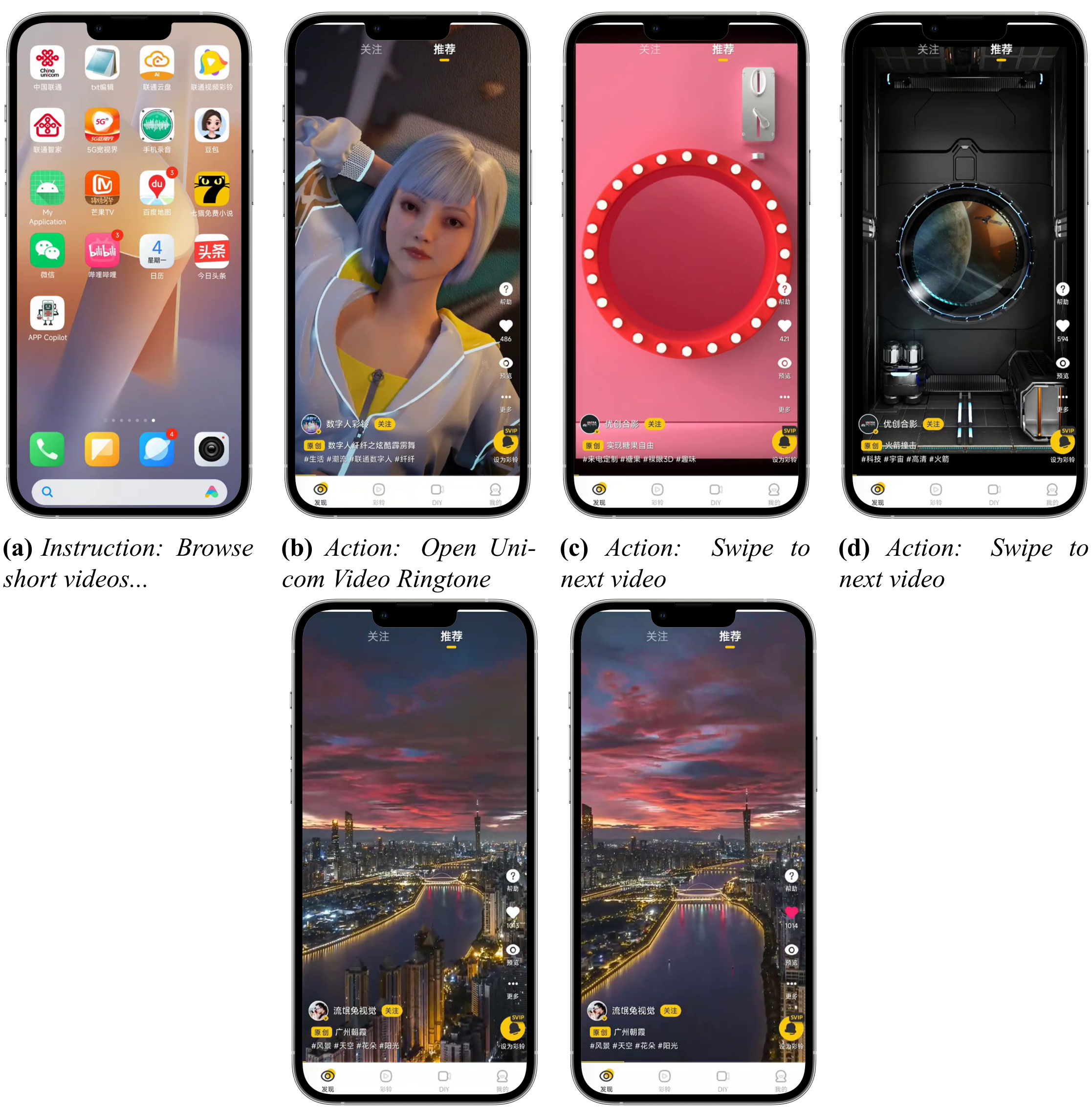

**(a)** *Instruction: Browse short videos...*

**(b)** *Action: Open Unicom Video Ringtone*

**(c)** *Action: Swipe to next video*

**(d)** *Action: Swipe to next video*

**(e)** *Action: Swipe to next video*

**(f)** *Action: Like*

**Figure 7.7:** *Instruction: Browse short videos in Unicom Video Ringtone and like favorites*

*Execution Process* As shown in Figure 7.7, during initial instruction execution, the agent demonstrates the ability to concretize the ambiguous requirement “favorites”. By invoking the “scenery” preference tag from the user module, it transforms abstract requirements into executable judgment criteria. This process relies not only on stored preference data but also requires semantic understanding of preference tags.

During video content recognition, since Unicom Video Ringtone’s short videos are dynamically loaded in a streaming manner, the agent needs to analyze frame content in real-time to determine compliance with “scenery” preference. This heavily depends on the model’s visual capabilities. In the application, the agent swipes the screen to load new videos, executing a play-identify-judge loop until finding scenery-aligned content. When identified, it proactively triggers the like operation. This dynamic decision-making relies on collaboration between the GUI control module and content understanding module.

The task highlights the core value of active service. Beyond simply “browsing videos”, the agent incorporates active judgment and feedback based on preferences. This active service capability depends on the agent’s deep understanding of user needs.

The task is successfully completed by combining the “scenery” preference tag to achieve personalized browsing and proactive liking, demonstrating capabilities in user preference understanding, real-time content recognition, dynamic interaction decision-making, and active service awareness.

Task 1.6: Watch Episode 10 of “Falling Flowers Meet Jun Again” on 5G Kuan Shijie and save to Unicom Cloud Drive

Task Objective: This task demonstrates capabilities in error tolerance and complex task decomposition.

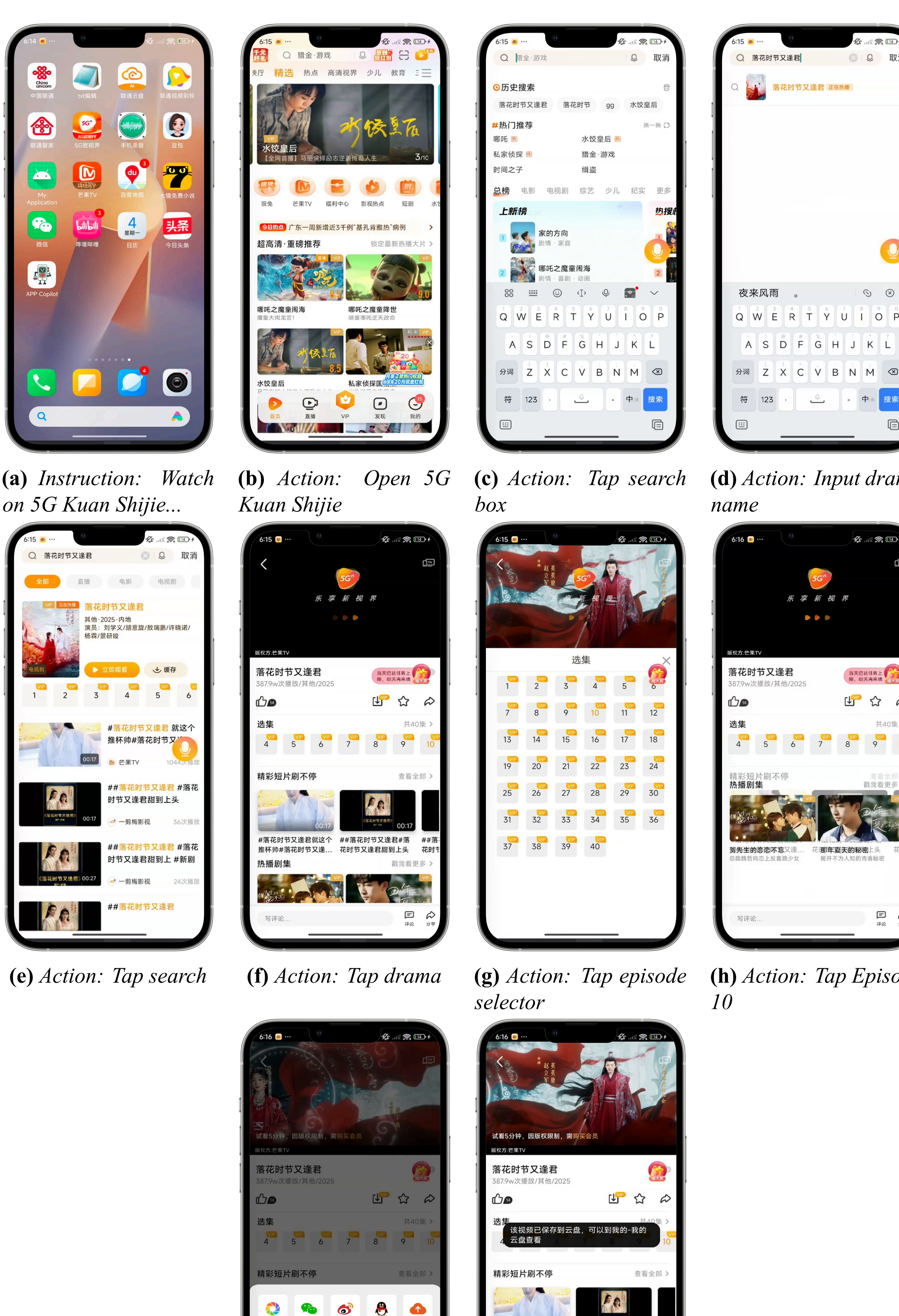

**(a)** *Instruction: Watch on 5G Kuan Shijie...*

**(b)** *Action: Open 5G Kuan Shijie*

**(c)** *Action: Tap search box*

**(d)** *Action: Input drama name*

**(e)** *Action: Tap search*

**(f)** *Action: Tap drama*

**(g)** *Action: Tap episode selector*

**(h)** *Action: Tap Episode 10*

**(i)** *Action: Tap share*

**(j)** *Action: Save to cloud drive*

**Figure 7.8:** *Instruction: Watch Episode 10 of "Falling Flowers Meet Jun Again" on 5G Kuan Shijie and save to Unicom Cloud Drive*

*Execution Process* As shown in Figure 7.8, during initial instruction execution, the agent accurately launches the 5G Kuan Shijie application and locates the drama through search functionality. This relies on understanding video application interaction logic. Precise drama name recognition prevents search failures.

During episode lookup error handling, the agent demonstrates active decision-making. When only the first six episodes are displayed by default, the agent proactively triggers the "All Episodes" list expansion. This behavior reflects problem-solving awareness, recognizing insufficient information and invoking alternative paths.

After successfully playing Episode 10, the agent completes the subsequent "save to cloud drive" task. This involves functional linkage between 5G Kuan Shijie and Unicom Cloud Drive. The agent locates the share-to-cloud function and confirms the storage operation.

The task validates the agent's capability to recognize and operate complex interface elements. As a video application, 5G Kuan Shijie contains rich multimedia elements. Experimental results show the agent can stably recognize various interface elements.

The task is successfully completed, demonstrating capabilities in contextual interaction understanding, error handling decision-making, cross-module collaboration, and interface adaptation robustness.

**Account and Information Query Tasks**

Task 1.7: Help me generate an account sheet for July's recharge records and send it to "Little Assistant" on WeChat

Task Objective: This task demonstrates capabilities in generalization, long-horizon context modeling, complex task decomposition, and planning.

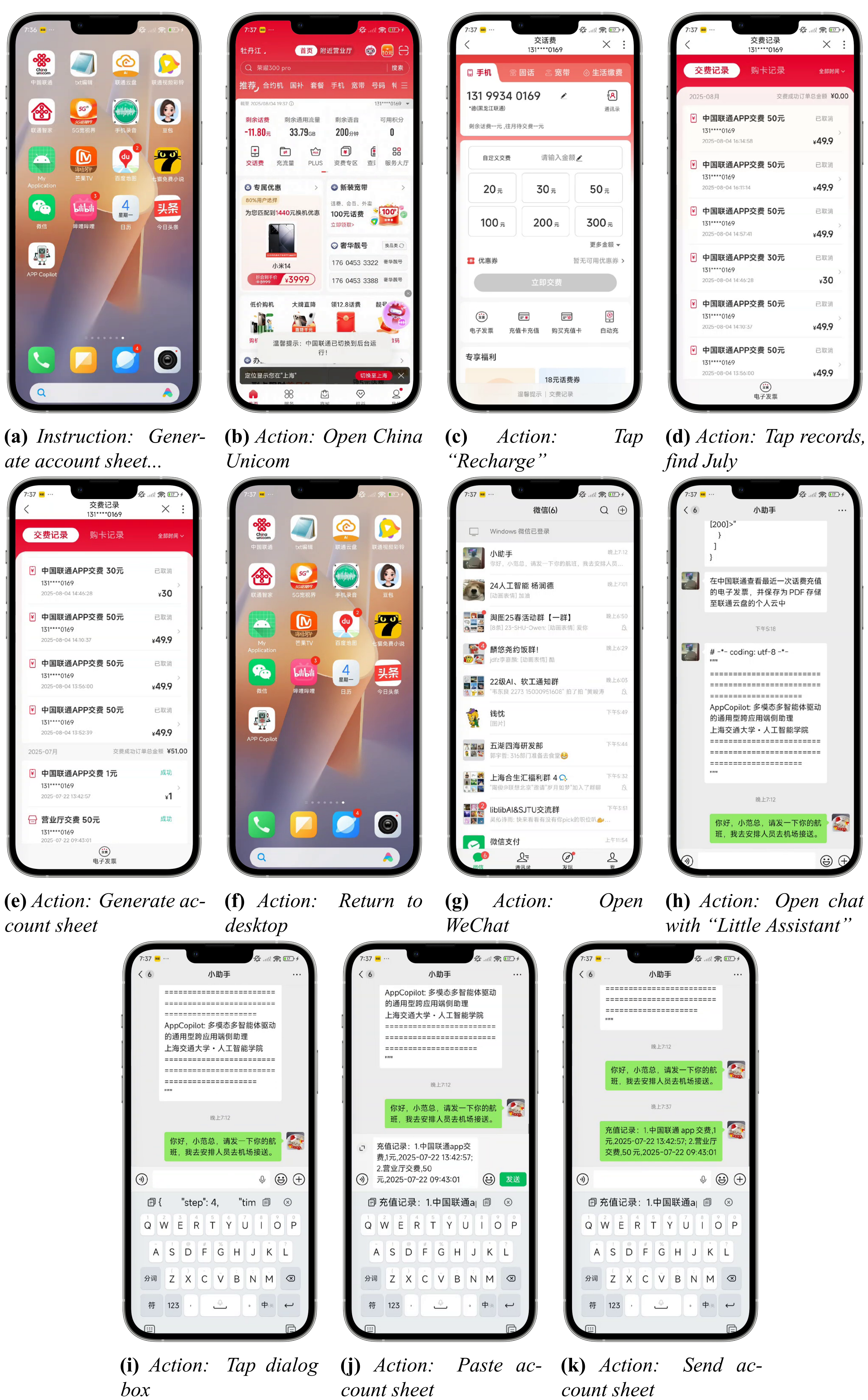

**(a)** *Instruction: Generate account sheet...* **(b)** *Action: Open China Unicom* **(c)** *Action: Tap "Recharge"* **(d)** *Action: Tap records, find July*

**(e)** *Action: Generate account sheet* **(f)** *Action: Return to desktop* **(g)** *Action: Open WeChat* **(h)** *Action: Open chat with "Little Assistant"*

**(i)** *Action: Tap dialog box* **(j)** *Action: Paste account sheet* **(k)** *Action: Send account sheet*

**Figure 7.9:** *Instruction: Generate account sheet for July's recharge records and send to "Little Assistant" on WeChat*

*Execution Process* As shown in Figure 7.9, this task chains four core processes: data retrieval, information extraction, text generation, and cross-APP sending. The entire instruction involves two applications with multiple critical steps.

The agent progresses in logical order: first finds recharge records, locates "July records" through swiping, generates the account text, then switches to WeChat for sending. This long-horizon coherence stems from global task understanding.

During record retrieval, the agent's swiping behavior adapts to real interaction scenarios. Unicom APP typically displays records in reverse chronological order. The agent simulates human operation while understanding temporal filtering logic.

After finding July records, the agent extracts key recharge information through OCR recognition and structured information extraction technologies.

After generating the account sheet, the agent switches from China Unicom APP to WeChat. This involves system-level operations. Smooth execution demonstrates global control in multi-app environments.

#### Tool Efficiency Tasks

Task 1.8: View the e-invoice for the last recharge in China Unicom and save as PDF to Personal Cloud in Unicom Cloud Drive

Task Objective: This task demonstrates capabilities in precise intent parsing, complex task decomposition, and planning.

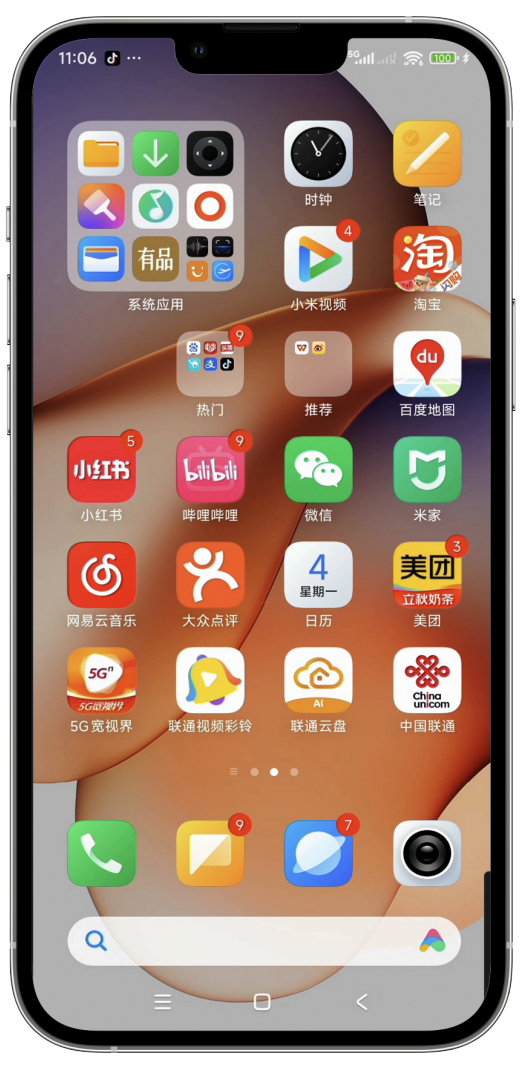

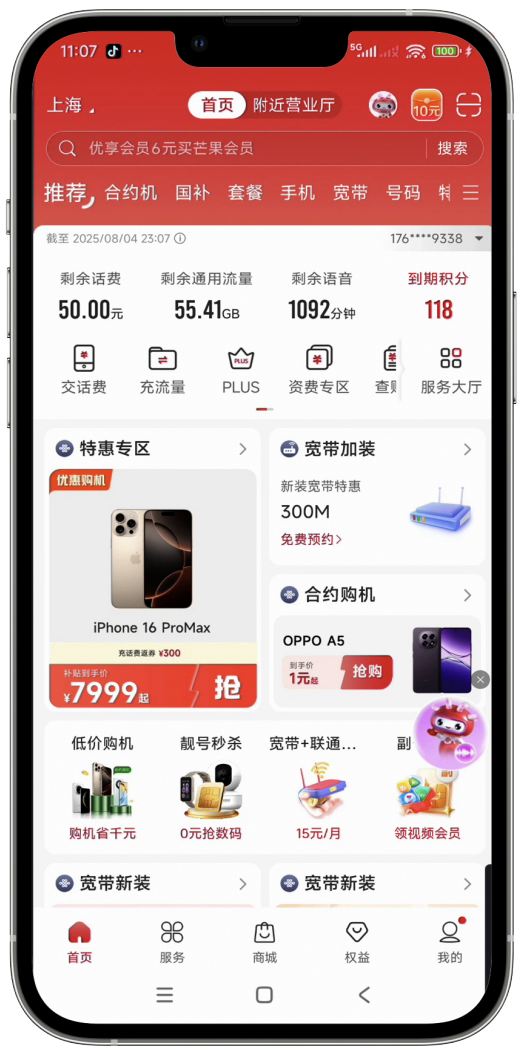

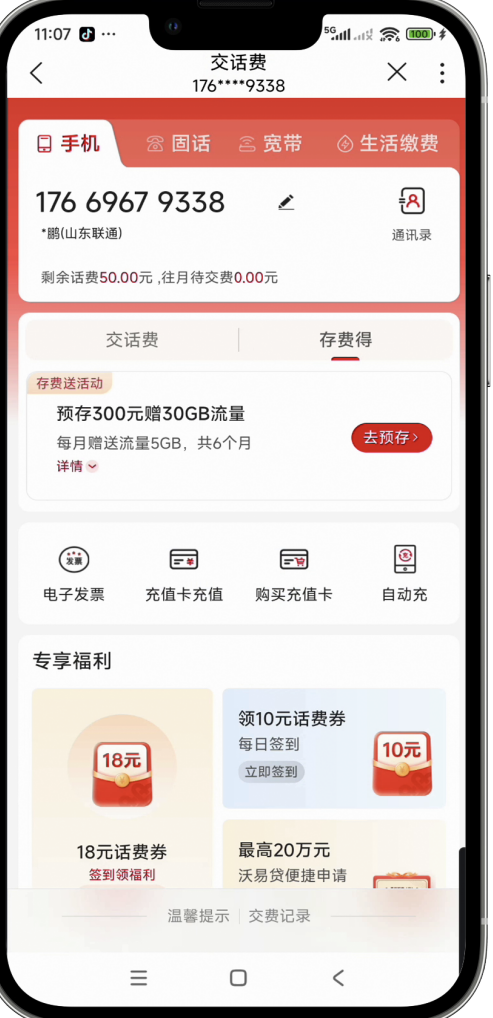

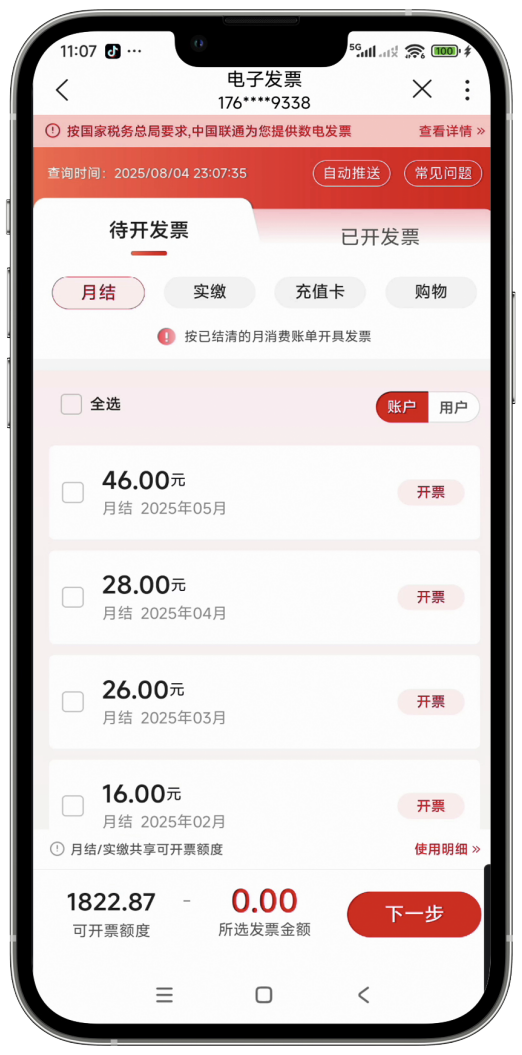

**(a)** *Instruction: View e-invoice...*

**(b)** *Action: Open China Unicom*

**(c)** *Action: Tap "Recharge"*

**(d)** *Action: Tap "E-invoice"*

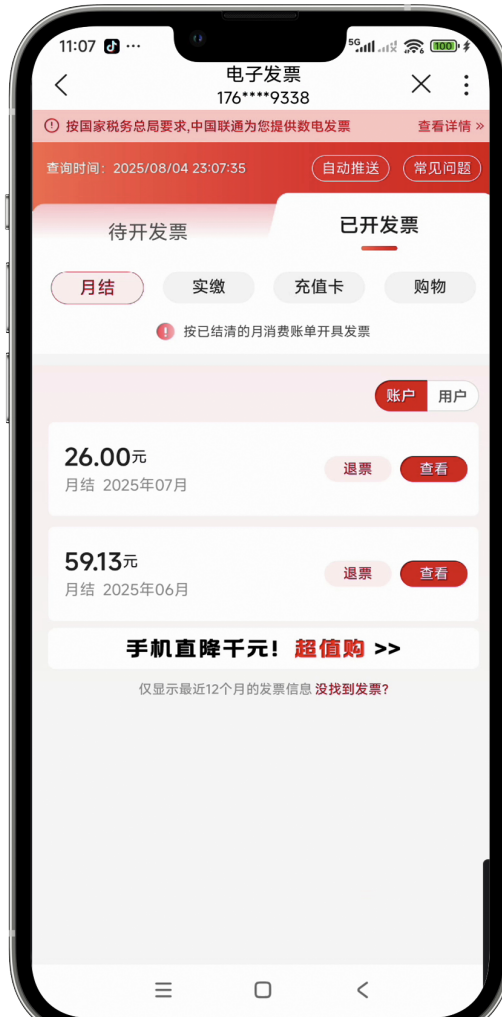

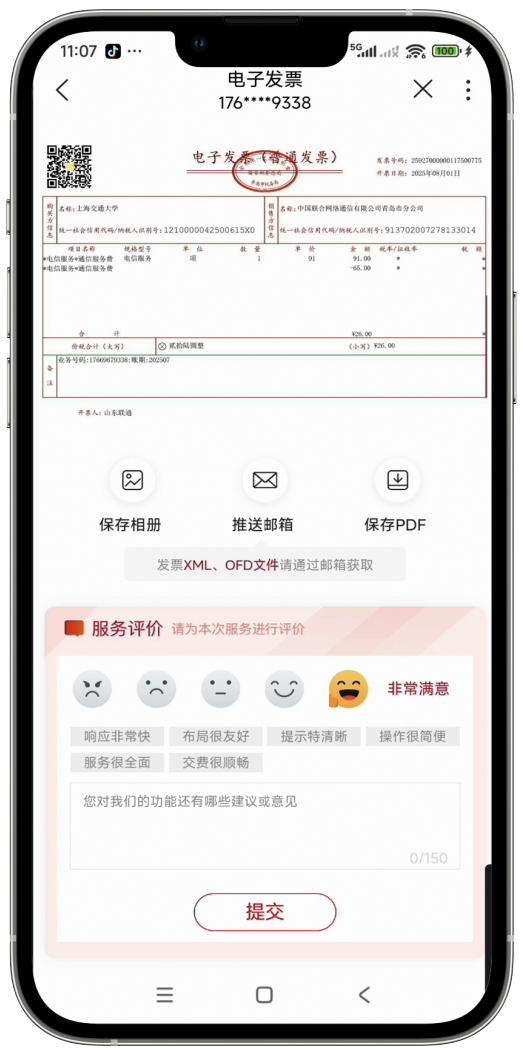

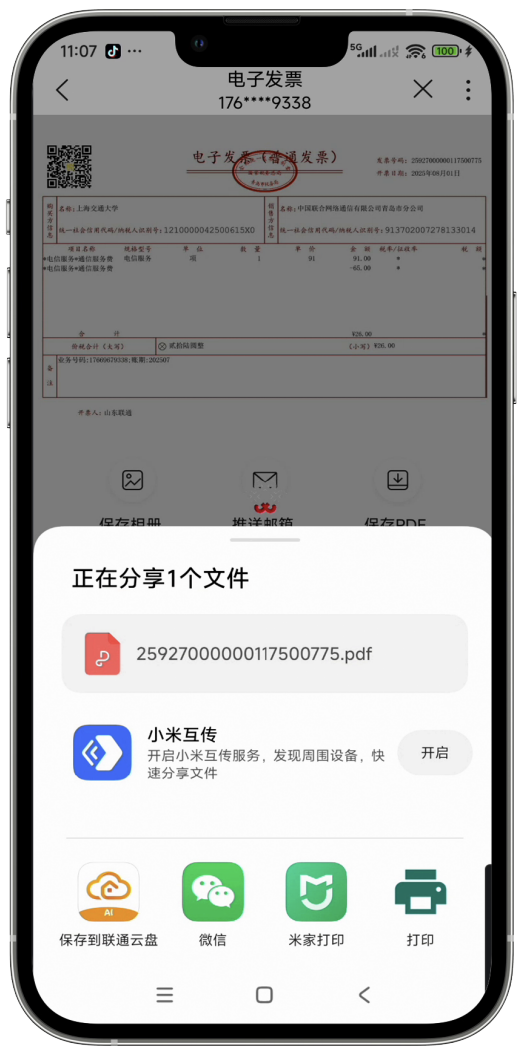

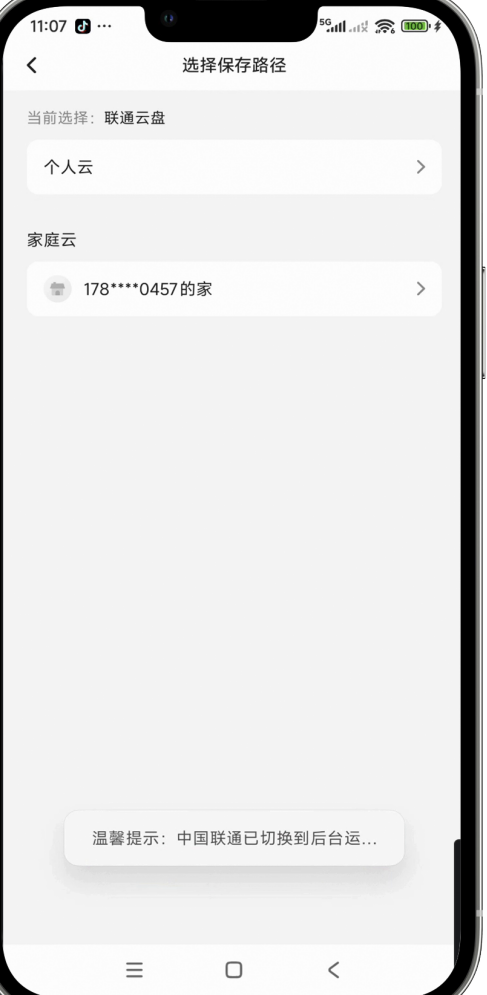

**(e)** *Action: Tap "Issued Invoices"*

**(f)** *Action: View last recharge's e-invoice*

**(g)** *Action: Tap "Save PDF"*

**(h)** *Action: Tap "Save to Cloud Drive"*

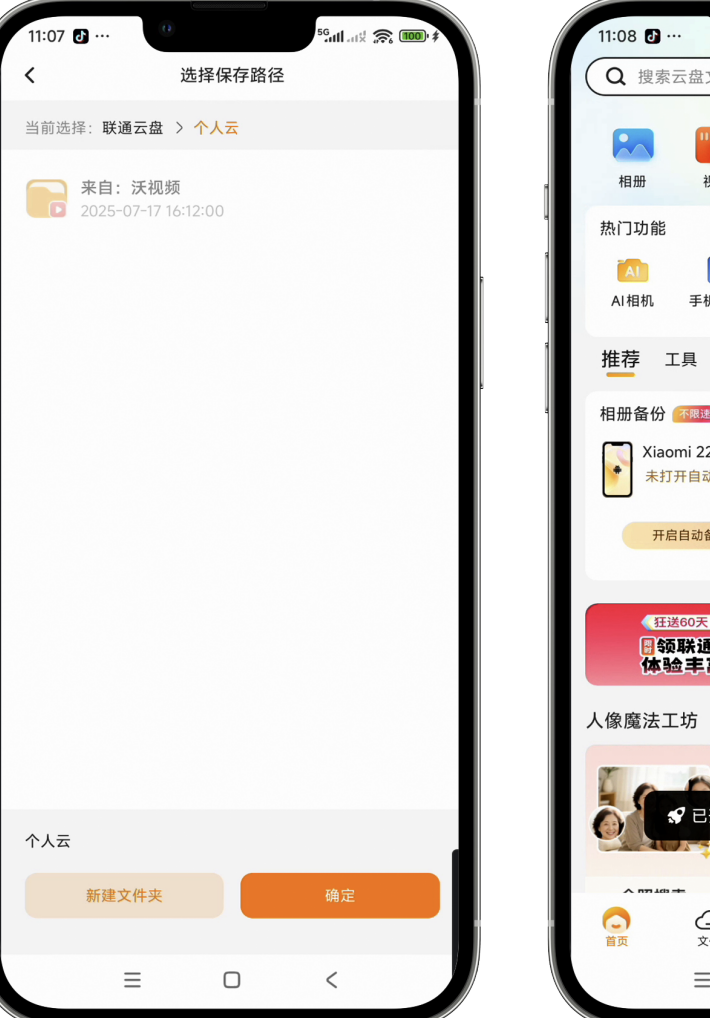

**(i)** *Action: Select "Personal Cloud"*

**(j)** *Action: Tap "Confirm"*

**Figure 7.10:** *Instruction: View e-invoice for last recharge in China Unicom and save as PDF to Personal Cloud in Unicom Cloud Drive*

*Execution Process* As shown in Figure 7.10, during initial instruction execution, the agent navigates from homepage to “Recharge” module, then to “E-invoice” module. This long-horizon operation contains multiple key steps executed in logical order without path deviation or step omission. This coherence stems from deep understanding of Unicom APP business logic.

The execution addresses the pain point of invoice management. Traditional management requires manual lookup, download, and APP switching. The agent integrates multiple steps into one automated flow. Precise location of “last” record ensures users get the latest standardized invoice file.

The task is successfully executed, demonstrating capabilities in precise intent parsing, long-horizon process control, cross-functional module collaboration, and file format handling, comprehensively validating comprehensive service capabilities in complex business scenarios.

### 7.2.2 Complex Scenario Task Evaluation Results and Analysis

#### Aging-Friendly and Vision-Impaired Tasks

Task 2.1: Listen to “A Record of a Mortal’s Journey to Immortality” on Qimao Free Novel

Task Objective: This task demonstrates precise content localization and voice accessibility support, highlighting intelligent interaction’s core value in accessibility scenarios.

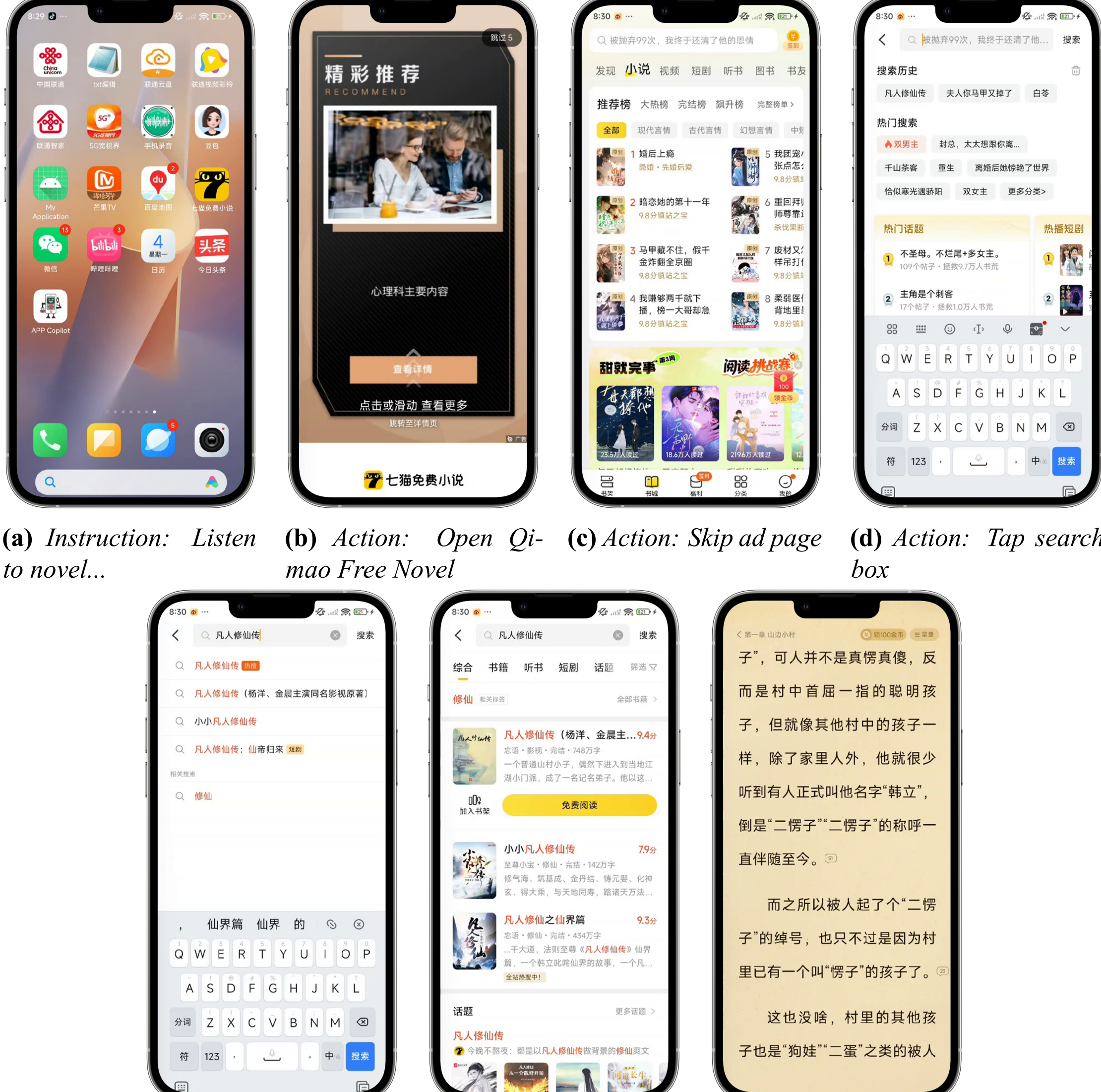

**(a)** *Instruction: Listen to novel...*

**(b)** *Action: Open Qimao Free Novel*

**(c)** *Action: Skip ad page*

**(d)** *Action: Tap search box*

**(e)** *Action: Search novel name*

**(f)** *Action: Tap confirm*

**(g)** *Action: Enter page and listen*

**Figure 7.11:** *Instruction: Listen to "A Record of a Mortal's Journey to Immortality" on Qimao Free Novel*

*Execution Process* As shown in Figure 7.11, for elderly or visually impaired users, manually searching novels in complex interfaces presents challenges. After launching Qimao Novel APP, the agent locates the search box, inputs the novel name, and quickly locks the target work. This approach minimizes visual recognition and manual input barriers.

Activation and adaptation of the voice module is core to this task. After entering the reading interface, the agent activates voice reading rather than displaying text. For visually impaired users, this removes visual information barriers; for elderly users, it alleviates visual fatigue. During reading, the agent automatically adapts to chapter transitions and controls speed/pauses.

The task is successfully completed, demonstrating precise content localization and voice accessibility support, providing equal access to digital content for elderly and visually impaired groups.

**Long-horizon Tasks**

Task 2.2: Check the highest-rated nearby restaurant on Dianping

Task Objective: This task demonstrates capabilities in task decomposition/planning and contextual understanding.

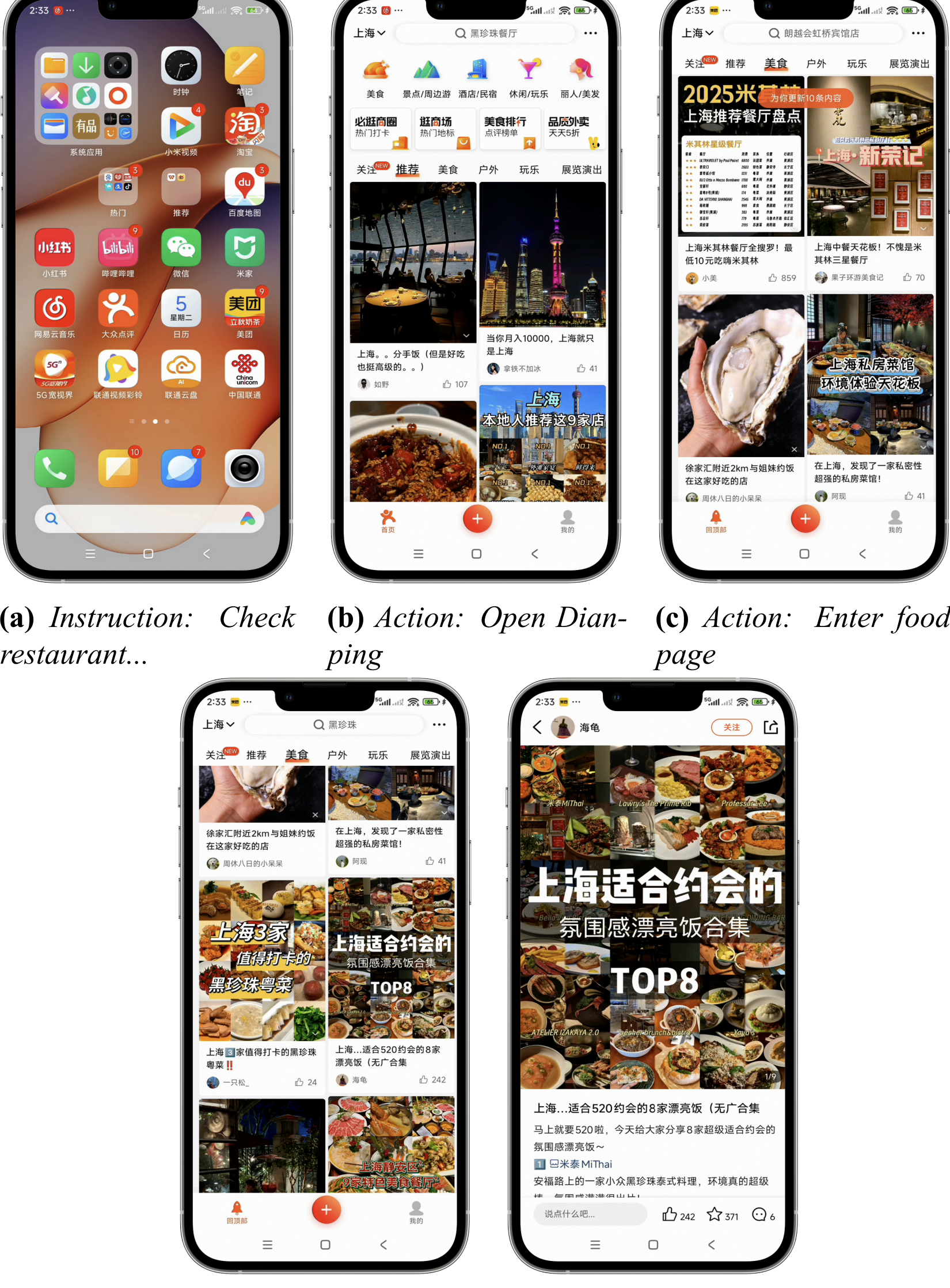

**(a)** *Instruction: Check restaurant...* **(b)** *Action: Open Dianping* **(c)** *Action: Enter food page* **(d)** *Action: Swipe page* **(e)** *Action: View recommendation posts*

**Figure 7.12:** *Instruction: Check highest-rated nearby restaurant on Dianping (First Execution)*

*Execution Process* Figure 7.12 shows the first execution choosing the "homepage recommendation post" path. Dianping's homepage typically aggregates personalized recommendations based on location and history. The agent prioritizes using platform recommendations rather than starting from scratch. This path leverages platform algorithms to reduce operational steps while meeting needs.

Figure 7.13 shows the second execution using the "search bar filtering" path, demonstrating active control over requirement boundaries. The core logic is precise requirement decomposition: first search "restaurant" for full coverage, then set "Nearby" distance filter, finally switch to "Highest Rating" sorting. Each step directly corresponds to core instruction conditions.

The path difference reflects the agent's dynamic balance between "efficiency" and "accuracy". The recommendation path prioritizes efficiency when content matches requirements; the search path prioritizes accuracy when recommendation reliability is uncertain. This balancing depends on real-time judgment of requirement clarity and recommendation reliability.

The task is successfully completed, demonstrating multi-strategy adaptation capabilities in complex application scenarios.

#### Cross-APP Tasks

Task 2.3: Send my itinerary from Umetrip to WeChat contact Manager Wu

Task Objective: This task validates capabilities in intent recognition, complex task decomposition, and planning.

*Execution Process* As shown in Figure 7.14, the agent navigates to the itinerary module and extracts core elements through structured parsing of the page. This automatically identifies and extracts key data (departure, destination, time, flight number), ensuring clear, complete information transmission.

After extraction, the agent switches from TravelSky to WeChat, involving system-level operations. The agent maintains state coherence: after completing extraction in TravelSky, it switches to WeChat via system operations, locates "Manager Wu" through search/contacts, enters chat interface, pastes integrated itinerary, and triggers sending. This cross-end execution relies on deep cognition of multi-APP interface logic.

The task is successfully completed, demonstrating cross-APP collaboration, information extraction/integration, and long-horizon task control capabilities, validating full-chain intelligent service in complex scenarios.

#### Cross-Device Tasks

Task 2.4: Buy a gift for Lili based on her 5G Kuan Shijie history

Task Objective: This task demonstrates multi-agent collaboration, user preference analysis, and error tolerance.

*Execution Process* As shown in Figure 7.15, Lili's device stores 5G Kuan Shijie viewing history data, while the user's device completes gift purchasing. This demonstrates cross-device multi-agent collaboration, user preference extraction, and cross-application decision-making.

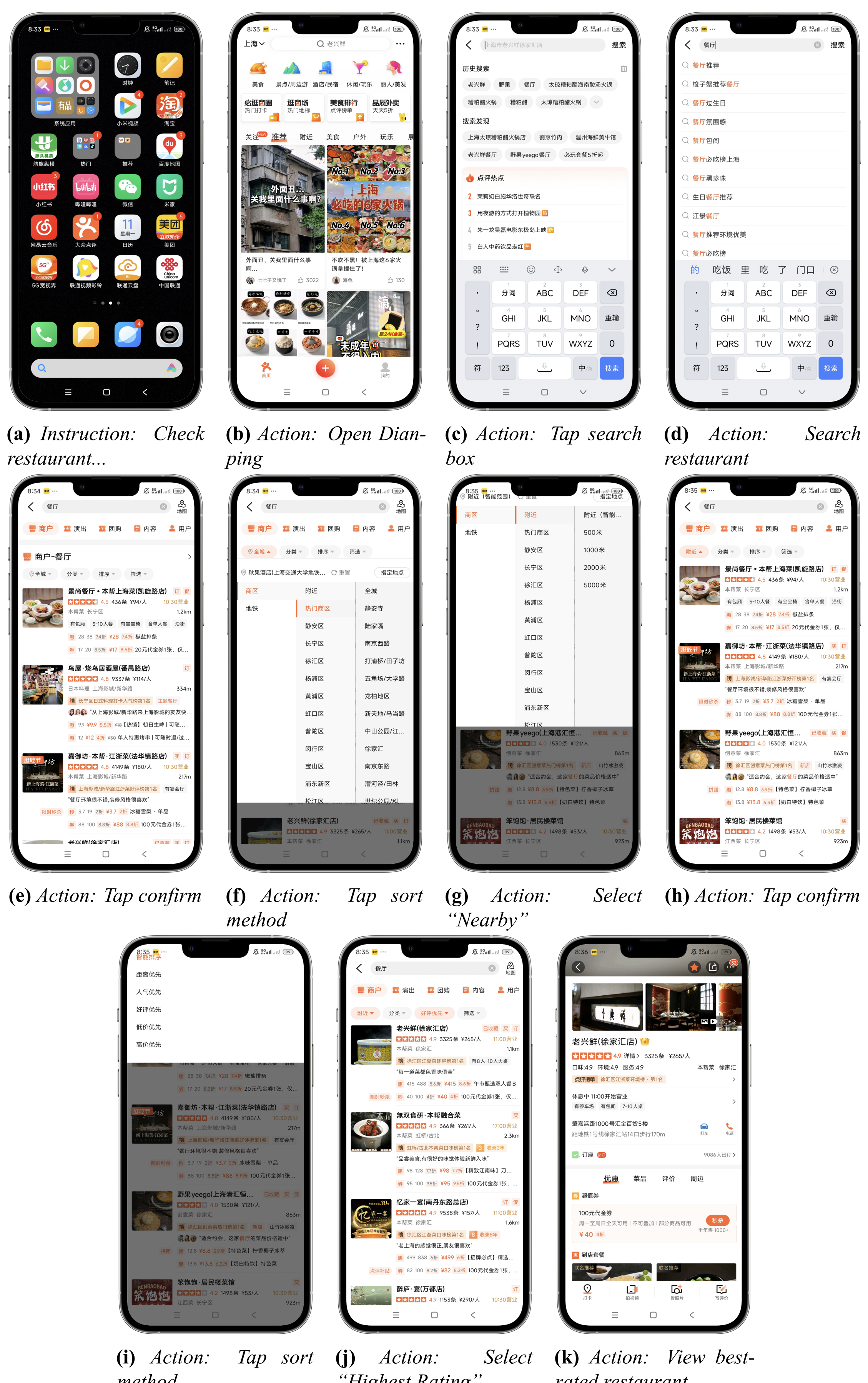

**(a)** *Instruction: Check restaurant...* **(b)** *Action: Open Dianping* **(c)** *Action: Tap search box* **(d)** *Action: Search restaurant*

**(e)** *Action: Tap confirm* **(f)** *Action: Tap sort method* **(g)** *Action: Select "Nearby"* **(h)** *Action: Tap confirm*

**(i)** *Action: Tap sort method* **(j)** *Action: Select "Highest Rating"* **(k)** *Action: View best-rated restaurant*

**Figure 7.13:** *Instruction: Check highest-rated nearby restaurant on Dianping (Second Execution)*

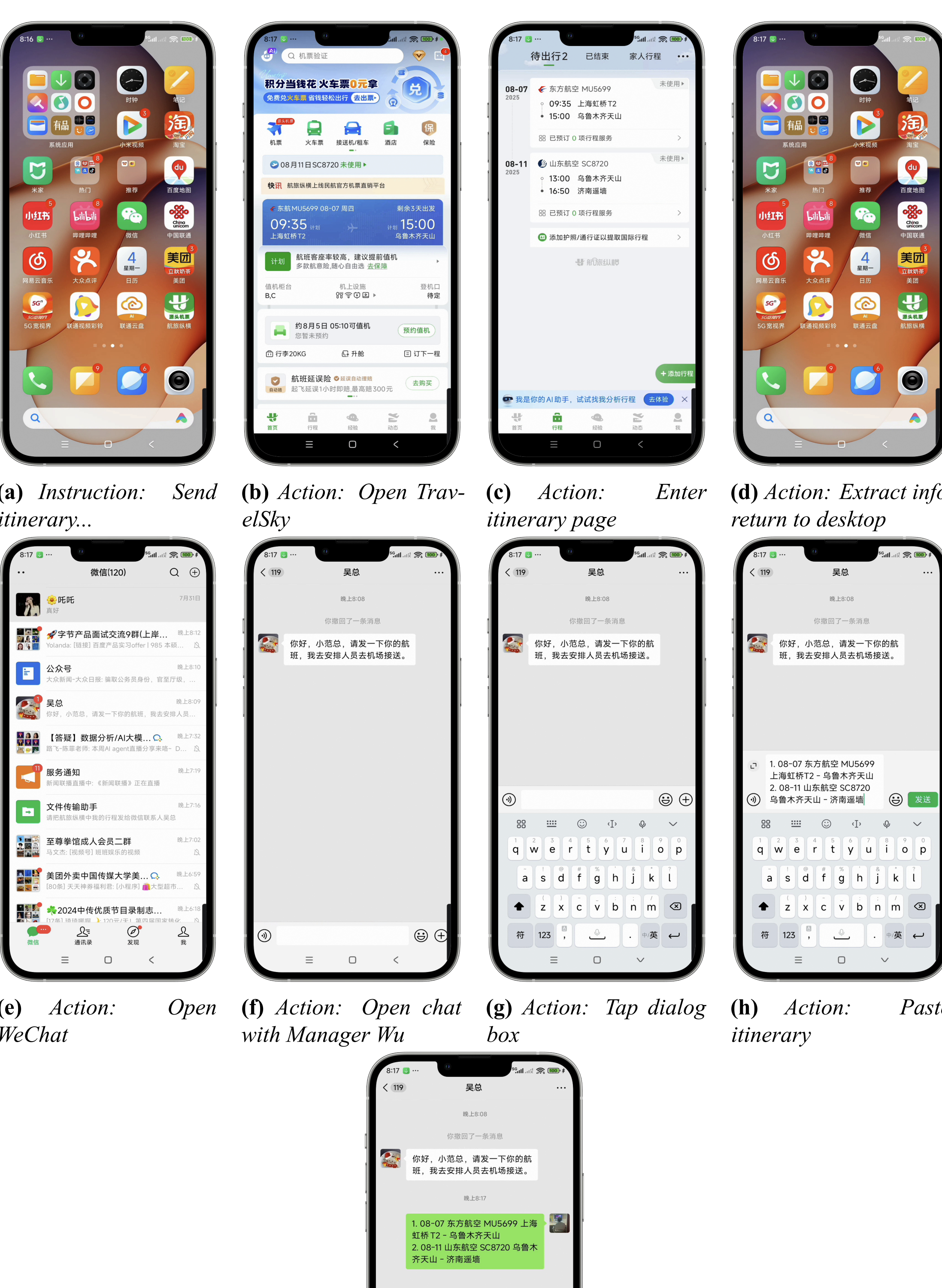

**(a)** *Instruction: Send itinerary...*

**(b)** *Action: Open TravelSky*

**(c)** *Action: Enter itinerary page*

**(d)** *Action: Extract info, return to desktop*

**(e)** *Action: Open WeChat*

**(f)** *Action: Open chat with Manager Wu*

**(g)** *Action: Tap dialog box*

**(h)** *Action: Paste itinerary*

**(i)** *Action: Send message*

**Figure 7.14:** *Instruction: Send my itinerary from TravelSky to WeChat contact Manager Wu*

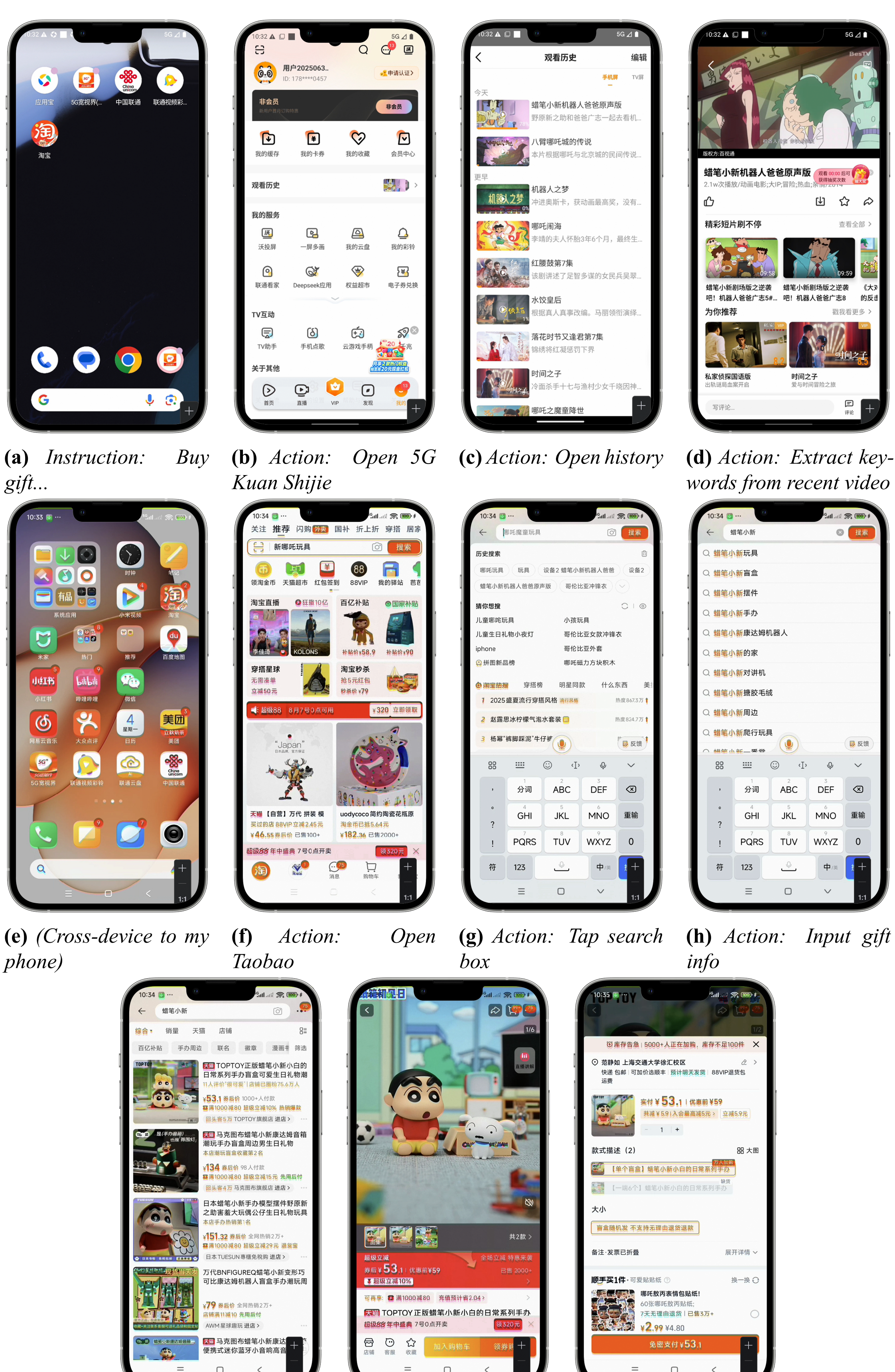

**(a)** *Instruction: Buy gift...*

**(b)** *Action: Open 5G Kuan Shijie*

**(c)** *Action: Open history*

**(d)** *Action: Extract keywords from recent video*

**(e)** *(Cross-device to my phone)*

**(f)** *Action: Open Taobao*

**(g)** *Action: Tap search box*

**(h)** *Action: Input gift info*

**(i)** *Action: Tap search*

**(j)** *Action: Tap gift*

**(k)** *Action: Add to cart*

**Figure 7.15:** *Instruction: Buy gift for Lili based on her 5G Kuan Shijie history*

After authorization verification, the agent locates the history module and extracts key information from the most recent video. But raw video lists can't directly guide gift selection. Here, the agent extracts IP keywords from "Crayon Shin-chan" to infer potential interests. This transcends simple data transfer by achieving a leap from data to preference to demand through content understanding.

On the user's device, the agent receives "Crayon Shin-chan" keywords and launches Taobao. It locates relevant gifts through search, maintaining process coherence across multiple operations.

Critically, the task highlights the core value of cross-device service,breaking device barriers to achieve precise data-to-service docking. Traditional scenarios require manual preference inquiry and product search; the agent automates the entire process from data collection to product recommendation.

Task 2.5: Purchase gifts based on Lili's and Fanfan's 5G Kuan Shijie history

Task Objective: This task extends agent capabilities from serving individual users to multi-user collaborative paradigms.

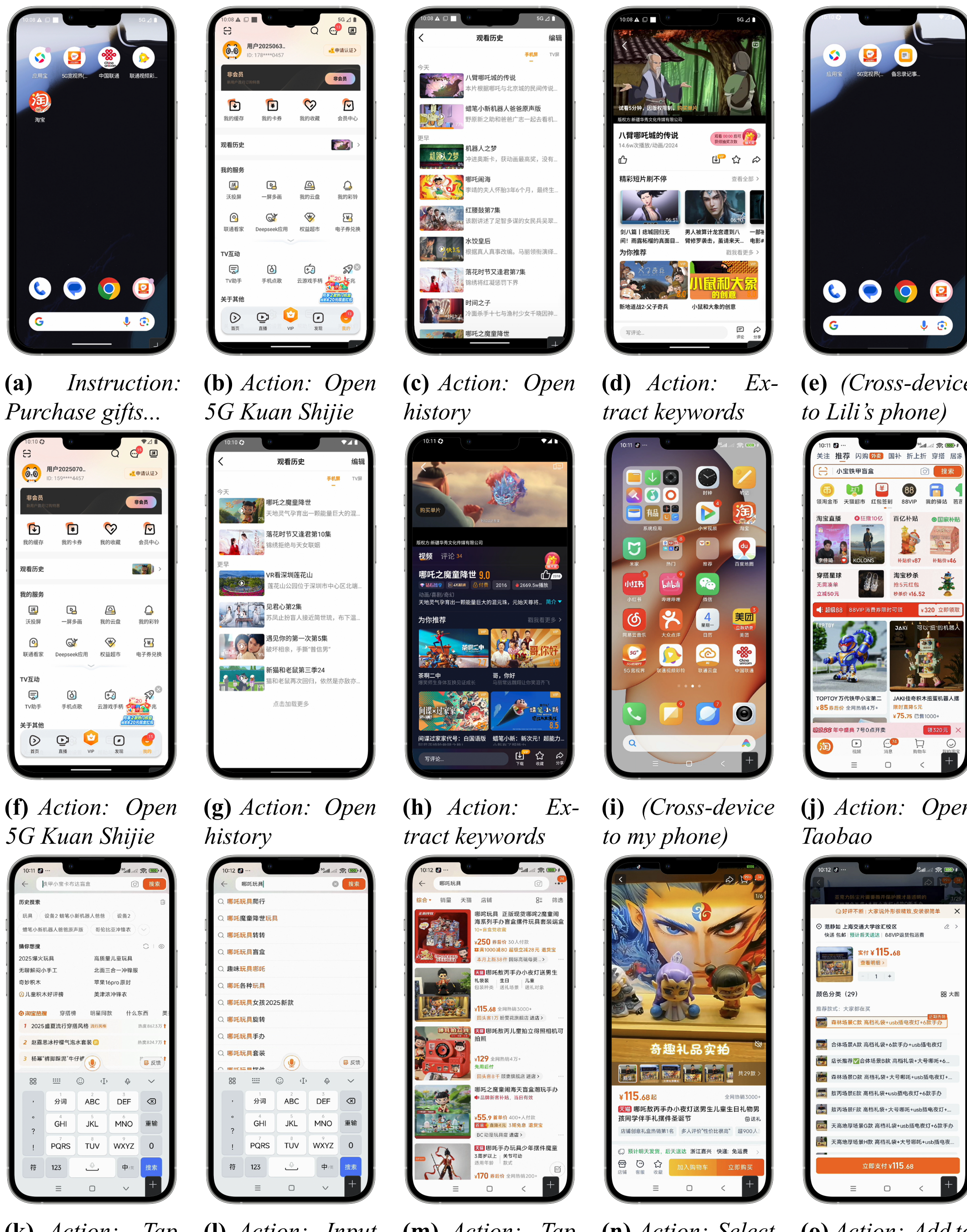

**(a)** *Instruction: Purchase gifts...*
**(b)** *Action: Open 5G Kuan Shijie*
**(c)** *Action: Open history*
**(d)** *Action: Extract keywords*
**(e)** *(Cross-device to Lili's phone)*
**(f)** *Action: Open 5G Kuan Shijie*
**(g)** *Action: Open history*
**(h)** *Action: Extract keywords*
**(i)** *(Cross-device to my phone)*
**(j)** *Action: Open Taobao*
**(k)** *Action: Tap search box*
**(l)** *Action: Input gift info*
**(m)** *Action: Tap search*
**(n)** *Action: Select gift*
**(o)** *Action: Add to cart*

**Figure 7.16:** *Instruction: Purchase gifts based on Lili's and Fanfan's 5G Kuan Shijie history*

*Execution Process* As shown in Figure 7.16, Lili's and Fanfan's devices store viewing history data, while the user's device completes gift purchasing. This extends agent capabilities from individual users to multi-user collaborative operations. This isn't simple technical stacking but a paradigm shift from personal assistants to distributed collaboration networks.

Each agent focuses on parsing its own 5G Kuan Shijie history to extract individual preferences. The user's agent integrates preference data from both ends to drive targeted gift selection. In this architecture, each agent has independent computational space and decision boundaries, preserving core attributes representing individual intent while breaking limitations through collaboration.

The multi-device task confirms the feasibility of a distributed intelligence system. First, addressing reasoning and coordination challenges with incomplete information: Lili's and Fanfan's agents only know their own task status; the user's agent cannot directly access raw data on other devices. Through data desensitization and intent recognition mechanisms, agents collaborate accurately despite information gaps.

Second, addressing communication and negotiation mechanisms: agents achieve precise intent transmission through unified protocols despite heterogeneous systems. The successful execution validates that the mobile agent system has upgraded from single-agent to a system-level architecture with multi-agent collaboration, distributed state modeling, and mechanism design capabilities. This upgrade's core value enables intelligent services to break single-user boundaries and complete complex cross-domain long-horizon tasks through multiple autonomous agents collaborating, moving toward realizing the "theoretically expandable to massive terminals" vision of collective intelligence.

### 7.2.3 Real Scenario Task Evaluation Results and Analysis

In the real scenario capability evaluation phase, based on the technical framework of the AppCopilot system, our team independently developed an Android mobile application using Java as the development language. This study conducted a system performance evaluation of the developed Android mobile application in real-world scenarios. The research designed a complete chain of daily user behaviors as the evaluation scenario: The user wants to call a friend for dinner but finds the mobile phone has no credit, so they recharge; then opens Dianping to search for the highest-rated nearby restaurant; then needs to drive to the restaurant, so opens Amap for navigation; finally, after the meal, bookmarks the restaurant. This involves the following four subtasks:

1. Task 3.1: Recharge credit on China Unicom: Simulates users completing online credit recharge operations when discovering insufficient balance during calls. Specific instructions and process screenshots are shown in Fig 7.17;
2. Task 3.2: Open Dianping to search for the highest-rated nearby restaurant: Implements restaurant information retrieval and sorting (by descending rating) through a third-party platform (Dianping). Specific instructions and process screenshots are shown in Figure 7.18;
3. Task 3.3: Navigate to the restaurant using Amap: Calls map navigation service (Amap) to search for the destination and enter the navigation interface. Specific instructions and process screenshots are shown in Figure 7.19;

4. Task 3.4: Bookmark the restaurant on Dianping: Completes the post-consumption bookmarking operation on the third-party platform (Dianping). Specific instructions and process screenshots are shown in Figure 7.20.

Task Objectives: This evaluation scheme aims to verify the system's generalization, accuracy, and efficiency in real complex scenarios. It simulates high-frequency daily user scenarios (recharge, restaurant search, navigation, bookmarking), demonstrating the practicality of the mobile application mobile agent for real needs.

*Execution Process* Currently, there is a growing demand for seamless multi-application collaborative operations. This evaluation, through the complete closed-loop of credit recharge → restaurant search → navigation → bookmarking, reflects the potential of mobile agent in enhancing life efficiency and optimizing service experience. For example, automatic recharge reminders when credit is low, precise local service recommendations, and integrated navigation can significantly reduce user operation costs, especially benefiting groups unfamiliar with digital interactions (e.g., the elderly), promoting inclusive development of digital services. As shown below, the Android mobile application developed based on AppCopilot can accurately recognize and jump to the corresponding APP to complete tasks in all four subtasks, demonstrating the system's generalization, accuracy, and efficiency in real complex scenarios.

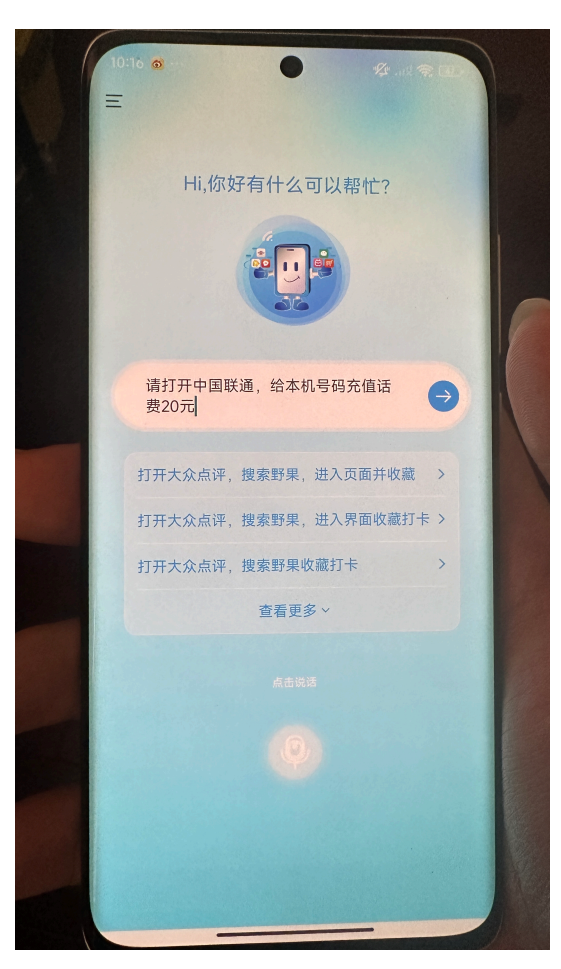

**(a)** *Operation: Voice input command*

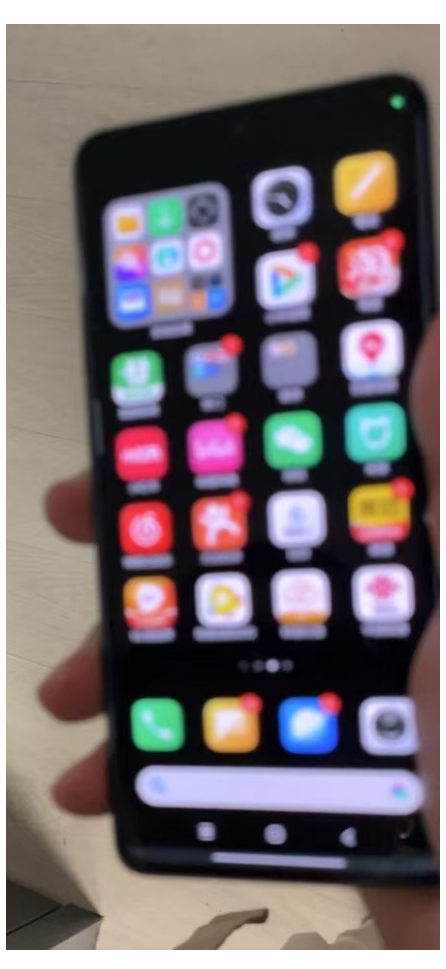

**(b)** *Operation: Open China Unicom APP*

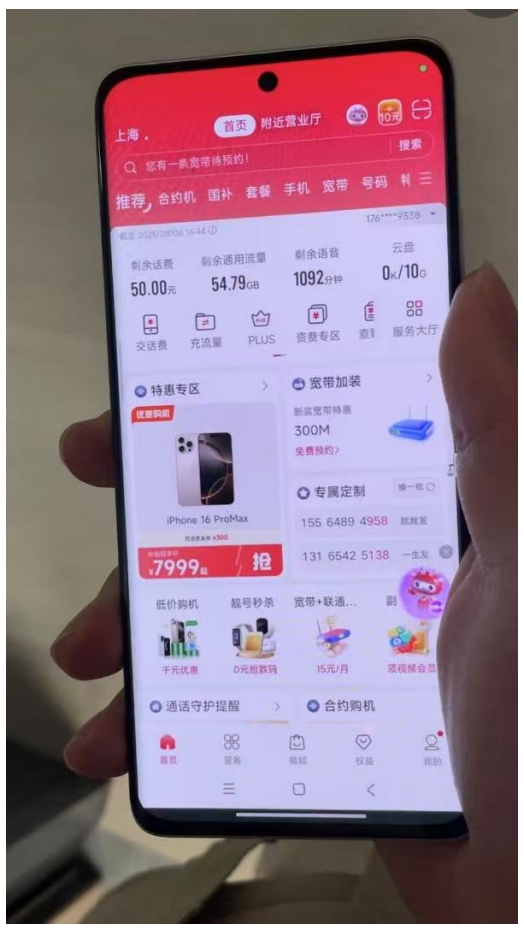

**(c)** *Operation: Click to recharge credit*

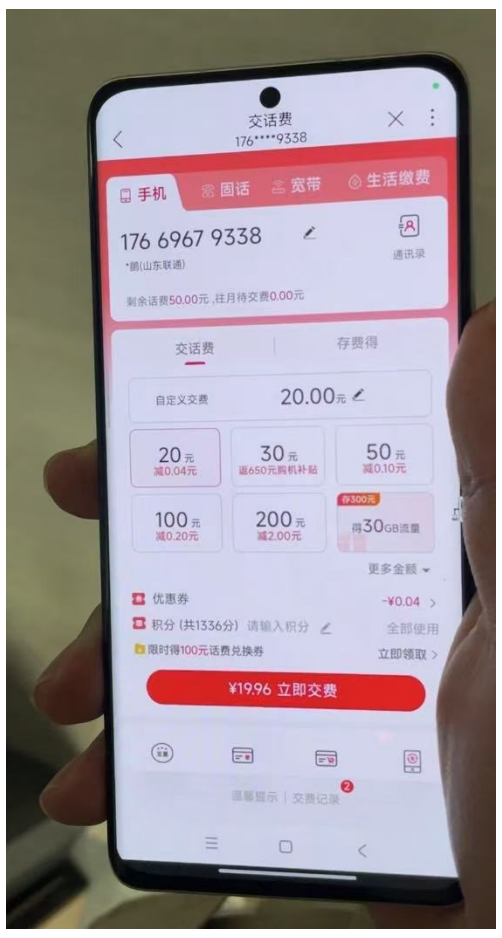

**(d)** *Operation: Select recharge amount*

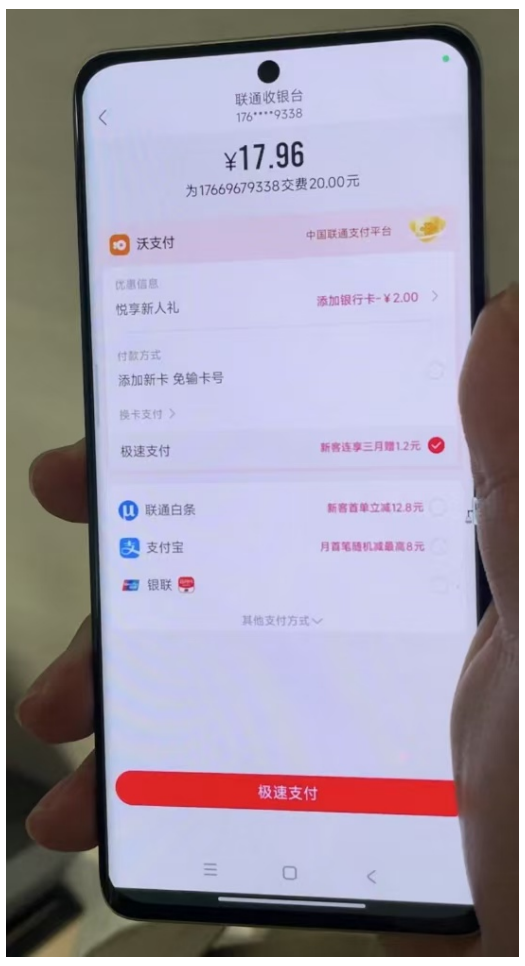

**(e)** *Operation: Enter payment page*

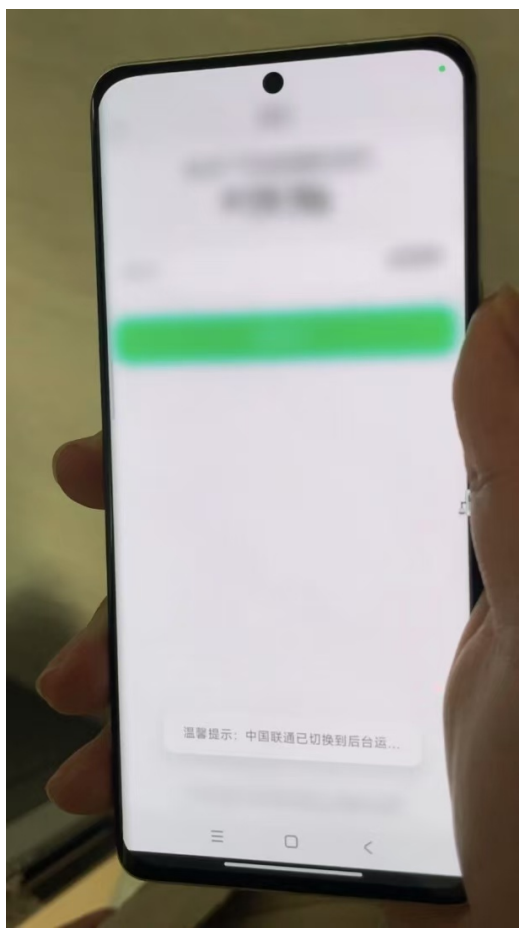

**(f)** *Operation: Confirm payment*

**Figure 7.17:** *Instruction: Recharge 20 yuan for this device number on China Unicom*

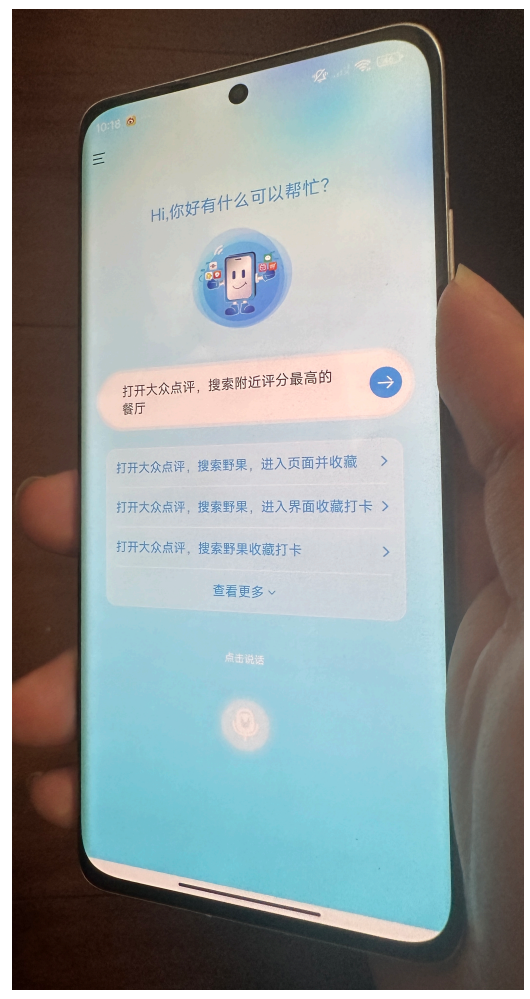

(a) *Operation: Voice input command*

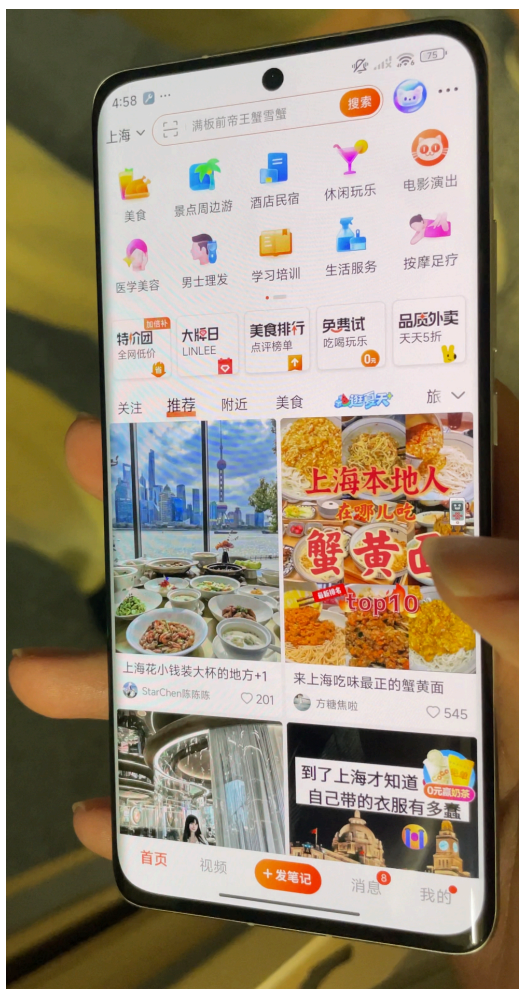

(b) *Operation: Open Dianping*

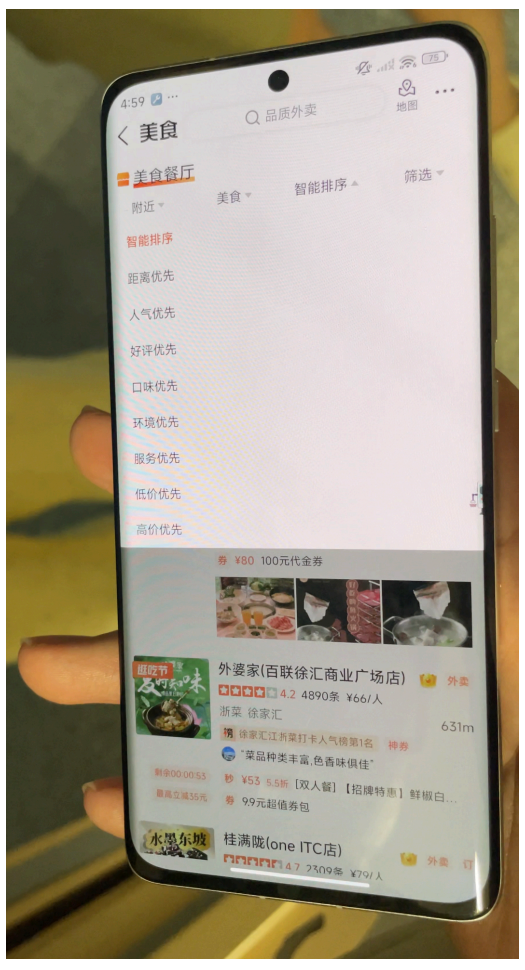

(c) *Operation: Click "Highest Rating First"*

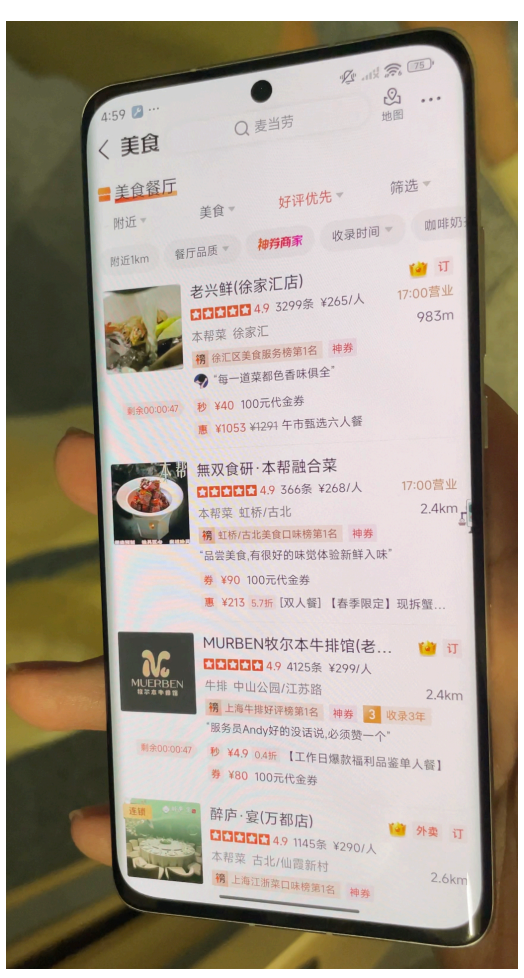

(d) *Operation: Locate the restaurant*

**Figure 7.18:** *Instruction: Search for the highest-rated nearby restaurant on Dianping*

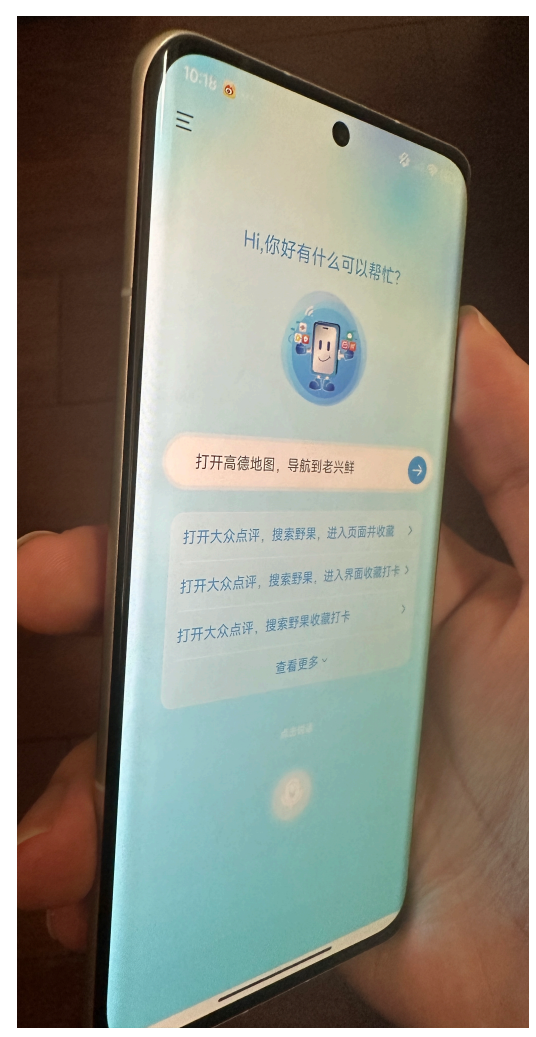

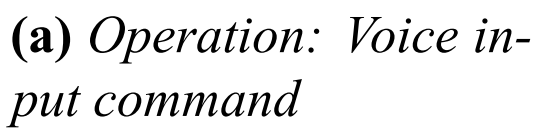

(a) *Operation: Voice input command*

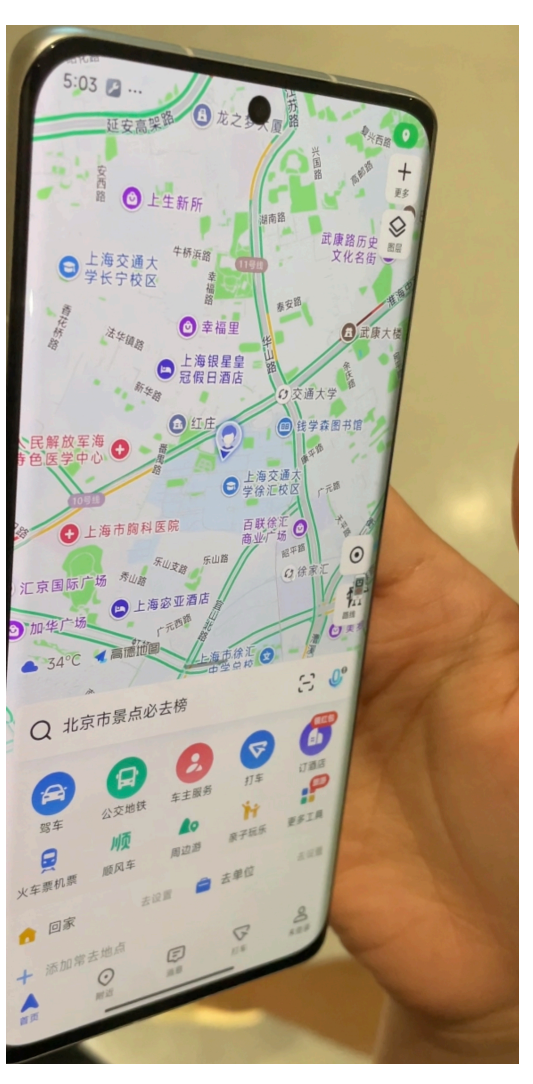

(b) *Operation: Enter Amap*

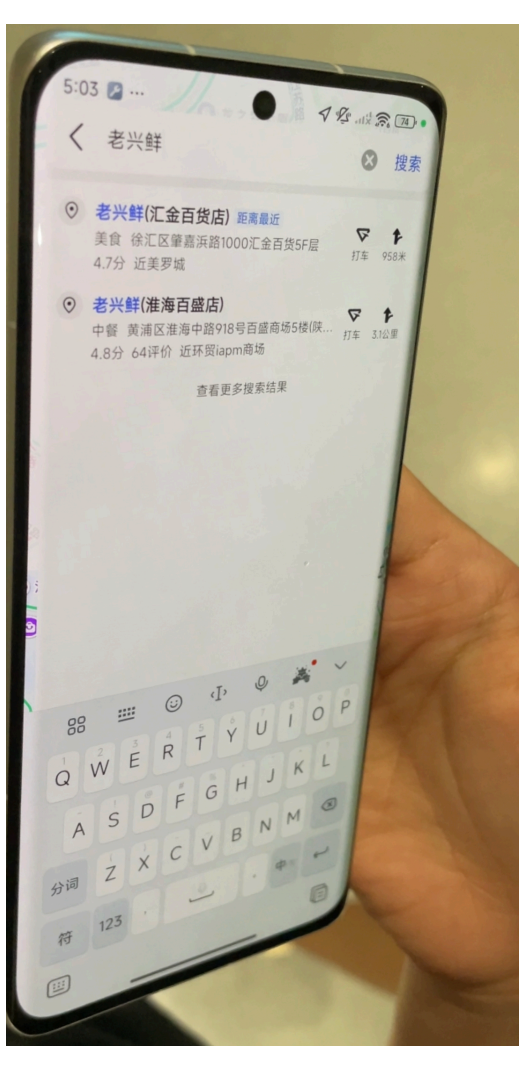

(c) *Operation: Search for Lao Xing Xian*

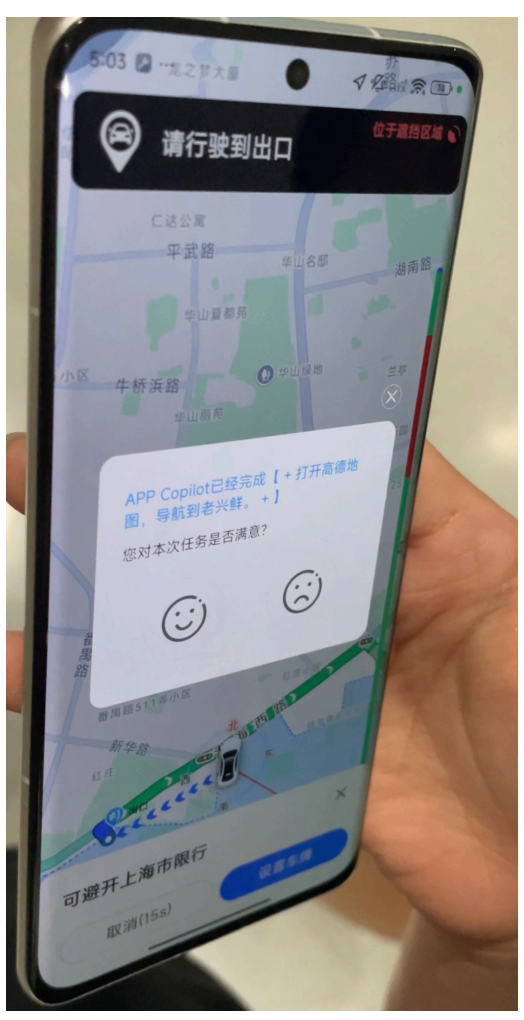

(d) *Operation: Enter navigation page*

**Figure 7.19:** *Instruction: Open Amap to navigate to Lao Xing Xian*

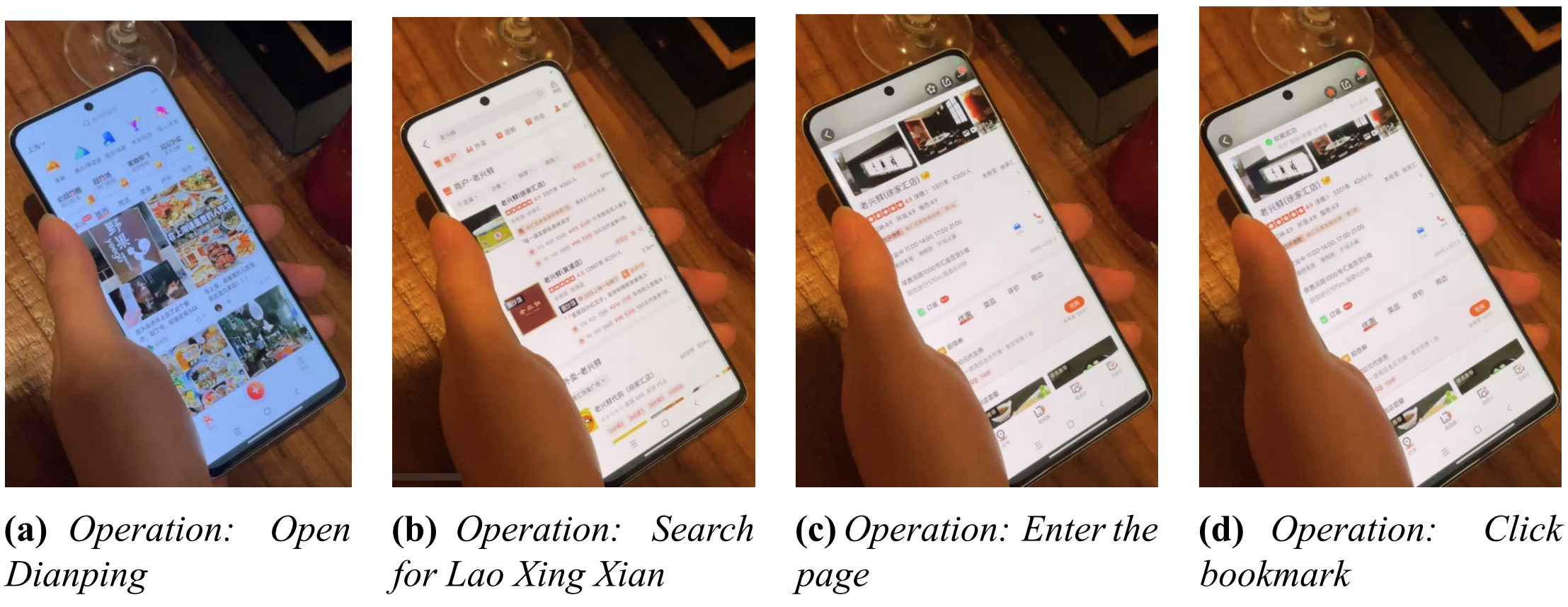

**(a)** *Operation: Open Dianping* **(b)** *Operation: Search for Lao Xing Xian* **(c)** *Operation: Enter the page* **(d)** *Operation: Click bookmark*

**Figure 7.20:** *Instruction: Bookmark Lao Xing Xian shop on Dianping*

### 7.2.4 Scenario Summary

The above presents the main scenario evaluation tasks. Guided by the four core capabilities of accuracy, generalization, long-horizon capability, and efficiency, and combined with scenario characteristics, the evaluation plan is systematically designed into three major categories: Basic scenarios, complex interaction scenarios, and real-life work scenarios. The evaluation results are summarized in Table 7.2. All scenarios were validated through standardized and detailed procedures. The results not only confirm the effectiveness of the solutions proposed in this study for the four core capability issues but also visually demonstrate that the project achieves precise task execution in both Basic scenarios and complex scenarios. Especially in long-horizon tasks, relying on the collaborative optimization of multiple mechanisms, the model exhibits stable and reliable execution performance, while the overall task execution efficiency is significantly improved.

**Table 7.2:** *Evaluation Metrics and Task Scenario Correlation Table*

| Evaluation Metric | Task Scenario | Associated Tasks |
| --- | --- | --- |
| **Accuracy** | **Basic Scenarios** | |
| | - Basic Communication Tasks | Task 1.1, Task 1.2 |
| | - Transaction Service Tasks | Task 1.3, Task 1.4 |
| | - Entertainment Content Tasks | Task 1.5, Task 1.6 |
| | - Account and Information Query Tasks | Task 1.7 |
| | - Tool Efficiency Tasks | Task 1.8 |
| | **Complex Scenarios** | |
| | - Aging-Friendly and Vision-Impaired Tasks | Task 2.1 |
| | - long-horizon Tasks | Task 2.2 |
| | - Cross-APP Tasks | Task 2.3 |
| | - Cross-Device Tasks | Task 2.4, Task 2.5 |
| | **Real Scenarios** | Task 3.1, Task 3.2, Task 3.3, Task 3.4 |
| **Generalization** | **Basic Scenarios** | |
| | - Transaction Service Tasks | Task 1.4 |
| | - Entertainment Content Tasks | Task 1.5, Task 1.6 |
| | - Account and Information Query Tasks | Task 1.7 |
| | - Tool Efficiency Tasks | Task 1.8 |
| | **Complex Scenarios** | |
| | - Aging-Friendly and Vision-Impaired Tasks | Task 2.1 |
| | - long-horizon Tasks | Task 2.2 |
| | - Cross-APP Tasks | Task 2.3 |
| | - Cross-Device Tasks | Task 2.4, Task 2.5 |
| | **Real Scenarios** | Task 3.2, Task 3.3, Task 3.4 |
| **long-horizon Capability** | **Basic Scenarios** | |
| | - Entertainment Content Tasks | Task 1.6 |
| | - Account and Information Query Tasks | Task 1.7 |
| | - Tool Efficiency Tasks | Task 1.8 |
| | **Complex Scenarios** | |
| | - long-horizon Tasks | Task 2.2 |
| | - Cross-APP Tasks | Task 2.3 |
| | - Cross-Device Tasks | Task 2.4, Task 2.5 |
| | **Real Scenarios** | Task 3.1, Task 3.2, Task 3.4 |
| **Efficiency** | **Basic Scenarios** | |
| | - Basic Communication Tasks | Task 1.2 |
| | - Transaction Service Tasks | Task 1.4 |
| | - Entertainment Content Tasks | Task 1.5, Task 1.6 |
| | - Account and Information Query Tasks | Task 1.7 |
| | - Tool Efficiency Tasks | Task 1.8 |
| | **Complex Scenarios** | |
| | - long-horizon Tasks | Task 2.2 |
| | - Cross-APP Tasks | Task 2.3 |
| | - Cross-Device Tasks | Task 2.4, Task 2.5 |
| | **Real Scenarios** | Task 3.2, Task 3.4 |

# 8

# Future Directions

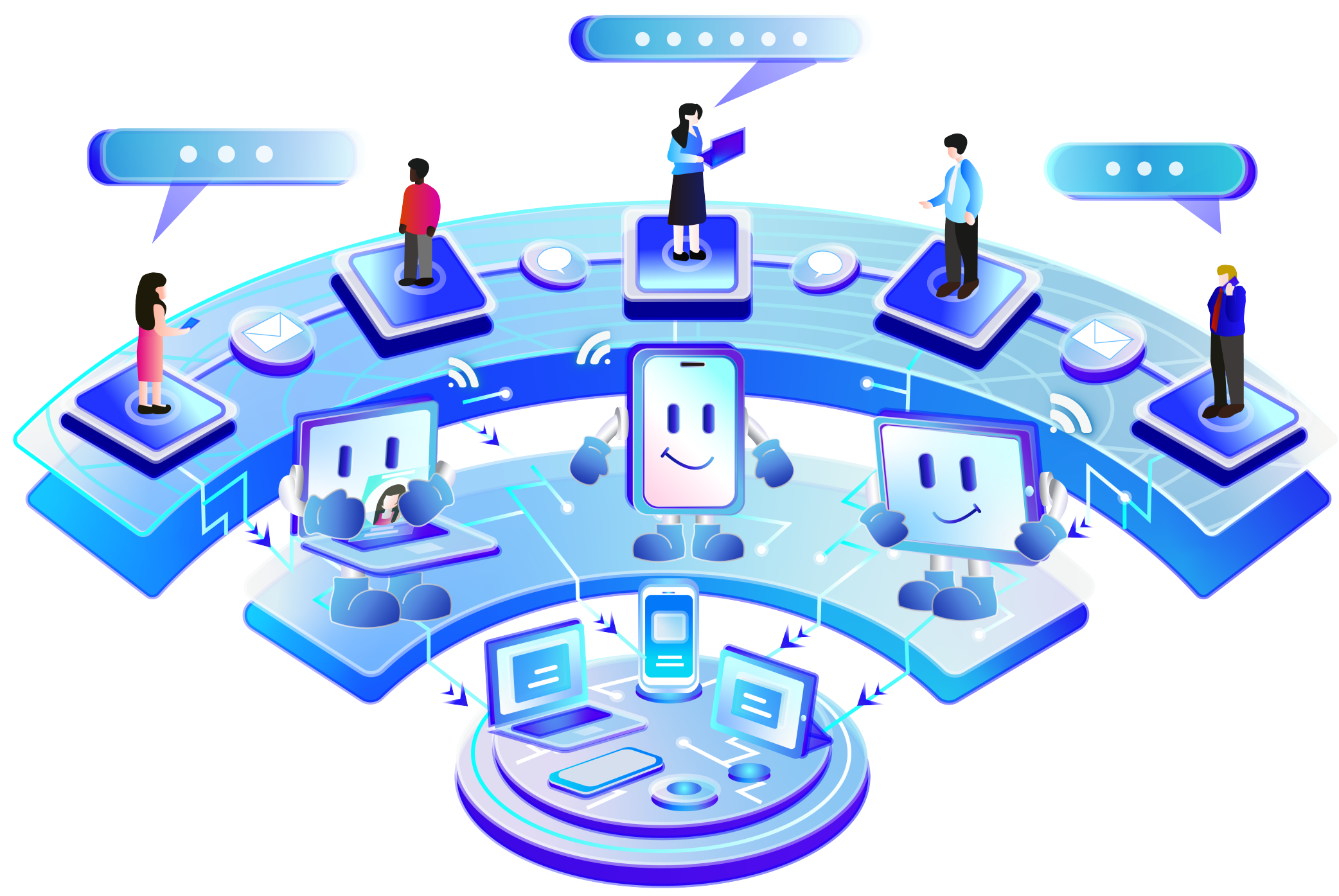

**Figure 8.1:** *Human-Agents-Physical Device Collaboration Architecture Diagram*

Looking forward, the envisioned implementation logic is illustrated in Figure 8.1, which presents a three-layer interaction framework: Human-Agents-Devices Layers. In this vision, the human layer initiates diverse needs across users, devices, and applications, transmitting personalized intentions, scenario-specific instructions, and relevant context to the agents layer. Acting as the core transformation hub, the agents layer would not only orchestrate complex cross-device and cross-APP tasks but also possess autonomy, enabling agents to independently plan and ex-ecute user-assigned tasks. Finally, the physical device Layer would carry out the instructions through coordinated device control, completing the full chain from human needs to physical operations.

To realize this envisioned framework, the further evolution of mobile agents should focus on four core capabilities: generalization, accuracy, long-horizon capability, and efficiency. These capabilities directly support the framework: generalization enables agents to handle diverse devices, applications, and user scenarios; accuracy ensures precise execution of personalized tasks; long-horizon capability allows autonomous planning and completion of multi-step operations; and efficiency guarantees flexible, scalable, and timely human-device interactions.

Overall, this framework envisions a flexible, efficient, and autonomous human-agent-device ecosystem that leverages these four core capabilities to overcome traditional single-agent or single-device limitations, supporting adaptive, cross-scenario task execution and paving the way for large-scale, stable, and innovative applications of mobile agents.

## 8.1 Generalization: Data Scale Expansion and Training Paradigm Innovation

The generalization ability of an agent determines its adaptability in unknown or rapidly changing environments. Future optimization can focus on the following three aspects.

*Expanding Heterogeneous Data Collection and Coverage* To enhance the platform's robustness across diverse terminals, it is essential to collect data from emerging and widely used applications, multilingual interfaces, and complex tasks. Achieving this requires systematic annotation pipelines and the automated collection of logs from mobile devices and other terminals, enabling the platform to comprehensively capture user interactions and interface features for improved generalization.

*Integrating Self-Supervised, Transfer, and Incremental Learning* Increasing data alone is insufficient to improve generalization; efficient training paradigms are essential. Self-supervised learning generates supervision from unlabeled data, transfer learning applies knowledge from previous tasks to new ones for faster adaptation, and incremental learning continuously updates models without retraining, preserving prior knowledge. Mobile agents should leverage these approaches to reduce manual labeling, accelerate adaptation, and handle shifting data distributions effectively.

*Establishing Open Data and Collaborative Learning Mechanisms* Sustained generalization requires openness and collaboration around data. Building on the open data movement, Open Data 2.0 allows data owners to retain control while enabling cooperation with model developers through decentralized agents and federated learning, where devices train locally and upload only parameter updates. This approach safeguards privacy while enabling collaborative innovation, further facilitating rapid expansion into new domains.

## 8.2 Accuracy: Dual-Track Integration of API and GUI with Enhanced Robustness

In enterprise deployment and large-scale adoption of mobile agents, operational accuracy is a key metric. Both API-based and GUI-based approaches are viable, each with distinct advantages: APIs provide high correctness in stable, well-defined interfaces, while GUI automation ensures adaptability to diverse UIs when APIs are missing or unstable. This motivates the need for integrated hybrid solutions that leverage the strengths of both approaches to maintain consistent accuracy across complex business scenarios.

*Robustness and Reliability in API Design* API-based approaches are well-suited for mobile agents because high accuracy stems from the API's design: clear, consistent endpoints and standardized semantics reduce errors, ensure backward compatibility, and support automated execution. Coupled with comprehensive error handling and self-healing mechanisms, APIs provide a stable, predictable foundation that enables mobile agents to perform complex enterprise tasks efficiently and accurately. However, many applications do not expose APIs for mobile agents, and those that do often offer limited interaction capabilities, which restricts task coverage and adaptability in real-world scenarios.

*Generality and Redundancy Safeguards in GUI Automation* In many legacy systems, closed platforms, or emerging applications, stable APIs are often unavailable. GUI automation provides strong generality and control by simulating mouse, keyboard, and touch gestures, leveraging visual recognition, OCR, and object tracking to handle multi-resolution, multilingual, and complex interfaces. Its zero-intrusion, low-dependency nature enables rapid deployment and iterative updates without altering system code. When APIs are absent, unreliable, or require rapid prototyping, GUI modules serve as a high-accuracy fallback, ensuring task continuity for mobile agents.

*Dual-Track Collaboration and Dynamic Decision-Making* Future mobile agent systems should treat API-based and GUI-based modes as complementary to ensure high task accuracy. APIs provide precise and efficient execution when available, while GUI automation maintains accuracy in unstable, missing, or complex interfaces. Systems should dynamically switch between modes based on context: prioritizing APIs for routine operations to maximize precision, using GUI to verify results or handle long-tail scenarios, and falling back to GUI immediately when APIs fail. By adaptively coordinating these modes, mobile agents can maintain consistent execution accuracy, leverage the strengths of both approaches, and handle a wide range of enterprise scenarios from stable APIs to unpredictable real-world interfaces.

## 8.3 Long-Horizon Capability: Personalized Memory, Cross-Application Collaboration, and Ecosystem Synergy

Complex tasks in real-world environments are inherently long-horizon, often involving sequential steps, conditional branching, and cross-application coordination. Mobile agents must handle dozens or hundreds of instructions, seamlessly switch between applications, and make decisions based on users' historical preferences and real-time intentions, requiring enhanced long-horizon capabilities for practical deployment.

*Deep User Data and Personalized Long-Term Memory* True long-horizon intelligence requires understanding users' long-term preferences and behavioral patterns, and making decisions in dynamically changing contexts. Long-Term Memory (LTM) enables mobile agents to store users' information, preferences, and behavioral patterns, allowing them to better handle long-horizon tasks. By building personalized databases and integrating them with contextual understanding and intent modeling, agents can decompose tasks, provide intelligent recommendations, maintain coherence across multi-session operations, and anticipate user needs more effectively.

*Cross-APP Collaboration and Multi-Agent Evolution* Long-horizon tasks often span multiple applications, involve complex conditional branches, and require long operation chains, which single agents can hardly handle alone. Cross-APP collaboration enables multiple specialized agents to coordinate subtasks across applications. Multi-agent evolution is necessary to further improve performance, and one effective approach is memory-based self-improvement, which allows agents to summarize past experiences, adapt strategies, and iteratively enhance further execution. Together, these mechanisms allow future mobile agents to dynamically plan cross-application paths, reuse accumulated experiences, and maintain robust performance on long-horizon, multi-application tasks.

*Cross-Ecosystem and Cross-Platform Capabilities* Some long-horizon tasks are inherently cross-ecosystem and cross-platform, spanning multiple devices and ecosystems. To handle such tasks, mobile agents should support cross-ecosystem and cross-platform collaboration. By leveraging system-level services such as accessibility APIs, scheduling interfaces, and notifications, agents can navigate across applications, recognize interface elements, and coordinate tasks reliably. Integrating these capabilities enables agents to perform seamless cross-platform operations, transfer data efficiently, and maintain security and stability, thereby supporting robust execution of long-horizon tasks and large-scale deployment.

## 8.4 Efficiency: Knowledge Density Enhancement and Lightweight Deployment on Devices

The inference efficiency and deployability of AI systems directly affect the cost and user experience of large-scale applications. As models grow in capability and multi-end collaboration becomes common, mobile agents must optimize not only model "slimming" but also dynamic scheduling between edge and cloud, as well as rapid task response.

*Model Compression and Knowledge Density Enhancement* Deep neural networks often contain redundant weights, causing high computation and energy costs that hinder deployment on mobile devices. Model compression techniques, such as pruning, quantization, and knowledge distillation, can reduce size without significant accuracy loss [21]. Knowledge distillation further transfers teacher knowledge to smaller student models, maintaining performance with fewer parameters. For LLM-powered mobile agents, these techniques together increase knowledge density per parameter and support efficient on-device inference.

*Dynamic Inference with Early Exit* Beyond model compression, dynamic inference strate-gies can further reduce costs. Early-exit techniques allow a network to stop inference once sufficient confidence is reached, lowering computation [70]. Mobile agents can adopt similar dynamic inference strategies, adjusting the depth of computation according to task complexity or input difficulty, skipping redundant processing for simpler tasks, and focusing resources on more complex operations. This enables flexible, context-aware execution, reduces latency and energy consumption, and ensures efficient performance across diverse tasks and devices.

*Edge-Cloud Collaboration and Dynamic Scheduling* Relying solely on cloud inference causes high latency and network costs, while fully local inference is constrained by device compute. Edge-cloud collaborative architectures address this by dynamically scheduling tasks based on complexity and resource availability. For example, a study on LLM services showed that such collaboration reduced energy consumption by over 50% while meeting latency requirements [80]. Mobile agents can adopt elastic local-cloud inference, executing simple, latency-sensitive tasks on-device and offloading complex or resource-intensive tasks to the cloud, with intelligent scheduling to balance performance, cost, and responsiveness.

*Local Caching for Frequent Interactions* In daily use, interactions often center on a few high-frequency tasks like launching APPs, text input, or system settings. Caching inference results can greatly improve responsiveness. Edge AI studies show that caching frequently used data across devices and cloud reduces storage and communication overhead [72]. For mobile agents, locally caching some UI elements and action sequences for common tasks enables near-instant responses, while model inference handles long-tail or less frequent scenarios to maintain generalization.

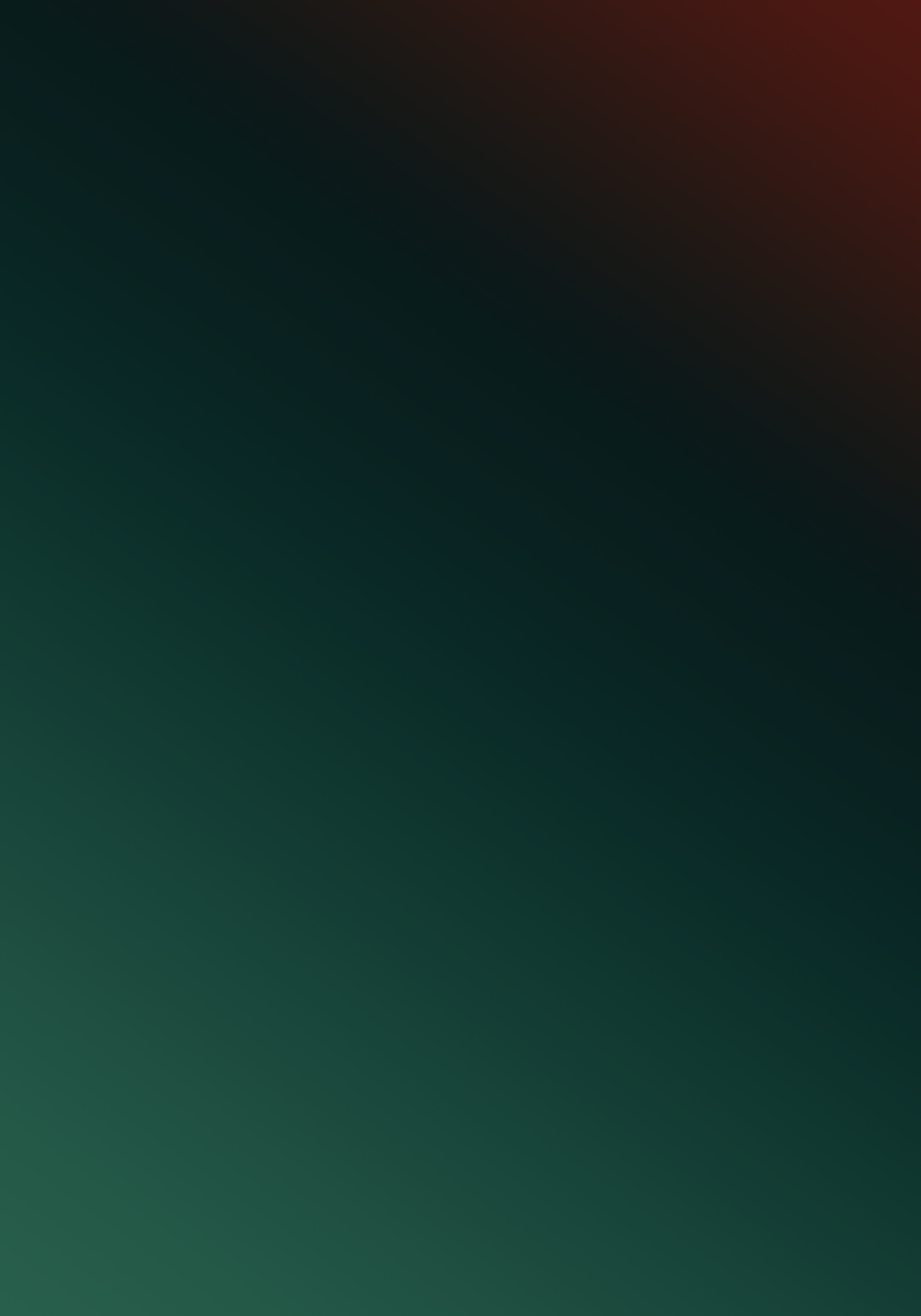